\documentclass[ a4paper, twocolumn]{article}
\usepackage[margin=1in]{geometry}

\usepackage[T1]{fontenc}
\usepackage{dfadobe}  

\usepackage{cite}  %
\usepackage[pdftex]{graphicx} \pdfcompresslevel=9

\usepackage{abstract}
\usepackage{subcaption}
\usepackage[labelfont=bf]{caption}
\usepackage{adjustbox,booktabs,multirow}
\usepackage{amsmath}
\usepackage{float}
\usepackage[table]{xcolor}    %
\definecolor{lightred}{HTML}{FF9999}
\definecolor{lightyellow}{HTML}{FFFF99}
\definecolor{lightorange}{HTML}{FFCC99}

\usepackage{url}
\usepackage{hyperref}
\usepackage{makecell}
\newsavebox\CBox

\renewenvironment{abstract}
{\begin{quote}
\noindent \rule{\linewidth}{.5pt}\par{\bfseries \abstractname.}}
{\medskip\noindent \rule{\linewidth}{.5pt}
\end{quote}
}

\newcommand{\MipNeRF}{\mbox{\textit{Mip-NeRF~360}}}
\newcommand{\TandT}{\textit{Tanks and Temples}}
\newcommand{\DeepBlending}{\textit{Deep~Blending}}
\newcommand{\SyntheticNeRF}{\textit{Synthetic~NeRF}}

\title{3DGS.zip: A survey on \\
      3D Gaussian Splatting Compression Methods}

\author{\parbox{\textwidth}{\centering 
    M.\,T. Bagdasarian$^{1}$%
    and P. Knoll$^{1}$%
    and Y. Li$^{3}$%
    and F. Barthel$^{1}$ \\%
    and A. Hilsmann$^{1}$%
    and P. Eisert$^{1,2}$%
    and W. Morgenstern$^{1}$
    }
        \\[0.5cm]
{\parbox{\textwidth}{\centering $^1$Fraunhofer HHI, Germany\\
         $^2$Humboldt-Universität zu Berlin, Germany\\
         $^3$Technische Universität Berlin, Germany
       }
}
}

\begin{document}

\twocolumn[
\begin{@twocolumnfalse}
\maketitle

\begin{abstract}
3D Gaussian Splatting (3DGS) has emerged as a cutting-edge technique for real-time radiance field rendering, offering state-of-the-art performance in terms of both quality and speed. 3DGS models a scene as a collection of three-dimensional Gaussians, with additional attributes optimized to conform to the scene's geometric and visual properties. Despite its advantages in rendering speed and image fidelity, 3DGS is limited by its significant storage and memory demands. These high demands make 3DGS impractical for mobile devices or headsets, reducing its applicability in important areas of computer graphics. 
To address these challenges and advance the practicality of 3DGS, this state-of-the-art report (STAR) provides a comprehensive and detailed examination of compression and compaction techniques developed to make 3DGS more efficient. We classify existing methods into two categories: compression, which focuses on reducing file size, and compaction, which aims to minimize the number of Gaussians. Both methods aim to maintain or improve quality, each by minimizing its respective attribute: file size for compression and Gaussian count for compaction. We introduce the basic mathematical concepts underlying the analyzed methods, as well as key implementation details and design choices. Our report thoroughly discusses similarities and differences among the methods, as well as their respective advantages and disadvantages. 
We establish a consistent framework for comparing the surveyed methods based on key performance metrics and datasets. Specifically, since these methods have been developed in parallel and over a short period of time, currently, no comprehensive comparison exists. This survey, for the first time, presents a unified framework to evaluate 3DGS compression techniques. 
To facilitate the continuous monitoring of emerging methodologies, we maintain a dedicated website that will be regularly updated with new techniques and revisions of existing findings: \\
\url{https://w-m.github.io/3dgs-compression-survey/}. 
\\
Overall, this STAR provides an intuitive starting point for researchers interested in exploring the rapidly growing field of 3DGS compression. By comprehensively categorizing and evaluating existing compression and compaction strategies, our work advances the understanding and practical application of 3DGS in computationally constrained environments. 
\end{abstract}  
\end{@twocolumnfalse}
]

\newpage
\newpage

\section{Introduction}

Computer graphics is a constantly evolving field, working towards more realistic and detailed representations of the world. Key milestones in this journey include the introduction of raster graphics in the 1960s, the development of ray tracing in the 1980s to enhance realism, and the adoption of real-time GPU rendering in the 2000s, which transformed gaming and interactive applications. Realistic 3D world representation remains a crucial and ambitious goal, often considered the "holy grail" for the future of graphics technology.

One way to represent the real world is to build it from the ground up with computer graphics methods, while another is to scan and reconstruct real-world scenes. The computer vision community has made significant strides with Neural Radiance Fields (NeRFs)\cite{mildenhall2020nerf}, a novel approach to scene representation that leverages neural networks to represent volumetric scenes by predicting the color and density at each point in space. NeRFs set a new quality standard for rendering, significantly boosting research in 3D rendering and novel view synthesis. However, implicit models like NeRFs come with inherent challenges. They are computationally intensive because the network must be evaluated billions of times for a single image by querying along rays for each pixel. This makes them difficult to manipulate and complicates direct editing of scene elements. Even with advances such as Instant-NGP\cite{muller2022instant}, which provide much improved training and rendering speeds for radiance fields, the representation remains implicit, limiting its flexibility for practical applications.

3D Gaussian Splatting (3DGS)\cite{kerbl3Dgaussians} has emerged as an explicit alternative to NeRFs, which delivers quality close to the state-of-the-art NeRF approach of Zip-NeRF\cite{barron2023zip}, while allowing explicit control over scene elements. 3DGS uses Gaussian Splatting to represent a scene by distributing a collection of Gaussian ellipsoids throughout the volume. These Gaussians approximate surfaces and volumes, providing an efficient means of rendering and allowing for more straightforward scene manipulation and editing. 3DGS builds on earlier computer graphics work on splatting techniques \cite{splatting91}, such as EWA splatting\cite{zwicker2002ewa}, which laid the foundation for point-based rendering methods. A key advantage of 3DGS is its ability to handle complex visual effects, such as view-dependent lighting and transparency, which are vital for achieving accurate and visually convincing representations of real-world scenes. As an explicit representation involving Gaussian ellipsoids, 3DGS is editable and can be composed with other representations, such as textured meshes, which are widely used across industries like gaming and film due to their flexibility, editability, and compatibility with existing rendering engines. By supporting features like transparency and view-dependent effects, 3DGS enhances the versatility of 3D scene recording.

Moreover, 3DGS supports real-time rendering on high-end GPUs, creating an optimistic outlook for broader availability of applications on mobile and VR devices in the near future. Its simplicity also contributes to its accessibility: the 3DGS format uses a simple renderer with multiple open implementations and a basic .ply file format\cite{turk1994ply}, which can be read and written by libraries or implementations in many programming languages, making 3DGS accessible from the day it was published. This simplicity has driven rapid adoption across various fields\cite{wu2024recent}, making 3DGS an attractive option for both researchers and practitioners. However, early implementations of 3DGS had significant file size challenges \cite{papantonakis2024reducing}. Storing all attributes with full precision often resulted in files of several gigabytes, which drove a wave of research focused on compression techniques. Compression of 3DGS files provides several benefits: it allows for faster transmission over slower connections, enables rendering on lower-end devices by reducing memory requirements, and supports the creation of larger, more complex scenes, particularly in video game development. There is thus a strong motivation for compression, leading to an ongoing race to discover the most effective techniques that balance realism and computational efficiency.

The goal of compact scene representation is to achieve an optimal balance between visual realism and efficient storage. While NeRFs perform compression as an integral part of their neural network-based representation, 3DGS requires explicit densification - adding more Gaussian primitives to fill in details - and pruning - removing redundant primitives to optimize memory usage and performance. Despite these challenges, the explicit nature of 3DGS makes it particularly effective for recording real-world scenes, especially with its ability to capture subtle lighting effects and transparency.

This survey provides an overview of the different techniques developed to achieve a compact 3DGS representation, summarizing numerous parallel efforts and evaluating their effectiveness. By examining these methods, the survey guides future research and applications, helping the community identify successful approaches and areas that require further exploration. Many of the techniques used for compressing 3DGS are adaptations of classical methods, tailored for Gaussian splatting, while others introduce entirely new approaches to reduce data size.

The community has been closing the gap in compression efficiency between 3DGS and NeRFs, which may currently serve as an optimal benchmark. By providing a comprehensive overview, we aim to help researchers identify effective strategies and avoid pitfalls, facilitating the development of improved 3DGS compression methods. Establishing a simple and consistent standard for compressed 3DGS will promote broader adoption in the computer graphics community, enhancing its versatility for diverse applications. Such a standard will streamline compatibility and usability, paving the way for 3DGS to achieve the ubiquity and practicality of textured meshes across various use cases.

\subsection*{Scope of this Survey}

In this state-of-the-art report, we focus on optimization techniques for 3D Gaussian Splatting (3DGS) representations, aiming to optimize memory usage while preserving visual quality and real-time rendering speed. We focus on compression and compaction methods. We provide a comprehensive comparison of various compression techniques, with quantitative results for the most commonly used datasets summarized in a tabulated format. Specifically, we aim to ensure transparency and create a basis for reproducibility of the included approaches. Additionally, we offer a brief explanation of each pipeline and compare and discuss the main compression and compaction approaches. Rather than covering all existing 3DGS methods and applications, our focus is specifically on techniques to optimize 3DGS representations for size or memory footprint; for a broader overview of 3DGS methods and applications, we refer readers to \cite{wu2024recent,fei20243d}. While we include common approaches shared between neural radiance field (NeRF)\cite{mildenhall2020nerf} compression and 3DGS compression, we direct readers to \cite{li2023compressing,chen2024far} for NeRF-specific compression methods. Furthermore, we acknowledge the inclusion of both peer-reviewed publications from conferences and journals, as well as preprints on arXiv, within our state-of-the-art report, recognizing the rapidly evolving nature of the field of 3D Gaussian splatting.

\section{Fundamentals of 3D Gaussian Splatting (3DGS)}\label{sec:3dgs}

3D Gaussian Splatting\cite{kerbl3Dgaussians} introduces a novel approach to real-time radiance field rendering, achieving state-of-the-art performance quality and rendering speed. The technique involves depicting a scene as an ensemble of 3D Gaussian primitives, which are optimized to align with the scene's geometry and visual characteristics. Each 3D Gaussian is characterized by its 59 attributes:
\begin{itemize}
    \item \textbf{Position ($\mu$):} A 3D vector representing the x, y, and z coordinates in world space; the mean of the Gaussian. (3 attributes)
    \item \textbf{Covariance matrix ($\Sigma$):} The covariance matrix, representing orientation and size of the Gaussian primitive, can be factorized into the scaling $S$ along the x, y, and z axes (3 attributes) and rotation $R$, a quaternion, which has 4 components. An exponential activation is used on the scaling parameters. (7 attributes)
    \item \textbf{Opacity ($o$):} A single scalar value. A sigmoid activation is used to constrain the values to range $(0, 1)$. (1 attribute)
    \item \textbf{Color ($c$):} View-dependent color is represented by Spherical Harmonics (SH) coefficients. The first degree of spherical harmonics is required to give splats a diffuse color, split into 3 channels (R, G, B). These are denoted $f\_dc\_\{0, 1, 2\}$ in the .ply point cloud format. Additional view-dependent colors are supported through higher degrees of SH. 3DGS proposes to use degree 3, thus 15 additional coefficients per color channel. (3 rgb x 16 coefficients = 48 attributes)
\end{itemize}

The training of a static 3D Gaussian Splatting scene utilizes a collection of images as input, in conjunction with calibrated cameras derived from  Structure from Motion (SfM), which yield a sparse initial point cloud. For each point within the resulting sparse cloud, a 3D Gaussian distribution ($G(x)$) is initialized. 
The rendering of 3D Gaussian primitives, also called splatting, is achieved by their projection from a three-dimensional space onto a two-dimensional image plane. Each three-dimensional Gaussian is transformed into a two-dimensional Gaussian (a splat) whose footprint is derived from its covariance matrix and the camera's view transformation parameters. For each pixel, colors are aggregated via alpha-blending, with contributions from each splat blended according to their depth order. Ultimately, the pixel's color, denoted as I, is determined as follows: 
\begin{equation}\label{eq:color}
    I(x) = \sum_{i \in N} \alpha_i(x) c_i  \prod_{j=1}^{i-1} (1 - \alpha_j(x)), 
\end{equation} 
\begin{equation*}
    \hspace{2em}  \alpha(x)=oG(x), \hspace{2em} G(x) = e^{-\frac{1}{2} (\mathbf{x-\mu})^T \Sigma'^{-1} (x-\mu)}
\end{equation*}
In this context, $N$ denotes the set of depth-sorted splats that overlap at a pixel, $c_i$ is the color and $\alpha_i$ is the contribution of a primitive, that is the product of opacity ($o$) and Gaussian falloff. This ensures that splats are blended in the appropriate sequence, with those situated closer exerting a more significant impact on the resulting pixel color. Throughout the training or optimization phase, the position, size (as represented by the covariance matrix), opacity, and color of the Gaussian functions are systematically refined to optimally correspond to the input views. Differentiable rendering is used to compute gradients, which allows the adjustment of Gaussian parameters so that the rendered images align with the training images.

\begin{figure}[htb]
    \centering
    \includegraphics[width=\linewidth]{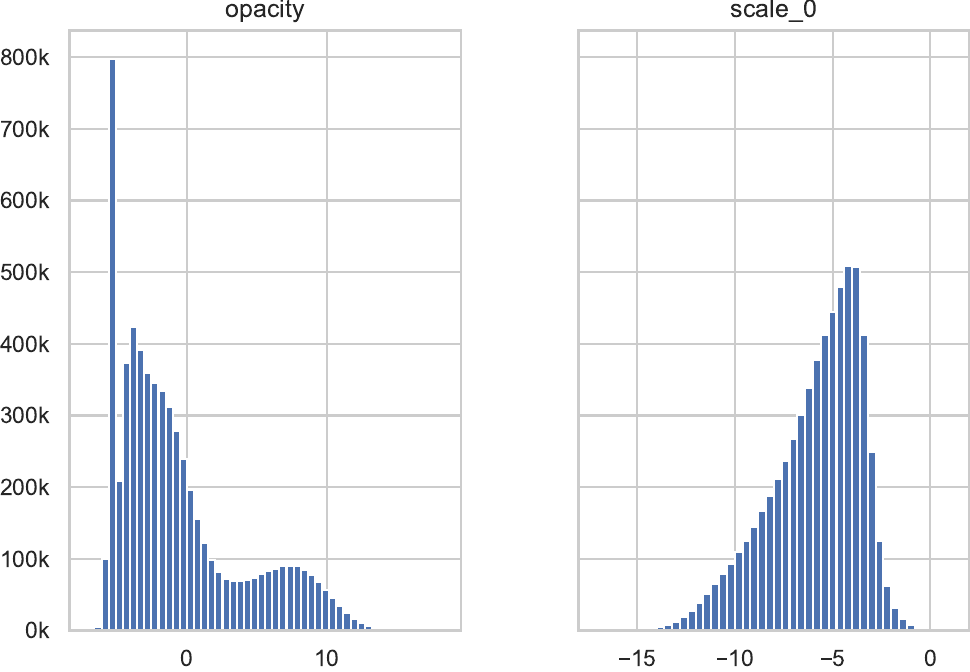}
    \caption{Histograms of the opacity and the first scaling attributes for all 3D Gaussians of the \textit{bicycle} scene (as trained by 3DGS\cite{kerbl3Dgaussians}).}
    \label{fig:opacity_dc0_scale_hist}
\end{figure}
Figure~\ref{fig:opacity_dc0_scale_hist} shows histograms of the opacity and the first scaling attributes for all 3D Gaussians of the \textit{bicycle} scene (as trained by 3DGS\cite{kerbl3Dgaussians}). Before passing the values to the renderer, their activation functions are applied: sigmoid for the opacity, and exponential for the scale. We can see a peak of very low opacity Gaussians, demonstrating potential candidates to be removed from the scene (see Figure \ref{fig:opacity}). In the scaling histogram, we can see that there are very few large Gaussians, an effect from applying the adaptive density control. Additional histograms for the same scene are shown in the Appendix~\ref{ap:stats}.

\section{Fundamentals of Compression and Compaction for 3DGS}

While Gaussian Splatting scenes can efficiently be rendered in real-time, they come with significant file size demands, necessitating efficient compression for managing large scenes. For other media data like images, video, or audio different coding strategies exist to significantly reduce data size. Usually, coding methods are distinguished between lossless and lossy coding. For lossless coding, redundancy is reduced by exploiting the different probabilities of the individual symbols. Such entropy encoding can be achieved, e.g., with Huffman \cite{huffcoding} or arithmetic coding \cite{arithcoding}. For correlated sources, coding efficiency can be further increased by appropriate prediction or transformation of symbols prior to entropy coding. However, depending on source statistics, lossless coding is often restricted to moderate compression around a factor of 2. Much higher coding efficiency can be expected when tolerating small deviations in the decoded data. Lossy compression targets at removing irrelevance in the data that cannot perceived by humans but would need additional bits for encoding. Quantization of the predicted or transformed symbols to fewer code words is a standard approach, either for scalar values or an entire vector as in vector quantization (see Sec.\ \ref{sec:vq}). The goal is then to adjust and shift the quantization error such that is does not become visible. 

\begin{figure}[htb]
    \centering
    \includegraphics[width =\linewidth]{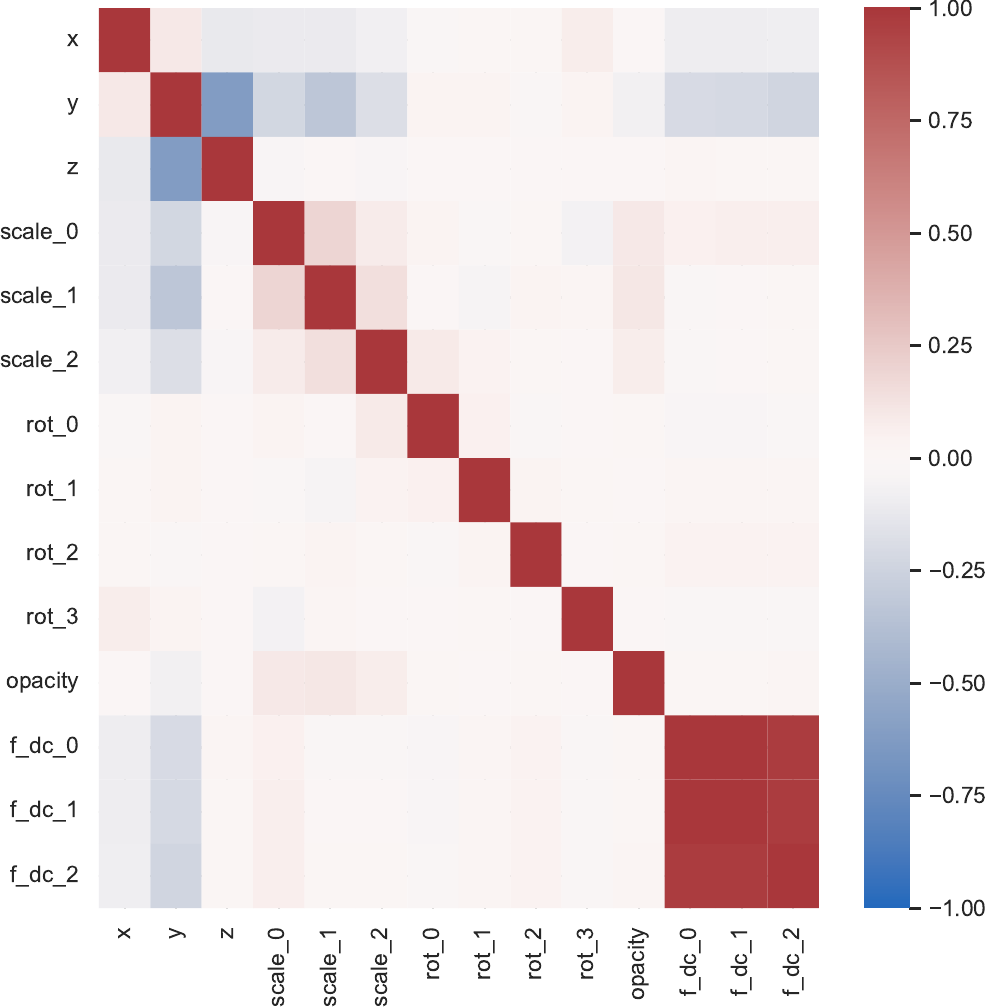}
    \caption{
    A correlation heatmap for attributes of all 3D Gaussians of the \textit{bicycle} scene (as trained by 3DGS\cite{kerbl3Dgaussians}).
    }
    \label{fig:relevant_cols_corr}
\end{figure}
As an example, Figure~\ref{fig:relevant_cols_corr} shows a correlation heatmap for attributes of all 3D Gaussians of the \textit{bicycle} scene, as provided by 3DGS\cite{kerbl3Dgaussians}. The color channels of the base colors of the splats ($f\_dc$) are nearly perfectly correlated, likely due to the dominance of luminance over chrominance changes in this natural scene. This significant correlation suggests the feasibility of jointly encoding the base color channels. Further correlation exists between the individual scale attributes and also between scale and opacity, which can be exploited in compression. A full correlation map including all spherical harmonics attributes can be found in the Appendix~\ref{ap:stats}

Beyond standard coding techniques, Gaussian Splatting provides numerous opportunities to modify data to support more efficient coding. The reason lies in the different ambiguities within the Gaussian Splatting structure. Different arrangements of splats can lead to the same visual appearance while differing in their compressibility. This can be exploited in considering additional losses during training and scene optimization to limit bit-rate while keeping visual quality. Reparameterizing the 3DGS representation offers another avenue for achieving compression without compromising visual quality. In the following sections, we will first present several concepts for compression and compaction of Gaussian Splatting scenes, while Section \ref{sec:strategies} will dive into details of individual approaches from the current state of the art.

\subsection{Vector Quantization}
\label{sec:vq}
\begin{figure}[htb]
    \centering
    \includegraphics[width=0.9\linewidth]{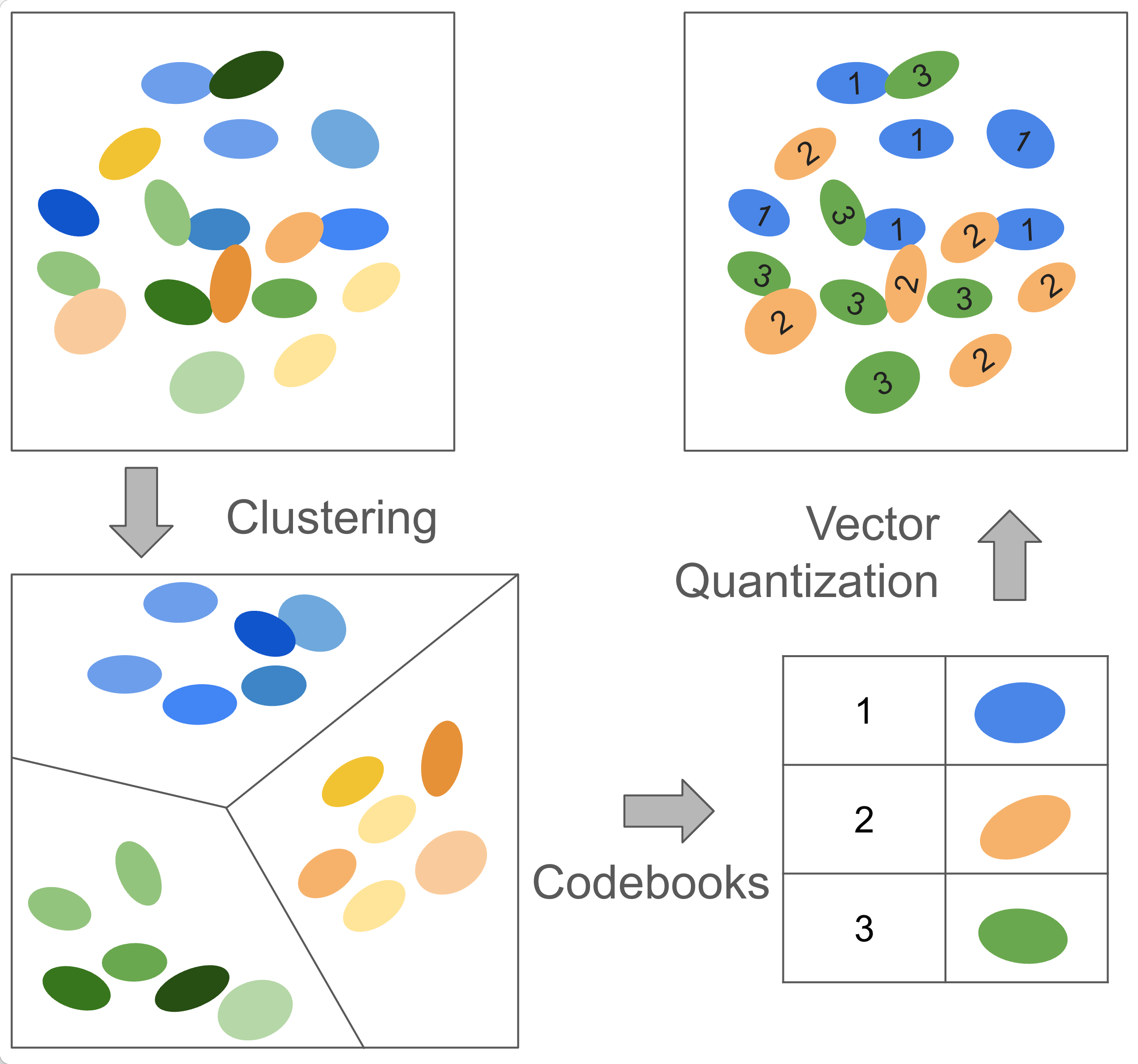}
    \caption{Vector Quantization steps.}
    \label{fig:VQ}
\end{figure}

Vector quantization techniques are central to many compression strategies, with the goal to reduce data complexity by grouping similar data points together and representing them with a shared approximation. Specifically, the original high dimensional dataset is partitioned into clusters, with each cluster being approximated by a representative feature. The process relies on algorithms like K-means \cite{lloyd1982kmean,macqueen1967kmean} or LBG \cite{linde1980lbg}, iteratively assigning data points to clusters, minimizing the distance between the vectors and their assigned centroids. This leads to the formation of a codebook. Data compression is achieved by replacing each original vector with the index of the closest representative vector in the codebook. Figure \ref{fig:VQ} illustrates the sequential process starting from unorganized Gaussian attributes, progressing through the clustering, the creation of a codebook, and the final result of Vector Quantization. The quality of the quantized representation heavily depends on the codebook design and the clustering algorithm used. A well-designed codebook will minimize the error introduced by quantization while maximizing compression efficiency. 

The quantization process can either be performed on the entire multi-dimensional vectors simultaneously or individual dimensions of the data space can be treated as separate quantization tasks. The latter case allows for more flexible handling of different data attributes, especially for applications where different properties of the data require different levels of precision and have different distributions and levels of redundancy. 
In essence, vector quantization algorithms improve a small set of vectors to represent a larger set of vectors under some optimization criterion. This results in significant compression gains, making VQ highly efficient when the data exhibits redundancy or for data-rich applications where a slight loss of precision is acceptable in exchange for dramatic reductions in data size. Hence, these approaches play a crucial role in various domains, from signal processing and image compression to the emerging field of 3D data representation to compress complex datasets while maintaining a close approximation to the original. Especially, it is highly effective for compressing 3D Gaussian splatting (3DGS) data, where attributes like positions, colors, or spherical harmonics are often highly redundant and exhibit natural clustering tendencies. 

\subsection{Structuring and Dimensionality Reduction} 

While vector quantization focuses on reducing redundancy by encoding attributes into shared clusters, an alternative perspective integrates structural organization, spatial redundancy exploitation, and dimensional reduction. Instead of encoding attributes in isolation, methods like octrees, hash-grids, and self-organizing grids reshape and recombine data spatially. These techniques organize scene elements hierarchically or around representative points, enabling efficient reuse of similar properties across regions. By structuring Gaussians to exploit contextual relationships—such as proximity, color, or shape—compression becomes more about the global arrangement of data, reducing the need to store individual attributes independently. This shift from attribute-focused encoding to structural compression highlights the potential for compact representations through spatial coherence, redundancy management, and transformations into lower-dimensional forms that preserve essential spatial relationships. In this section, we look at different methods that provide either a re-structuring of the data, a dimensionality reduction or factorization, as well as combinations thereof.

\subsubsection{Octrees}
\label{sec:octrees}
\begin{figure}[htb]
    \centering
    \includegraphics[width=0.98\linewidth]{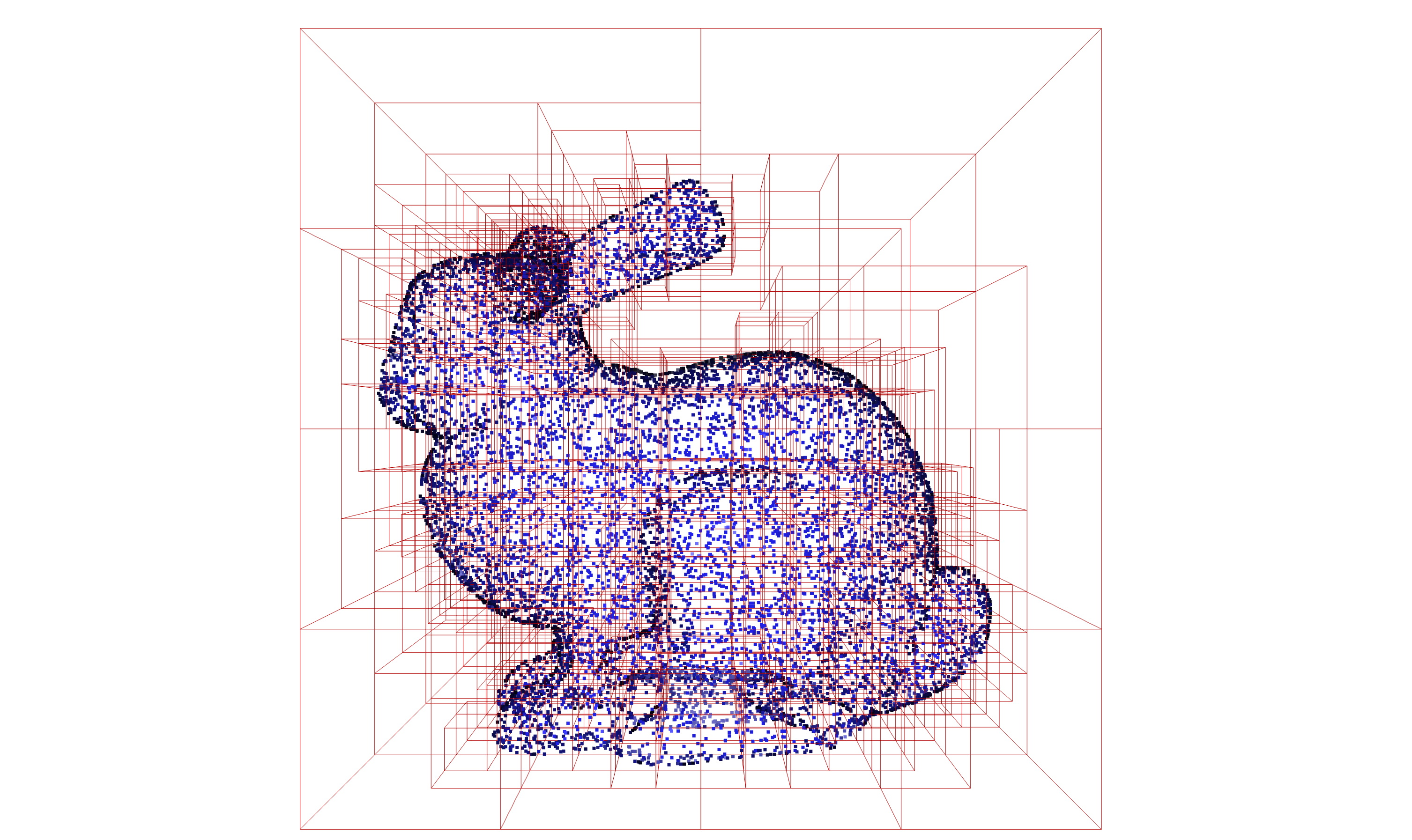}
    \caption{Example of a point cloud with Octree partitioning.}
    \label{fig:octrees}
\end{figure}

Octrees are a spatial partitioning technique used in computer graphics to efficiently represent 3D data like point clouds and volumetric models~\cite{meagher1982geometric}. They have been successfully applied to point cloud compression~\cite{schnabel2006octree}. Furthermore, octrees have demonstrated their effectiveness in speeding up Neural Radiance Field (NeRF) rendering through hierarchical volumetric representation, enabling real-time performance~\cite{yu2021plenoctrees}. Since 3DGS scenes are effectively point clouds with additional attributes, using octrees for 3DGS becomes straightforward. Octrees recursively subdivide 3D space into smaller cubes (see Figure \ref{fig:octrees}), allowing efficient memory allocation by focusing on non-empty regions. In addition, coordinates can be referenced relative to sub-cubes leading to a smaller average bit-length. In 3DGS, octrees help allocate memory only to occupied parts of the scene, skipping over empty regions and, reducing memory usage while preserving detail, making them ideal for large-scale scenes.

\subsubsection{Anchor-based Representations}
\begin{figure}[htb]
    \centering
    \includegraphics[width=0.98\linewidth]{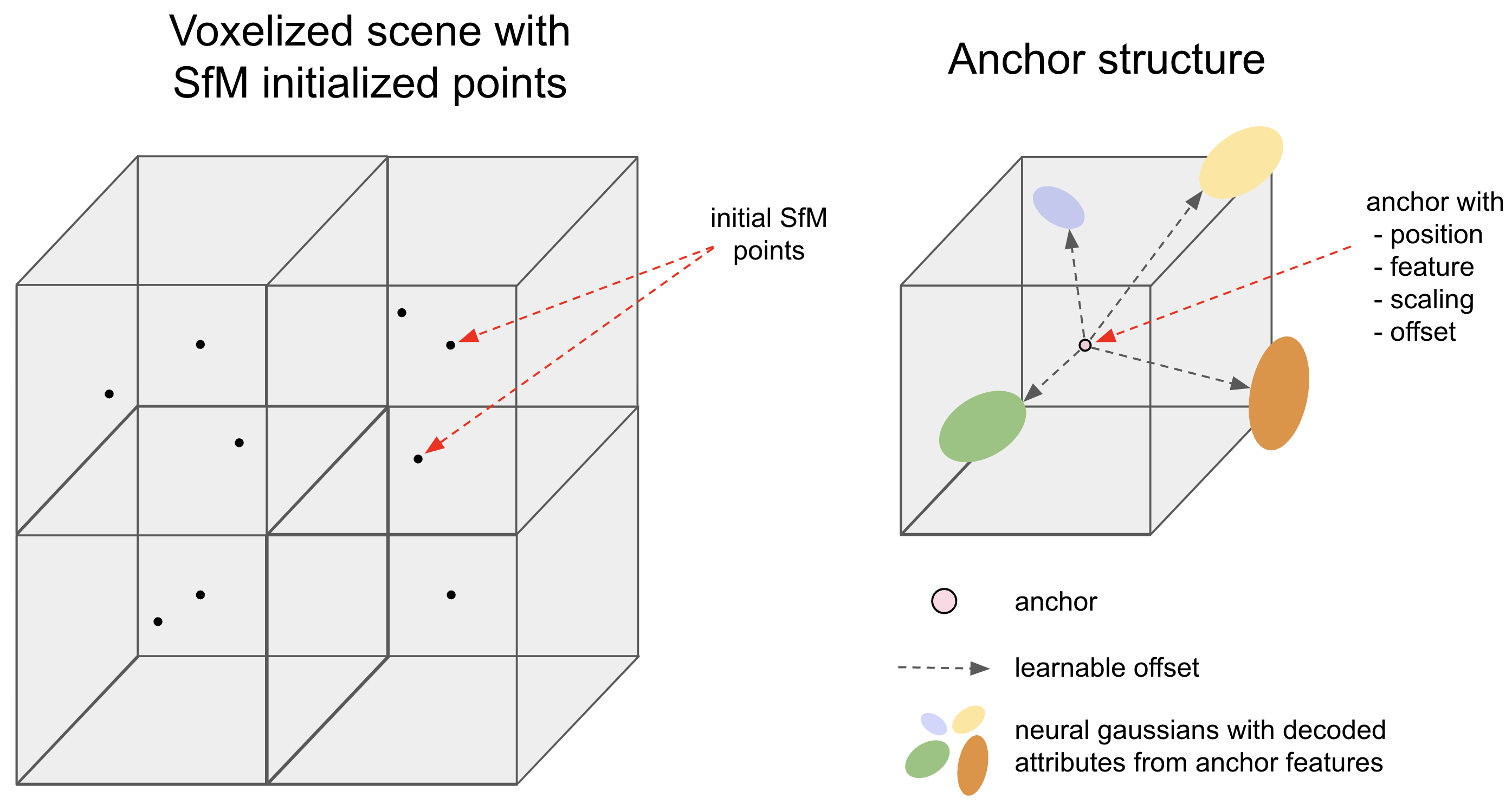}
    \caption{Anchor-based structure. Left: A voxalized scene with SfM inistialized points. Right: The center of each voxel becomes an anchor and is associated with position, feature, scaling and offset. From each anchor neural Gaussians are spawned.}
    \label{fig:anchors}
\end{figure}

Anchor points can be used as proxies to predict the properties of associated Gaussian kernels. Unlike octrees, which partition space into a hierarchical grid, anchors group Gaussians by associating them with representative points, allowing efficient compression without strict spatial subdivision~\cite{lu2024scaffold}. Anchors are initialized by voxelizing the 3D scene and are assigned position, context features, scaling, and learnable offsets. By deriving Gaussian attributes from these anchors rather than storing them individually, redundancy is reduced, leading to decreased memory usage (Fig. \ref{fig:anchors}). This approach is conceptually related to the codebooks used in vector quantization presented in Section \ref{sec:vq}, where shared representations efficiently reduce storage requirements while retaining high fidelity. 

\subsubsection{Multi-Resolution Hash Grids}
Hash-grid assisted context modeling leverages multi-resolution hash grids, inspired by Instant-NGP~\cite{muller2022instant}, to efficiently represent spatial relationships in 3D Gaussian Splatting. Instant-NGP's multiresolution hash encoding uses a compact hash table to store trainable feature vectors, which are optimized during training to represent complex spatial details with fewer parameters compared to dense grid encodings. This method enables a trade off between memory, performance, and quality by adjusting parameters such as hash table size, feature vector size, and the number of hash levels. Unlike traditional spatial encodings that explicitly manage collisions, the hash-grid approach uses an MLP to implicitly handle these collisions. This makes this a hybrid explicit/implicit method, and introduces a small computational overhead for the feature decoding. For application in 3DGS, the features stored in the hash table are decoded into attributes for individual primitives. Hash grids are particularly effective in allocating memory to regions of high importance, as demonstrated by Instant-NGP, which achieves small, high-quality radiance fields with minimal parameters. This makes them a very effective tool for 3DGS compression.

\subsubsection{Z-order Curves}
\begin{figure}[htb]
    \centering
    \includegraphics[width=0.85\linewidth]{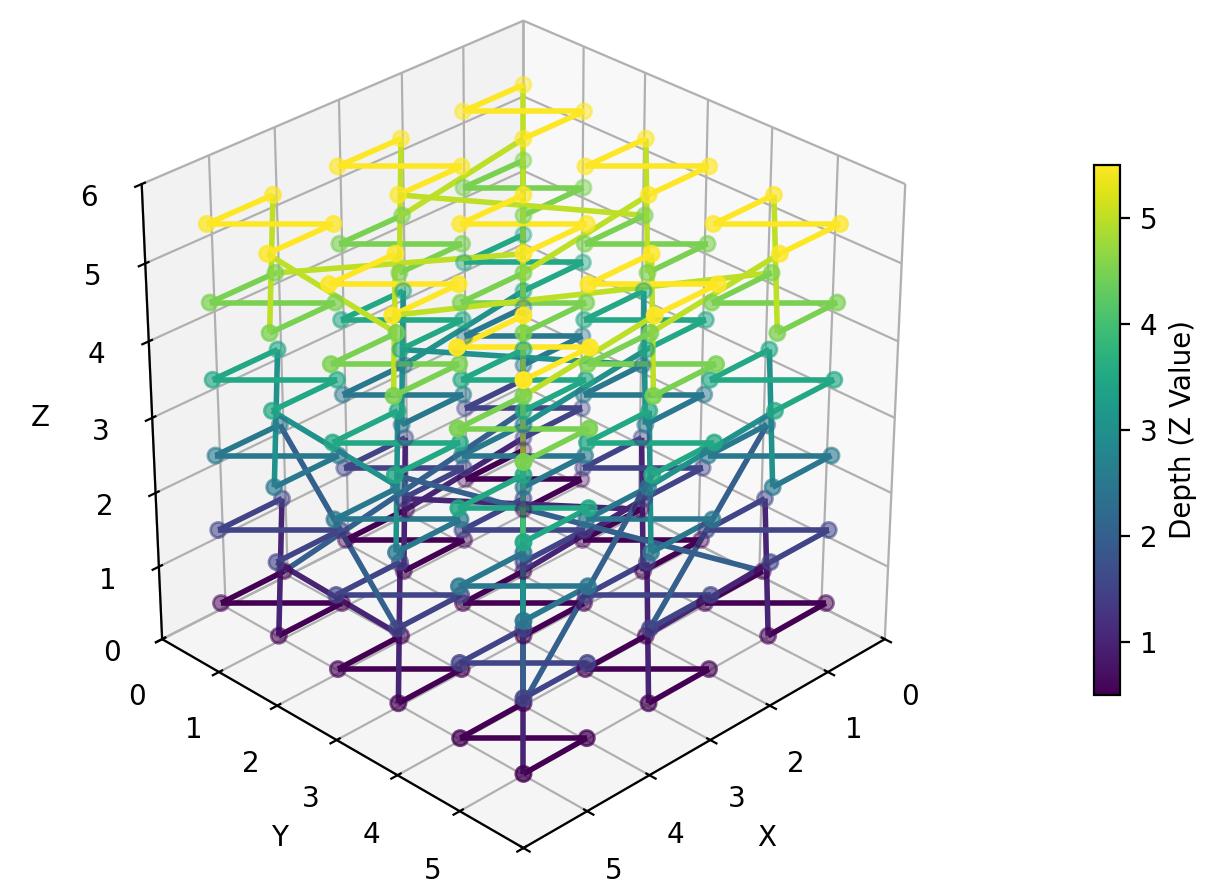}
    \caption{Example of a regular 3D point grid traversed with Z-order curves.}
    \label{fig:z-order-curves}
\end{figure}

Z-order curves, or Morton ordering, work by interleaving the bits of the coordinate values from multiple dimensions to create a single one-dimensional index, effectively mapping multidimensional data into a linear sequence while preserving spatial locality. Figure \ref{fig:z-order-curves} exemplifies how a Z-order curve traverses a regular point grind in 3D space. By ordering the Gaussians according to their positions along a Z-order curve, spatial coherence can be exploited to improve the efficiency of run-length or predictive encoding techniques. This makes Z-order curves useful for spatial indexing and creating a coherent order for splats. In the context of sparse point clouds in 3DGS, Z-order curves are used to order splats based on their positions, which is memory efficient as it avoids allocating storage to empty regions. However, this approach has limitations in terms of maintaining neighborhood relationships: high-dimensional neighbors are not always close when mapped linearly, which affects the efficiency of operations that rely on true spatial proximity. More sophisticated methods, like Hilbert curves, can sometimes provide better locality for such datasets\cite{chen2022hilbert}, offering improved efficiency in spatial indexing by ensuring that spatial neighbors are more likely to remain neighbors in the linear mapping.

\subsubsection{Tri-planes and K-planes}
\begin{figure}[htb]
    \centering
    \includegraphics[width=0.65\linewidth]{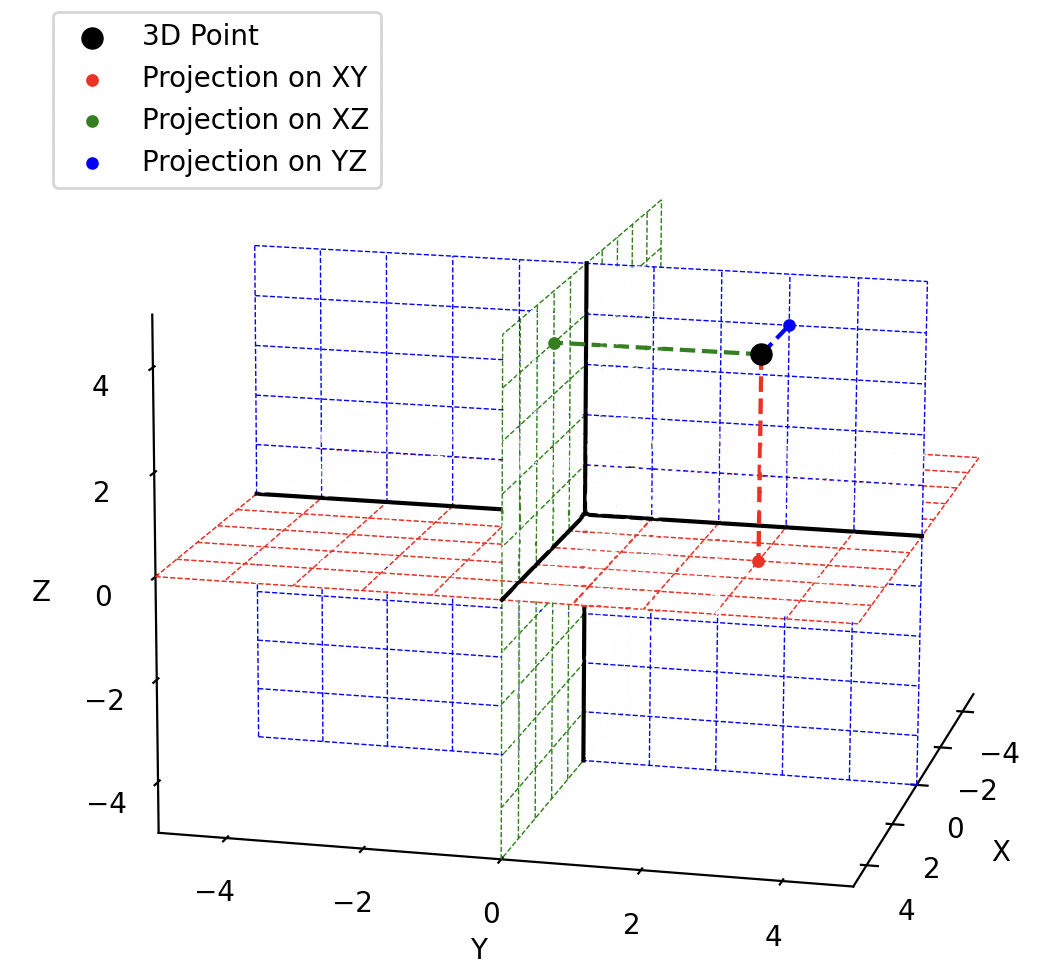}
    \caption{The tri-plane model: A 3D query point is projected onto three orthogonal planes: xy, xz and yz. For 3DGS compression, the query can be the position of the primitive. The planes can store features, which are decoded into the other attributes (e.g. color, opacity) with a small MLP.}
    \label{fig:tri-planes}
\end{figure}

Building on the idea of mapping multidimensional data into lower-dimensional grids, tri-plane factorization instead projects 3D positions onto three orthogonal 2D planes to store and retrieve attributes \cite{chan2022efficient}. This creates a natural 2D organization, where Gaussians that are spatially close in 3D tend to occupy proximate coordinates on each plane, allowing 2D feature maps to be saved or edited like images while preserving some local neighborhood relationships. Rather than storing each primitive’s parameters individually, tri-plane factorization encodes attributes (e.g., color, opacity) as 2D feature images in the xy, xz, and yz planes. This process is visualized in Figure \ref{fig:tri-planes}. The features are combined via simple operations such as concatenation, summation or multiplication and then decoded through a small MLP. Thus, the method is a hybrid of explicit storage (the planes) and implicit representation (the decoder).

Compared to Z-order curves, which provide a single linear mapping of 3D points to 1D indices, tri-planes rely on three parallel 2D projections for more direct spatial correlation in each axis-aligned view. This makes it straightforward to compress or synthesize the plane images with standard 2D techniques. It also allows for synthesis of the feature planes with generative frameworks such as StyleGAN\cite{karras2019style}. Furthermore, extension to higher-dimensional spaces yields K-Planes \cite{fridovich2023k}, where temporal or additional dimensions can be incorporated by factorizing them into extra 2D planes.

As with other dimensionality reduction methods, key parameters include the overall resolution of each plane (trading off granularity versus memory), how features from the three planes are aggregated, and the complexity of the MLP decoder that maps these aggregated features to the final per-Gaussian attributes. Similar to the Z-order approach, the aim is to exploit spatial coherence so that splats sharing similar properties lie close in the 2D feature maps. Because tri-plane factorization scales as \(\mathcal{O}(N^2)\) instead of \(\mathcal{O}(N^3)\), higher resolutions can be used than in dense voxel grids. This makes tri-plane factorization useful tool for 3DGS compression.

\subsubsection{Self-Organizing Gaussians}
\begin{figure}[htb]
    \centering
    \includegraphics[width=0.65\linewidth]{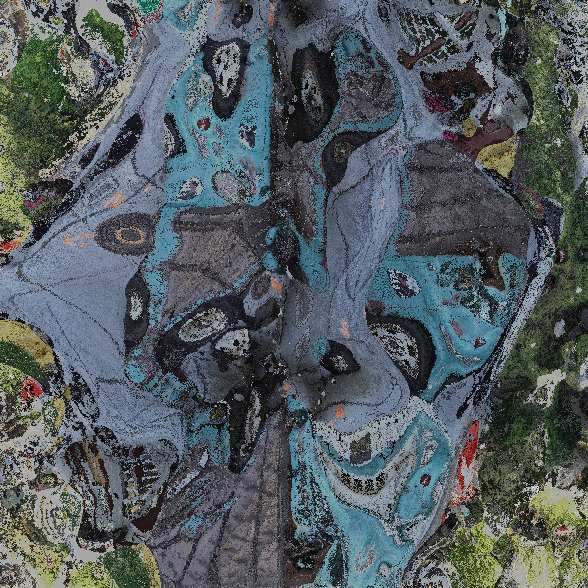}
    \caption{The Gaussian primitives of the \textit{Truck} scene mapped into a 2D layout using the \textit{Self-Organizing Gaussians} approach\cite{morgenstern2024compact}. An optimal arrangement is learned which preserves local relationships. This learned organization can utilize the available grid space more efficiently than fixed methods like Z-curves tri-planes.}
    \label{fig:self-organizing-gaussians}
\end{figure}

Another approach to map high-dimensional Gaussian parameters into 2D grids is through self-organization. This method was developed specifically for 3DGS compression\cite{morgenstern2024compact}. Figure \ref{fig:self-organizing-gaussians} depicts an example of the primitives of the \textit{Truck} scene from the \TandT dataset reorganized into a 2D grid. The idea is based on the concept of Self-Organizing Maps\cite{kohonen1990self}, an unsupervised learning model that projects high-dimensional data onto a lower-dimensional grid, preserving the topological relationships between data points through competitive learning. Similar in motivation to Z-order curves and tri-planes, this 2D representation enables the use of standard image compression techniques, exploiting perceptual redundancies and ensuring local smoothness between neighboring splats. Unlike Z-order curves and tri-planes, which impose a fixed ordering, the Self-Organizing Gaussians representation is optimized for each scene, allowing neighbors in all dimensions to be modeled effectively. By organizing attributes into multiple data layers that share a consistent 2D layout, this technique provides a highly compressible and efficient representation of the original 3D scene.

\subsubsection{Region Adaptive Hierarchical Transform (RAHT)}
RAHT is a bottom-up transform that reorganizes color attributes along an octree (see also Section \ref{sec:octrees} on octrees). At each step, it pairs neighboring voxels according to their weights—the number of underlying voxels—and produces one low-frequency (DC) and one high-frequency coefficient\cite{de2016compression}. The DC coefficients propagate upward for further merging, while the high-frequency coefficients are encoded immediately. Because the transform adapts to local density by adjusting the transform matrix to the weights, it effectively preserves small-scale details while reducing redundancy. Compared to graph-based transforms that require costly eigen-decompositions, RAHT relies on simple pairwise operations at each level, making it computationally efficient and suitable for real-time compression. After applying the transform, the resulting coefficients are quantized and entropy coded using an arithmetic coder with a Laplacian model, with sub-band parameters encoded with a Run-Length Golomb-Rice (RLGR) coder. This approach achieves compression performance comparable to state-of-the-art methods for point-cloud compression, but at significantly lower complexity. This hierarchical scheme is especially effective for attributes that exhibit strong local correlation, thus it can be adapted to encode attributes like color or density for 3D Gaussian splatting.

\subsubsection{Discussion}

The methods discussed form a toolkit to achieve compact and efficient 3D Gaussian Splatting, each with distinct advantages and disadvantages. Z-curves, tri-plane factorizations and 2D grid mappings reduce dimensional complexity in a simple and straightforward way. Fixed schemes like Z-order curves provide efficient ordering but may struggle with spatial coherence in sparse datasets. Adaptive techniques, such as anchor-based grouping or learned self-organizing mappings, can mitigate these issues by dynamically adjusting to scene-specific redundancies. Tri-planes may not use the available space efficiently, as empty space in 3D translates to empty patches in the planes. Self-Organizing Gaussians overcome this by learning a custom, non-linear mapping per scene. Hash-grids use adaptive, multiresolution indexing, which is fast and memory efficient, but is not an invertible mapping. Hash grids and tri-planes both store features, which need to be decoded into attributes with an MLP. While explicit methods allow direct access to Gaussian attributes, this implicit feature storage introduces compactness at the cost of requiring additional decoding.

Octrees adaptively allocate memory only to occupied regions, optimizing resource use for complex scenes, but require advanced attribute compression schemes like RAHT to exploit local correlations for compression. Anchor-based methods group Gaussians around representative points, reducing local redundancy efficiently, similar to vector quantization. While these methods are often considered independently, combining anchor-based representations with structured grids (e.g., hash grids, octrees) can enhance both local and global efficiency. Hybrid approaches allow for flexible spatial partitioning while maintaining fine-grained control over Gaussian placement.

All discussed methods are designed for efficient decoding, making them suitable for real-time applications. However, encoding complexity varies: Self-Organizing Gaussians require a computationally expensive optimization to determine element order, anchor-based methods involve clustering and optimization to define representative points, and feature-based methods like tri-planes and hash grids require a training process to establish features and MLP weights for decoding. This feature decoding adds a small computational overhead at decoding and rendering time.

Memory consumption at decoding and rendering time depends on the representation. For explicit methods, VRAM (GPU memory) usage is directly tied to the number of primitives, meaning that reducing their count through compaction (as discussed in the following sections) directly lowers memory requirements. In contrast, methods like hash grids, tri-planes, and vector quantization can use less memory overall by storing features in compact structures such as feature planes, hash tables, or codebooks. However, their memory usage is constant, meaning that reducing the number of primitives does not further decrease VRAM consumption. Depending on whether a system is memory- or compute-constrained, different trade-offs may be preferable. Large scenes often demand significant VRAM, making compact representations beneficial, while compute-constrained devices, such as VR headsets, may have sufficient memory but benefit from pre-decoding implicit representations into explicit attributes to reduce runtime computation.



\subsection{Attribute Pruning}
Attribute pruning refers to the selective reduction of attributes associated with each Gaussian, such as spherical harmonics (SH), to optimize memory usage and computational efficiency. In 3DGS, spherical harmonics are often used to encode complex lighting and color information. SH coefficients can become expensive to store and process, especially as the degree of precision increases. For example, adopting SH up to degree 3 for RGB data can result in 48 coefficients (16 per channel), consuming $81\%$ (48 out of 59 attributes) of the storage size required for a single Gaussian.

To address this, attribute pruning techniques dynamically adjust the number of SH coefficients for each Gaussian. Rather than applying a uniform level of SH precision across the entire scene, the degree of SH is tailored based on the specific requirements of each Gaussian. In less complex regions of the scene, some Gaussians only need a single degree of SH (basic RGB color), while more complex regions require higher degrees for more detailed representation.

It is also possible to completely forgo spherical harmonics, and model the view-dependent effects with other approaches, such as NeRF-like MLPs, or classical computer graphics shading\cite{jiang2024gaussianshader}. These approaches may use far fewer attributes then third-degree spherical harmonics, but on the downside require custom, and potentially slower, rendering.

\subsection{Compaction}
The key tool for achieving compaction in 3D Gaussian Splatting (3DGS) is Adaptive Density Control (ADC). ADC dynamically manages the number of Gaussians during the optimization process, adjusting their density based on scene requirements rather than relying on a fixed number of elements. It adds or removes Gaussians depending on their contribution to the scene, ensuring that only the most essential elements are retained.

\begin{figure}[htb]
    \centering
    \includegraphics[width=\linewidth]{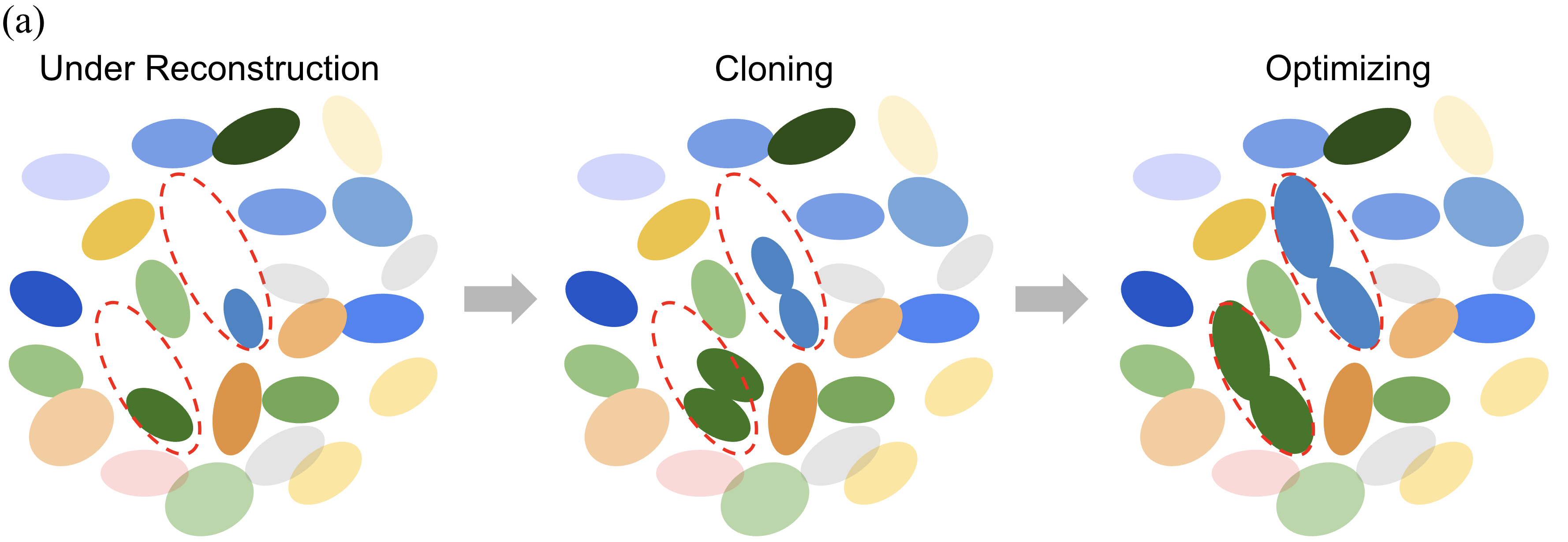}
    \centering
    \includegraphics[width=\linewidth]{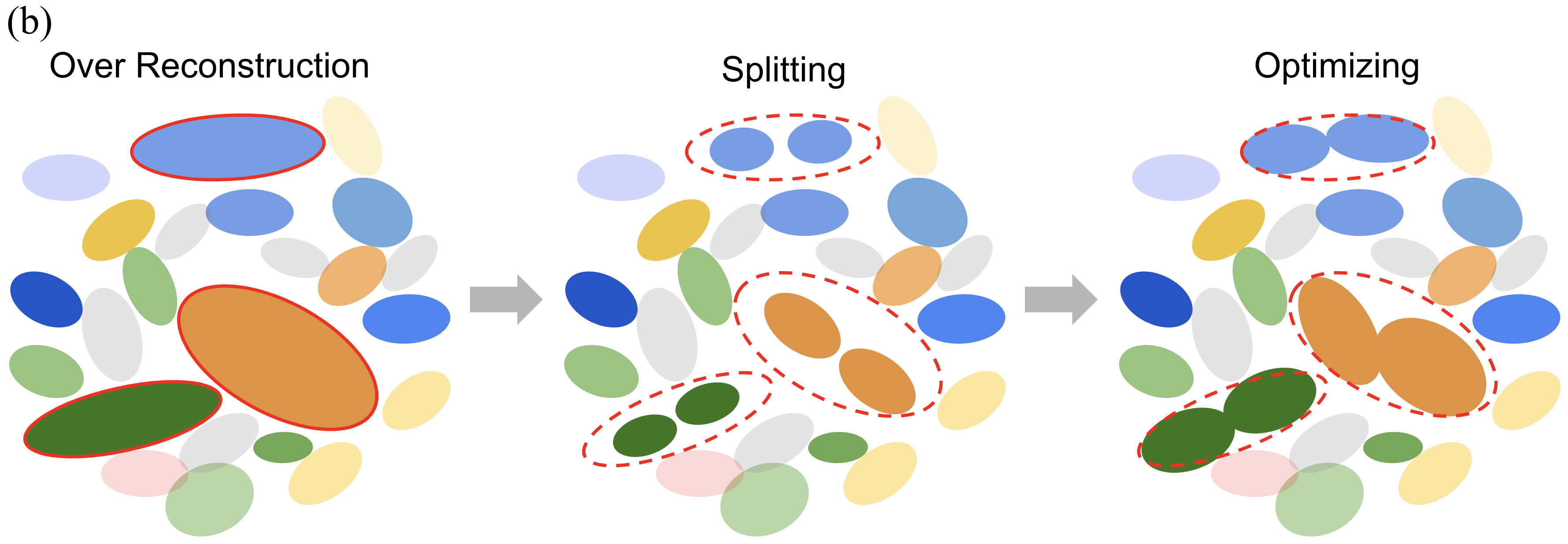}
    \centering
    \includegraphics[width=\linewidth]{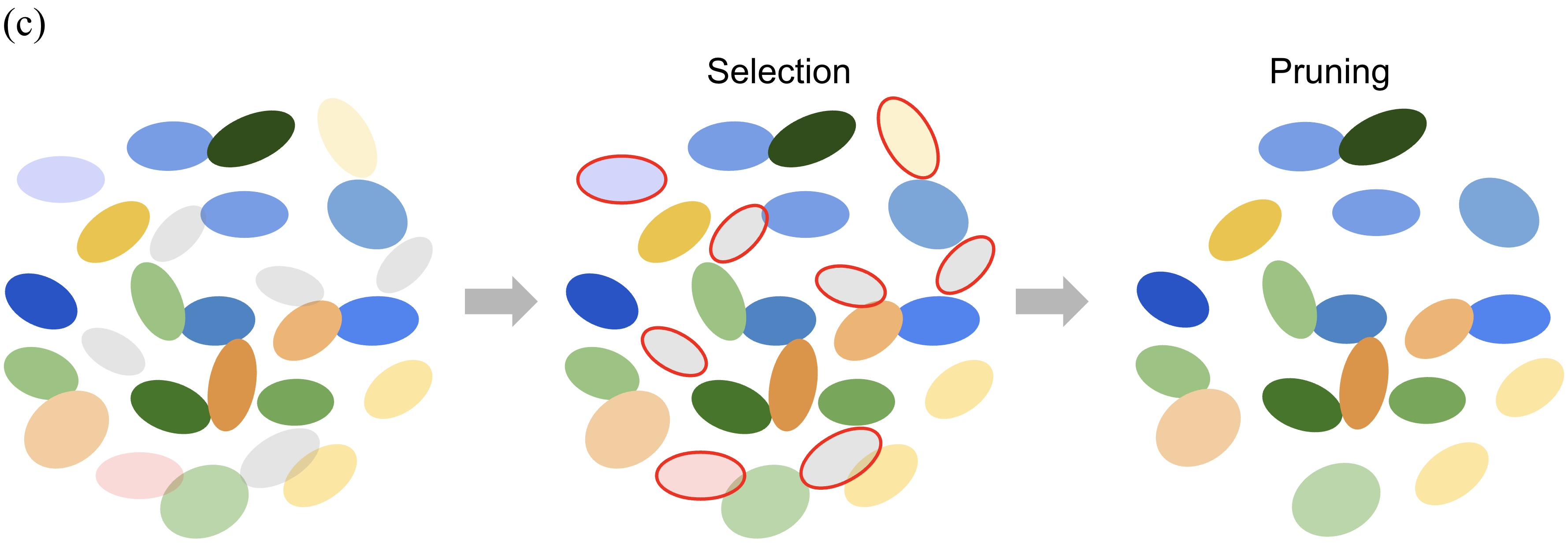}
    \caption{From top to bottom, this figure shows the following processes: (a) Gaussian cloning, (b) Gaussian splitting, and (c) Gaussian pruning.}\label{fig:cloneSplitRemove}
\end{figure}

By assessing criteria such as gradients, pixel coverage, and saliency maps, ADC intelligently determines whether Gaussians should be cloned, split, or removed. This ensures that additional Gaussians are allocated where they are most needed, such as in high-frequency regions, while redundant or less impactful Gaussians are pruned. Figure \ref{fig:cloneSplitRemove} exemplifies the processes for (a) Gaussian primitive cloning, (b) Gaussian primitive splitting, and (c) Gaussian primitive pruning. Cloning produces an identical replica of the chosen Gaussian which is then incorporated into the scene. The splitting algorithm substitutes a Gaussian with a set number (default N = 2) of child Gaussians. These child Gaussians, after splitting, are situated within the bounds of the removed parent Gaussian. The children's scaling is derived from the parent Gaussian's scaling, reduced by a factor. The remaining attributes are directly inherited from the parent Gaussian.

\begin{figure*}[htb]
    \includegraphics[width=\textwidth]{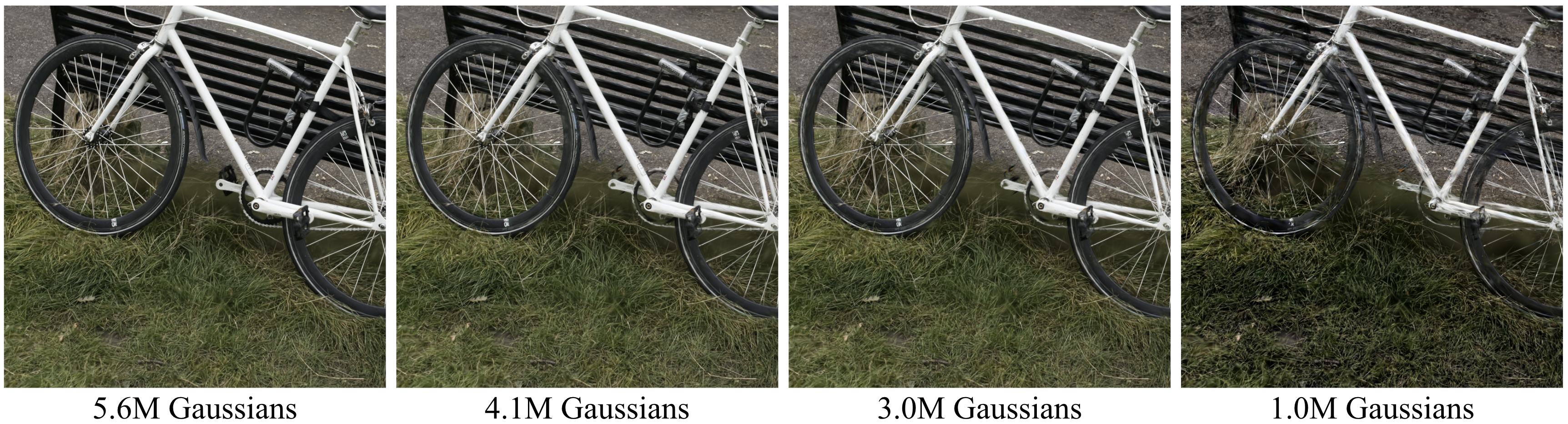}
    \caption{Render of the Gaussian scene \textit{bicycle} from the \MipNeRF{} dataset with Gaussians pruned based on an opacity criterion.}
    \label{fig:opacity}
\end{figure*}

Figure \ref{fig:opacity} shows how the number of Gaussians, in this example filtered by opacity, impact the visual quality of the rendered scene. While there is only minimal visual quality loss between 5.6 and 4.1 Million Gaussians, you can already spot changes in the rendering of the bicycle's spokes. This becomes more evident with 3.0 Million Gaussians. With even less Gaussians (1.0 Million) the bicycle reconstruction becomes transparent and the grass reconstruction also suffers.

Operating dynamically during both training and rendering, ADC continuously refines the scene representation. As the scene evolves, ADC ensures efficient Gaussian allocation, leading to a more compact and high-quality model over time.

\section{Efficient Strategies for Compression and Compaction in 3D Gaussian Splatting}
\label{sec:strategies}

\begin{figure}[htb]
    \centering
    \includegraphics[width=0.9\linewidth]{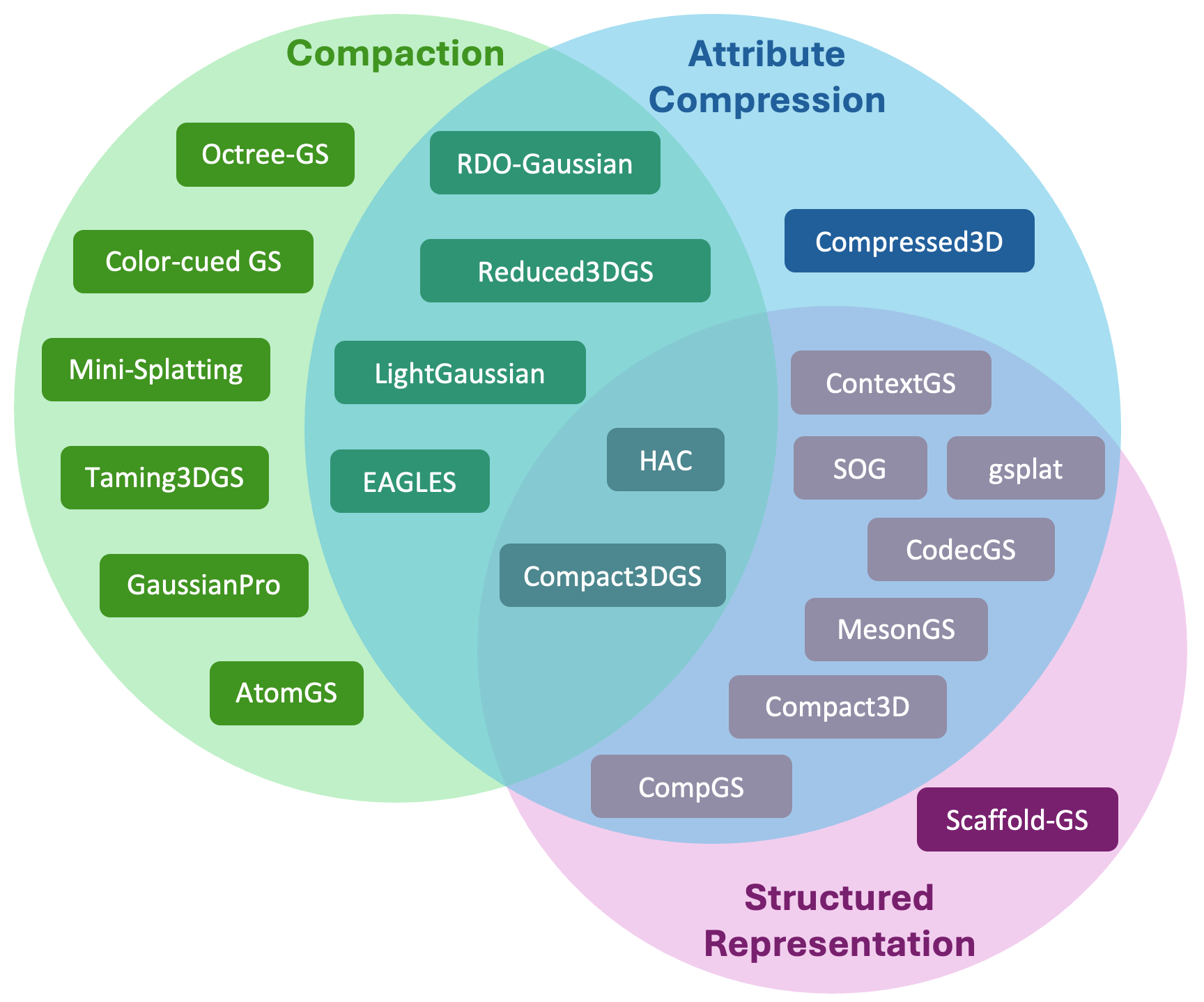}
    \caption{Venn diagram showing how Compression approaches notably Attribute Compression and Structured Representions of Gaussians intesect with Compaction approaches.}
    \label{fig:venn}
\end{figure}

As 3D Gaussian Splatting (3DGS) evolves as a prominent method for real-time scene rendering, its increasing adoption is challenged by substantial storage and computational requirements. This section addresses two essential optimization strategies: Compression (Sec. \ref{sec:compression}) and Compaction (Sec. \ref{sec:compaction}). 
Compression reduces storage usage by employing methods such as vector quantization, which clusters similar Gaussian attributes to reduce redundancy, and structured representations that organize Gaussians into more compact forms like grids or anchor points. Compaction, on the other hand, focuses on optimizing the number and distribution of Gaussians, ensuring that only the most essential elements are retained while reducing unnecessary data. Figure \ref{fig:venn} illustrates how Compression -- attribute compression and structured representations -- and Compaction intersect. Compaction here includes Gaussian pruning and densification approaches. Together, these strategies enhance the efficiency of 3DGS, making it more practical for a wide range of applications and devices.

\subsection{Compression}\label{sec:compression}

The principal objective of compression in 3D Gaussian Splatting is to preserve the high fidelity of the original information while substantially reducing data volume. This objective is realized through different strategies that address the primary factors contributing to significant memory usage. 3DGS scenes are composed of a substantial quantity of Gaussians; hence, the most evident method to minimize the memory footprint is to decrease the number of Gaussians utilized (see Sec.~\ref{pruning}).
When further examining individual Gaussian kernels, the numerous attributes associated with each Gaussian necessitate the consideration of efficient representations for these attributes. Lastly, Gaussians representing a 3D scene are often not randomly distributed but exhibit spatial relationships and patterns. This inherent structure can be leveraged to organize and compress the Gaussian data more efficiently. Most compression methodologies integrate these various approaches. 

In the following sections, we first discuss efficient methodologies for representing Gaussian attributes (Sec. \ref{sec:efficientRepresentionGaussianAttributes}), which is then followed by a perspective on the structured representations of Gaussians (Sec.~ \ref{sec:StructuredRepresentationOfGaussians}).

\subsubsection{Efficient Representation of Gaussian Attributes} \label{sec:efficientRepresentionGaussianAttributes}

3D Gaussian splatting requires significant storage due to the extensive number of Gaussians and their associated attributes. A more efficient representation of these attributes can mitigate storage requirements without appreciable degradation in quality, thereby achieving the objective of compression. 

Based on the assumption that many Gaussians share similar attributes, they can be quantized using Vector Quantization (VQ). Most commonly quantized attributes  are then stored in codebooks based on K-means as in \cite{navaneet2023compact3d,niedermayr2024compressed,papantonakis2024reducing,lee2024compact,fan2024lightgaussian}.  In \textit{LightGaussian}~\cite{fan2024lightgaussian}, the authors employ VQ on Spherical Harmonics in combination with a significance score to omit VQ on Spherical Harmonics with higher significance. Similarly, \cite{niedermayr2024compressed} introduces a sensitivity-parameter which describes the sensitivity of the reconstruction quality to changes of the Gaussian attributes. This sensitivity measure is then used for sensitivity-aware VQ of Gaussian attributes (e.g.~ Spherical Harmonics, shape). 
In \cite{xie2024mesongs}, the authors categorize SH coefficients in degrees greater than 0 as unimportant and thus compressable through VQ. Furthermore for key attributes (opacity, scales, Euler angles, and 0-degree SH coefficients), Xie et al.\ reduce the entropy using RAHT before quantization. \textit{RDO-Gaussian}~\cite{wang2024end} use entropy-constrained VQ with codebooks to quantize covariance and color parameters for a more compact representation. \textit{HAC} \cite{chen2024hac} uses an anchor structure with associated Gaussians and introduces an Adaptive Quantization Module (AQM) designed to dynamically select quantization steps to facilitate entropy coding of the anchor attributes. 
The researchers in \textit{EAGLES} \cite{girish2024eagles} employ a methodology for quantizing attributes such as rotation, view-dependent color, and opacity by leveraging a latent vector associated with each attribute. This latent vector is integrated with a multilayer perceptron (MLP) decoder, which facilitates the decoding of latent representations into attribute values. To ensure differentiability during the training process, an additional latent approximation is preserved. A Straight-Through Estimator (STE) is then utilized to round the latent approximation before the propagation of gradients.

As described in Section \ref{sec:3dgs}, 3D Gaussians are characterized by 59 attributes, with the majority pertaining to color representation. Of these, the Spherical Harmonics (SH) coefficients encompass 48 attributes across three bands of SH. Notably, 45 out of the 48 SH coefficients are responsible for depicting view-dependent colors. To reduce storage, attribute pruning SH coefficients is a common strategy.
In light of this, \cite{papantonakis2024reducing} introduces an adaptive adjustment mechanism for SH. The authors advocate for the calculation of an average transmittance per pixel and per view. In instances where there is minimal variation per view, this allows for the reduction of SH coefficients to lower bands or their complete removal, thereby significantly diminishing the memory footprint associated with each Gaussian primitive. 

\textit{Self-Organizing-Gaussians}\cite{morgenstern2024compact} provide an ablation experiment, showing that training scenes completely without any higher-degree spherical harmonics is possible, still achieving a competitive quality with a much smaller storage size.
Building upon the preliminary concept of reducing the number of SH coefficients, \textit{LightGaussian} \cite{fan2024lightgaussian} proposes a Knowledge Distillation scheme, accompanied by pseudo-view augmentation, to efficiently encapsulate information from higher-order coefficients into a more condensed form. 

It should be noted, that whenever a band or degree of spherical harmonics is removed, the higher frequency components of the color representation are also eliminated, which inherently leads to a loss of fine detail in view-dependent effects. This reduction, while beneficial for storage efficiency, inevitably sacrifices some information that would otherwise capture subtle variations in lighting and shading across different viewpoints.


Another straightforward way to reduce memory is using lower bit-depth representations (e.g., 16-bit half-floats instead of 32-bit floats) for attributes that do not require high precision. In \textit{SOG}\cite{morgenstern2024compact}, the authors clip various attribute ranges, including RGB, opacity, and spherical harmonics, based on percentile thresholds to ensure consistent normalization across models. After clipping, they quantize these attributes by rounding to the nearest value within predefined linear ranges $q$, with $q=2^{14}$ for coordinates and $q=2^6$ for scale, opacity, and rotation and $q=2^5$ for SH.
Fan et al.~\cite{fan2024lightgaussian} quantize selective Spherical Harmonics, position, shape, rotation and opacity attributes. Additionally, packing attributes using a codec like LZ77 as in \textit{MesonGS}~\cite{xie2024mesongs} or JPEG XL as in\cite{morgenstern2024compact} reduces remaining storage size.


\subsubsection{Structured Representation of Gaussians}
\label{sec:StructuredRepresentationOfGaussians}

In the vanilla approach to \textit{3DGS}~\cite{kerbl3Dgaussians}, a final high-fidelity scene usually consists of millions of unordered Gaussians. Besides efficiently representing attributes of single Gaussians by leveraging similarities within attributes as discussed in the previous Sec.~\ref{sec:efficientRepresentionGaussianAttributes}, some compression approaches leverage the correlation between neighboring Gaussians and find structured representations, opening new pathways to compression. 

One approach for structuring 3D Gaussians whithin a scene is the Anchor-Based representation, as introduced in \textit{Scaffold-GS}~\cite{lu2024scaffold}. From the initial Structure from Motion (SfM) derived point cloud a scene is first voxelized and each voxel center is then treated as an anchor. Each anchor is associated with a local context and learnable offsets, acting as representative point  which can spawn new Gaussians in close proximity. Subsequently a neural network predicts the attributes i.e. opacity, color and covariance, of the associated Gaussians based on anchor features and viewing conditions. This approach effectively reduces the overall number of parameters required to represent the scene.
\textit{ContextGS}~\cite{wang2024contextgs} builds upon the Anchor representation introduced in \textit{Scaffold-GS}~\cite{lu2024scaffold} but divides the anchor points into hierarchical levels, from coarse to fine. The anchors are encoded progressively, starting from the coarsest level. The decoded values of the anchors at a coarser level are then used to predict the distribution of nearby anchors at the next finer level, using an MLP. The hierarchical anchor approach efficiently exploits the spatial relationships between anchors.
In \textit{HAC}~\cite{chen2024hac}, the Anchor-Based representation is enhanced through the inclusion of a Hash-Grid Assisted Context. The primary concept involves simultaneously learning a structured, compact hash grid for context modeling of anchor attributes. For each anchor, the anchor's position is used to query the hash grid and retrieve an interpolated hash feature, which subsequently predicts the value distributions of the anchor attributes to aid in entropy coding for a highly efficient representation.
While also based on an Anchor-Based representation with anchor primitives and coupled primitives the primitives differ in their structure \textit{CompGS}~\cite{liu2024compGS}. Anchor primitives serve as reference and contain geometry attributes (location and covariance) and reference embeddings. Coupled primitives only contain residual embeddings to capture deviations. The attributes of coupled primitives are predicted by warping corresponding anchor primitives using affine transforms derived from anchor and coupled primitive embeddings.

In order to leverage the similarities among the color attributes of adjacent Gaussians and to eliminate the need to store attributes for each Gaussian, \textit{Compact3DGS}~\cite{lee2024compact} proposes the use of a hash-grid followed by a small MLP specifically for view-dependent color attributes. Positions are input into the hash-grid, and the resultant feature along with the view-direction are subsequently provided to the MLP in order to retrieve the color.

An alternative structuring methodology is employed by \textit{SOG}~\cite{morgenstern2024compact} and also used in \textit{gsplat}\cite{mcmc}, where unstructured Gaussians are mapped onto a structured 2D grid to spatially organize Gaussians with similar attributes closely, enhancing the smoothness of attribute values. This configuration allows for efficient compression using standard image compression techniques. To further improve smoothness and compressibility, a smoothing regularization term can be added during training to promote locally smooth configurations of Gaussians on the grid.
An other approach \textit{CodecGS}~\cite{lee2025compression3dgaussiansplatting} applies feature planes for efficient attribute representation and uses a standard video coding technique for feature plane compression. The authors propose a tri-plane architecture to predict Gaussian attributes instead of storing them directly. A progressive training strategy allows to capture coarse geometric information initially and gradually add finer details. The feature planes are optimized by applying a block-wise discrete cosine transform (DCT) allowing the model to leverage the spatial correlations within the feature planes as conventional image signals.

The authors in \textit{Compact3D}~\cite{navaneet2023compact3d} suggest an approach for organizing the unstructured 3D Gaussians by sorting them according to one of the quantized indices and subsequently storing them employing Run-Length Encoding (RLE).

In \textit{MesonGS}\cite{xie2024mesongs}, octrees are employed to achieve compression of the geometrical structure, specifically the 3D positions of the 3D Gaussians.

\subsubsection{Quantitative Comparison of Compression Methods}

This section presents the comparative analysis of 3DGS compression methods across four datasets: \TandT, \MipNeRF, \DeepBlending~and, \SyntheticNeRF. More details on the datasets are provided in Sec. \ref{sec:datasets}. For better comparison Table \ref{tab:compression} only includes approaches that mainly focus on compression, we provide a separate Table \ref{tab:compaction} to compare approaches that mainly focus on compaction (i.e. densification and pruning). Furthermore some approaches were missing the necessary data for quantitative comparison, approaches that we consider worth mentioning nevertheless are included in the survey but not in the tables. As shown in Table \ref{tab:compression}, the metrics used to evaluate the performance (see Sec. \ref{sec:metrics}) of each compression method are PSNR, SSIM, LPIPS and model size measured in megabytes. 

The objectives for compressing 3DGS vary depending on the application, with some requiring minimal model size, while others prioritize a smaller size alongside optimal perceptual quality. As there is no definitive winner that excels across all categories, we introduce a simple rank, reflecting the average rankings of the methods across all available datasets, thereby offering general guidance on the overall performance of the approaches. To determine the compression dataset $rank_s$, the ranks for the quality metrics PSNR, SSIM, and LPIPS are equally weight with the model size. Consequently, each quality metric contributes one-sixth to the overall ranking, while the model size accounts for the remaining half: 
\\
$rank_s = \frac{rank(PSNR)}{6} + \frac{rank(SSIM)}{6} + \frac{rank(LPIPS)}{6} + \frac{rank(Size[MB])}{2}$.
\\
The overall method ranking is calculated by averaging the dataset ranks across all available datasets. This approach ensures that methods with incomplete data are fairly included in the overall comparison. 
The \textit{min} operator is applied to resolve ties in metric rankings, assigning the lowest rank available to all methods in the group, while subsequent ranks skip the number of tied methods.

{
\rowcolors{2}{gray!25}{white}
\footnotesize
\setlength{\tabcolsep}{3pt}
\begin{table*}
\caption{Performance comparison of 3DGS compression methods across four datasets: \TandT, \MipNeRF, \DeepBlending and, \SyntheticNeRF. The included metrics are PSNR, SSIM, LPIPS and, model size in MB. The best methods in each category are highlighted (\colorbox{lightred}{first}, \colorbox{lightorange}{second}, \colorbox{lightyellow}{third}). The rank represents the average rankings of the methods across all available datasets. The method highlighted in bold: \textbf{3DGS-30K} is the original 3DGS method.\label{tab:compression}}
\begin{adjustbox}{center}
\begin{tabular}{ll|lllr|lllr|lllr|lllr}
\toprule
Method & Rank & \multicolumn{4}{c|}{Tanks and Temples} & \multicolumn{4}{c|}{Mip-NeRF 360} & \multicolumn{4}{c|}{Deep Blending} & \multicolumn{4}{c}{Synthetic NeRF} \\
 & \tiny  & \tiny PSNR$\uparrow$ & \tiny SSIM$\uparrow$ & \tiny LPIPS$\downarrow$ & \tiny \makecell{Size \\ MB$\downarrow$} & \tiny PSNR$\uparrow$ & \tiny SSIM$\uparrow$ & \tiny LPIPS$\downarrow$ & \tiny \makecell{Size \\ MB$\downarrow$} & \tiny PSNR$\uparrow$ & \tiny SSIM$\uparrow$ & \tiny LPIPS$\downarrow$ & \tiny \makecell{Size \\ MB$\downarrow$} & \tiny PSNR$\uparrow$ & \tiny SSIM$\uparrow$ & \tiny LPIPS$\downarrow$ & \tiny \makecell{Size \\ MB$\downarrow$} \\
\midrule
ContextGS\_lowrate & \cellcolor{lightred}4.3 & \cellcolor{lightyellow}24.12 & .849 & .186 & 9.9 & \cellcolor{lightyellow}27.62 & .808 & .237 & \cellcolor{lightorange}13.3 & 30.09 & \cellcolor{lightyellow}.907 & .265 & \cellcolor{lightred}3.7 &  &  &  &  \\
HAC-highrate & \cellcolor{lightorange}4.4 & \cellcolor{lightred}24.40 & \cellcolor{lightyellow}.853 & \cellcolor{lightyellow}.177 & 11.8 & \cellcolor{lightred}27.77 & \cellcolor{lightorange}.811 & .230 & 22.9 & \cellcolor{lightorange}30.34 & .906 & .258 & 6.7 & \cellcolor{lightred}33.71 & \cellcolor{lightred}.968 & \cellcolor{lightorange}.034 & \cellcolor{lightorange}2.0 \\
CodecGS & \cellcolor{lightyellow}4.8 & 23.63 & .841 & .192 & \cellcolor{lightred}7.8 & 27.30 & \cellcolor{lightyellow}.810 & .236 & \cellcolor{lightred}10.3 & 29.81 & .906 & .251 & 9.0 &  &  &  &  \\
HAC-lowrate & 5.0 & 24.04 & .846 & .187 & \cellcolor{lightorange}8.5 & 27.53 & .807 & .238 & \cellcolor{lightyellow}16.0 & 29.98 & .902 & .269 & \cellcolor{lightorange}4.6 & 33.24 & \cellcolor{lightorange}.967 & .037 & \cellcolor{lightred}1.2 \\
gsplat-1.00M & 5.3 & 24.03 & \cellcolor{lightred}.857 & \cellcolor{lightred}.163 & 16.1 & 27.29 & \cellcolor{lightorange}.811 & \cellcolor{lightyellow}.229 & \cellcolor{lightyellow}16.0 &  &  &  &  &  &  &  &  \\
ContextGS\_highrate & 5.8 & \cellcolor{lightorange}24.29 & \cellcolor{lightorange}.855 & \cellcolor{lightorange}.176 & 12.4 & \cellcolor{lightorange}27.75 & \cellcolor{lightorange}.811 & .231 & 19.3 & \cellcolor{lightred}30.41 & \cellcolor{lightorange}.909 & .259 & 6.9 &  &  &  &  \\
Compact3D 32K & 9.3 & 23.44 & .838 & .198 & 13.0 & 27.12 & .806 & .240 & 19.0 & 29.90 & \cellcolor{lightyellow}.907 & .251 & 13.0 &  &  &  &  \\
Compact3D 16K & 9.7 & 23.39 & .836 & .200 & 12.0 & 27.03 & .804 & .243 & 18.0 & 29.90 & .906 & .252 & 12.0 &  &  &  &  \\
RDO-Gaussian & 9.8 & 23.34 & .835 & .195 & 12.0 & 27.05 & .802 & .239 & 23.5 & 29.63 & .902 & .252 & 18.0 & 33.12 & \cellcolor{lightorange}.967 & \cellcolor{lightyellow}.035 & \cellcolor{lightyellow}2.3 \\
CompGS & 9.8 & 23.70 & .837 & .208 & 10.1 & 27.26 & .803 & .239 & 17.3 & 29.69 & .901 & .279 & 9.2 &  &  &  &  \\
Reduced3DGS & 10.4 & 23.57 & .840 & .188 & 14.0 & 27.10 & .809 & \cellcolor{lightorange}.226 & 29.0 & 29.63 & .902 & \cellcolor{lightorange}.249 & 18.0 &  &  &  &  \\
SOG w/o SH & 10.5 & 23.15 & .828 & .198 & \cellcolor{lightyellow}9.3 & 26.56 & .791 & .241 & 16.7 & 29.12 & .892 & .270 & \cellcolor{lightyellow}5.7 & 31.37 & .959 & .043 & \cellcolor{lightorange}2.0 \\
MesonGS c3 & 12.0 & 23.29 & .835 & .197 & 17.4 & 26.99 & .797 & .246 & 25.9 & 29.48 & .903 & .252 & 29.0 & 32.96 & \cellcolor{lightred}.968 & \cellcolor{lightred}.033 & 3.5 \\
Compressed3D & 12.4 & 23.32 & .832 & .194 & 17.3 & 26.98 & .801 & .238 & 28.8 & 29.38 & .898 & .253 & 25.3 & 32.94 & \cellcolor{lightorange}.967 & \cellcolor{lightred}.033 & 3.7 \\
MesonGS c1 & 12.6 & 23.31 & .835 & .196 & 18.5 & 26.99 & .796 & .247 & 28.5 & 29.50 & .903 & .251 & 31.1 & 32.94 & \cellcolor{lightred}.968 & \cellcolor{lightred}.033 & 3.9 \\
SOG & 12.7 & 23.56 & .837 & .186 & 22.8 & 27.08 & .799 & .230 & 40.3 & 29.26 & .894 & .268 & 17.7 & 33.23 & \cellcolor{lightyellow}.966 & \cellcolor{lightorange}.034 & 4.1 \\
Compact3DGS+PP & 13.3 & 23.32 & .831 & .202 & 20.9 & 27.03 & .797 & .247 & 29.1 & 29.73 & .900 & .258 & 23.8 & 32.88 & \cellcolor{lightred}.968 & \cellcolor{lightorange}.034 & 2.8 \\
EAGLES & 14.4 & 23.37 & .84 & .20 & 29.0 & 27.23 & \cellcolor{lightyellow}.81 & .24 & 54.0 & 29.86 & \cellcolor{lightred}.91 & \cellcolor{lightyellow}.25 & 52.0 &  &  &  &  \\
Scaffold-GS & 14.6 & 23.96 & \cellcolor{lightyellow}.853 & \cellcolor{lightyellow}.177 & 87.0 & 27.50 & .806 & .252 & 156.0 & \cellcolor{lightyellow}30.21 & .906 & .254 & 66.0 &  &  &  &  \\
Compact3DGS & 14.9 & 23.32 & .831 & .201 & 39.4 & 27.08 & .798 & .247 & 48.8 & 29.79 & .901 & .258 & 43.2 & \cellcolor{lightorange}33.33 & \cellcolor{lightred}.968 & \cellcolor{lightorange}.034 & 5.8 \\
LightGaussian & 15.3 & 23.11 & .817 & .231 & 22.0 & 27.28 & .805 & .243 & 42.0 &  &  &  &  & 32.72 & .965 & .037 & 7.8 \\
\textbf{3DGS-30K} & 15.3 & 23.14 & .841 & .183 & 411.0 & 27.21 & \cellcolor{lightred}.815 & \cellcolor{lightred}.214 & 734.0 & 29.41 & .903 & \cellcolor{lightred}.243 & 676.0 & \cellcolor{lightyellow}33.32 &  &  &  \\
EAGLES-Small & 17.2 & 23.10 & .82 & .22 & 19.0 & 26.94 & .80 & .25 & 47.0 & 29.92 & .90 & \cellcolor{lightyellow}.25 & 33.0 &  &  &  &  \\
\bottomrule
\end{tabular}
\end{adjustbox}
\end{table*}
}

The methods evaluated show significant variation in file size, with some achieving high compression rates at the cost of visual quality, while others strike a balance between compression and maintaining higher fidelity.
While our proposed ranking puts \textit{ContextGS-lowrate}~\cite{wang2024contextgs} on the first place, high variations between datasets and quality metrics show that there is not one winning compression strategy. Depending on the application and goals, different approaches should be considered. 
When size is the main concern \textit{CodecGS}\cite{lee2025compression3dgaussiansplatting} stands out for its very small file size (e.g., just 7.8 MB on \TandT), all while maintaining PSNR/SSIM near the top tier. Furthermore, \textit{HAC-lowrate}~\cite{chen2024hac}, \textit{SOG w/o SH}~\cite{morgenstern2024compact} and \textit{ContextGS\_lowrate}~\cite{wang2024contextgs} are the best-performing compression methods, achieving file sizes under 10 MB on the \TandT~dataset with acceptable levels of visual quality.

When the emphasis is on perceptual quality (LPIPS) \textit{gsplat}~\cite{Ye_gsplat} proves to be an excellent choice, alongside \textit{ContextGS\_highrate}~\cite{wang2024contextgs} and \textit{HAC\_highrate}~\cite{chen2024hac}. It is noteworthy that the original \textit{3DGS-30K}~\cite{kerbl3Dgaussians} exhibits the best LPIPS values on both the \MipNeRF~ and the \DeepBlending~ dataset, at the cost of being up to 70 times larger in size. 

Taken together, the top-ranking methods (\textit{ContextGS}~\cite{wang2024contextgs}, \textit{HAC}~\cite{chen2024hac}, \textit{CodecGS}~\cite{lee2025compression3dgaussiansplatting}, \textit{gsplat}~\cite{Ye_gsplat}) achieve a balanced compromise among PSNR, SSIM, LPIPS, and size, with slight differences depending on whether absolute fidelity or smaller storage is the priority. The prevailing conclusion is that methods integrating context modeling (\textit{ContextGS}~\cite{wang2024contextgs}) or hierarchical approaches (\textit{HAC}~\cite{chen2024hac}) tend to dominate in terms of image quality, whereas \textit{CodecGS} and the "lowrate" variants are particularly robust for model compression.

\begin{figure*}[htb]
    \hspace{1cm}
    \includegraphics[width=0.4\textwidth]{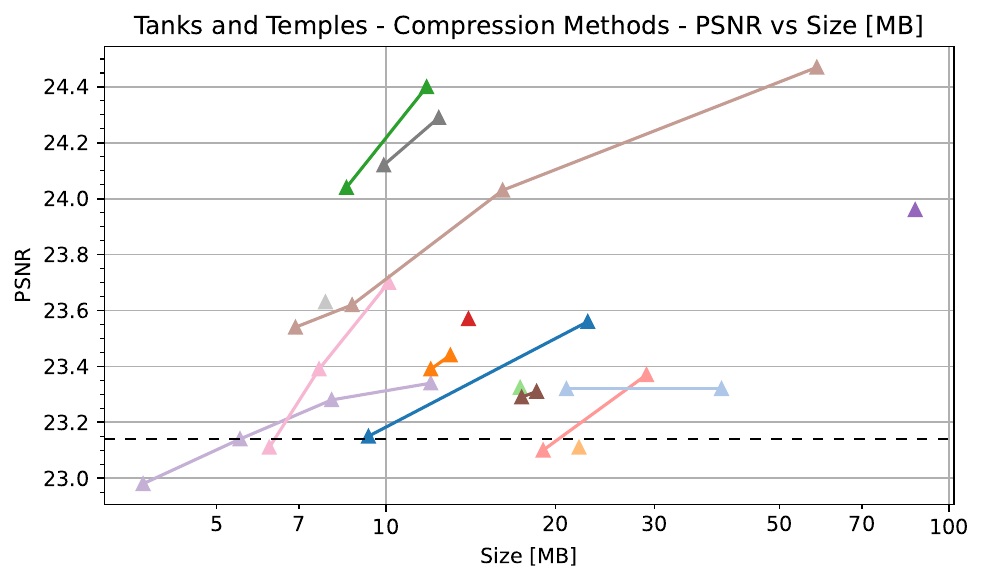}
    \includegraphics[width=0.497\textwidth]{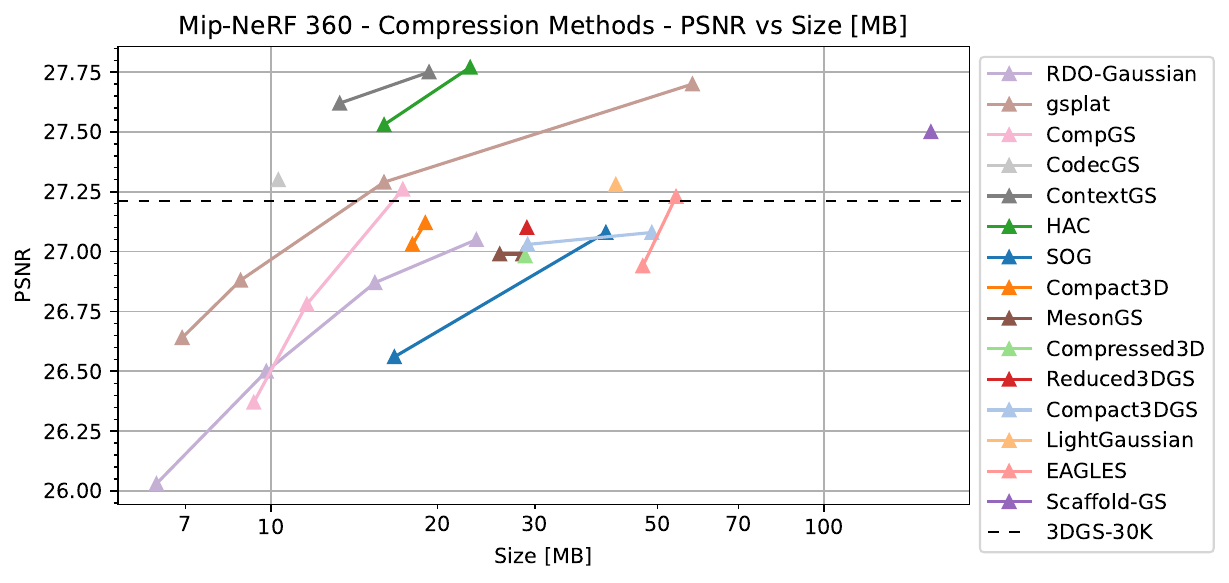}

    \hspace{1cm}
    \includegraphics[width=0.4\textwidth]{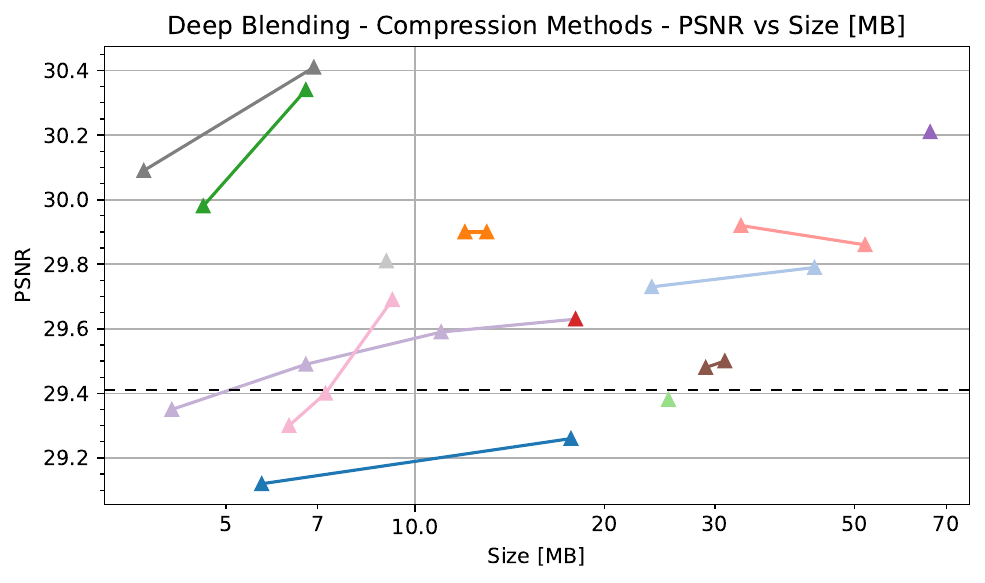}
    \includegraphics[width=0.4\textwidth]{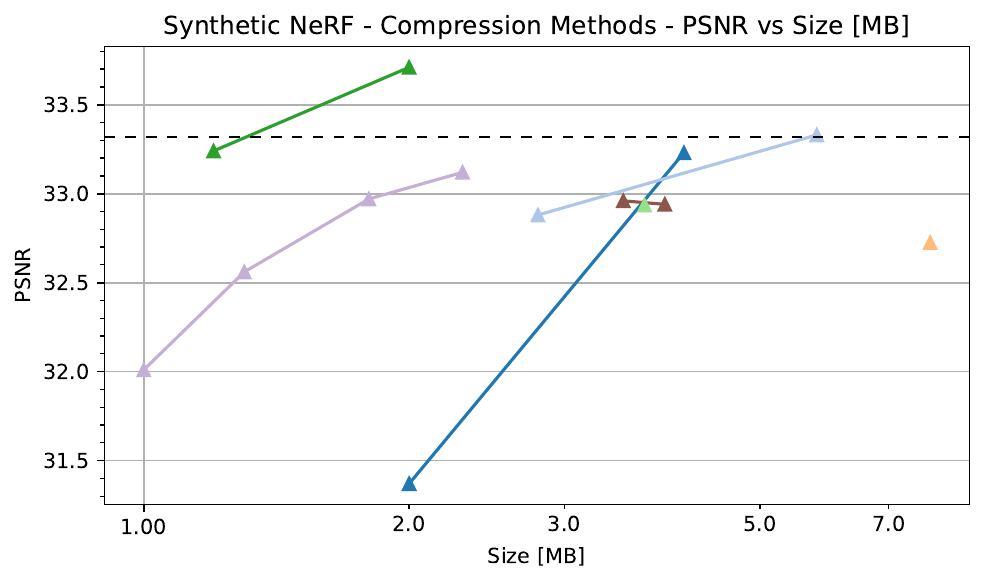}
    \caption{PSNR vs. Model Size (MB) for 3D Gaussian Splatting Compression Methods. The graphs compare different 3DGS compression methods across the \TandT{}, \MipNeRF{}, \DeepBlending{}, and \SyntheticNeRF{} datasets. The x-axis represents the model size (in MB), while the y-axis represents the PSNR, indicating the visual quality.}\label{fig:graphs}
\end{figure*}

Figure \ref{fig:graphs} illustrates the trade-offs between model size and performance measured by PSNR for various 3DGS compression methods across different datasets.  The curves highlight how smaller file sizes typically result in lower PSNR, showing the balance between compression efficiency and visual quality. Some approaches have additional data points, which for clarity were not included in the table. Further figures for the performance comparison measured by SSIM and LPIPS as well as the corresponding compaction figures can be found in Appendix~\ref{ap:compressionCompaction}.

\subsection{Compaction}\label{sec:compaction}
Compaction in 3D Gaussian Splatting (3DGS) refers to the optimization of Gaussian kernel distribution in 3D space to accurately represent scene features while maintaining computational efficiency. The initial kernel set often struggles to capture complex details, especially in high-frequency regions or geometrically intricate areas. Compaction in 3D Gaussian Splatting (3DGS) leverages Adaptive Density Control (ADC) to dynamically manage the distribution and density of Gaussians. ADC can be broadly categorized into two key approaches: Densification and Pruning. This classification is based on the primary objectives of each method: Densification techniques focus on selectively adding Gaussians where they are most needed to improve scene fidelity, while Pruning techniques focus on removing Gaussians that do not contribute effectively to the scene, avoiding over-reconstruction and inefficiencies. We first introduce densification in Sec. \ref{densification} and pruning in Sec. \ref{pruning}, outlining how each process contributes to optimizing the representation.

While most of the following methods do not specifically intend to reduce the memory footprint of 3DGS scenes, they implicitly do so, as they improve the quality of the scene without using more Gaussian primitives. Consequently, at similar quality, these methods require less Gaussians and therefore less memory. Note that some methods mentioned in this section are not included in Table \ref{tab:compaction} due to incomplete evaluations, particularly in the number of Gaussians, which excludes them from the comparison.

\subsubsection{Densification} \label{densification}
These methods use different criteria to determine where and how to introduce new Gaussians. For instance, the \textit{Color-cued Efficient Densification} method~\cite{Kim_2024_CVPR} leverages view-independent spherical harmonics coefficients to assess color cues, refining areas where traditional structure-from-motion techniques may struggle to capture fine details. \textit{FreGS}~\cite{zhang2024fregs} addresses over-reconstruction by regularizing frequency discrepancies in rendered images, focusing on the frequency domain. Meanwhile, \textit{Pixel-GS}~\cite{zhang2024pixelgs} introduces a pixel-aware gradient, targeting under-reconstruction artifacts by incorporating cues from multiple views, thus emphasizing pixel-level information in the densification decision-making process. Similarly, \textit{Revising Densification in Gaussian Splatting (RDGS)}~\cite{Bul2024RevisingDI} adopts the criteria that determines whether a Gaussian should be cloned or split during the optimization. Instead of using the accumulated positional gradient of each Gaussian, RDGS uses a structural similarity function to address a loss for each individual Gaussian. If this loss is high enough, the Gaussian will be split during the regular densification intervals.

\textit{GaussianPro} \cite{cheng2024gaussianpro} employs a different approach by using depth and normal maps to guide the growth and adjustment of Gaussians. It utilizes patch matching \cite{barnes2009patchmatch} to propagate depth and normal information from neighboring pixels and applies geometric filtering and selection to identify pixels requiring additional Gaussians. \textit{MVG-Splatting} \cite{li2024mvg} and \textit{Mini-Splatting} \cite{Fang2024MiniSplattingRS} also employs a depth map to enforce geometric consistency, but with a more targeted application. Specifically, MVG-Splatting applies the geometric consistency only in the near and far regions of the scene to effectively mitigate under-reconstruction in these critical regions while avoiding the risk of over-reconstruction. Mini-Splatting incorporates depth information after a few optimization iterations. It relies on the initial training process to estimate depth, requiring a few iterations to obtain this information. This approach ensures geometric accuracy while still optimizing the overall scene representation. \textit{Taming 3DGS} \cite{taming20243dgs} employs a global scoring approach to guide the addition of Gaussians, ensuring efficient densification. The global score consists of 1) gradient, 2) pixel coverage, 3) per-view saliency, and 4) core attributes like opacity, depth, and scale. By calculating this score that reflects both the scene’s structural complexity and visual importance, only the most critical areas are targeted for Gaussian splitting or cloning, resulting in more effective scene representation. \textit{AtomGS} \cite{liu2024atomgs} takes a different approach by employing a density control process similar to the original 3DGS, relying on positional (scaling) gradients. However, it enhances performance through the addition of edge-aware loss, guiding the gradient to better align with the scene's geometry. While it does not explicitly control densification with varying criteria, its refinement in gradient guidance leads to improved density control outcomes.

Some densification approaches adopt a multi-level strategy for greater flexibility. By down sampling or up sampling images, this approach generates multiple Gaussian sets, each finely tuned to different resolution. For instance, \textit{Octree-GS} \cite{ren2024octree} organizes Gaussians according to levels of detail using an octree structure \cite{bai2023dynamic}. It selectively trains Gaussians based on different views, adaptively adjusting the levels for training to ensure optimal rendering based on the observer's perspective.

Lastly, \textit{Markov Chain Monte Carlo (MCMC)} \cite{mcmc} uses a completely different approach for densifying the 3DGS scene. Instead of cloning and splitting the individual Gaussians, they sample a fixed number of Gaussians from a learned probability distribution. Specifically, they add a noise term to the position of the Gaussians to incorporate exploration during the training, while they also re-spawn Gaussians that drop below an opacity threshold randomly at Gaussians with high opacity. This way, the method improves the quality of regions with few initial Gaussian primitives and the number of Gaussians can directly be controlled.

\subsubsection{Pruning} \label{pruning}
These techniques focus on identifying and removing redundant Gaussians that contribute minimally to the scene representation, such as those that are excessively large, transparent, or provide overlapping information. \textit{Compact3DGS} \cite{lee2024compact}, \textit{RDO-Gaussian} \cite{wang2024end}, and \textit{HAC} \cite{chen2024hac} introduce additional mask parameters to regularize the volume of Gaussians, using binary masks throughout the training process to iteratively eliminate Gaussians based on their contribution, as indicated by the mask values. Both \textit{LightGaussian} \cite{fan2024lightgaussian} and \textit{EAGLES} \cite{girish2024eagles} use importance-based scoring systems to efficiently eliminate unnecessary Gaussians. LightGaussian assigns each Gaussian a global significance score by evaluating its impact on intersected pixels, pruning those with lower significance to enhance efficiency without compromising rendering quality. Similarly, \textit{EAGLES}~\cite{girish2024eagles} employs a simplified approach by using a weight composed of transmittance and opacity to represent each Gaussian's importance, ensuring that less significant Gaussians are removed while maintaining overall scene accuracy. Papantonakis et al. \cite{papantonakis2024reducing} propose a pruning strategy that eliminates Gaussians based on overlap—those with significant overlap with others are considered redundant and removed to reduce overlap without compromising scene accuracy. \textit{SUNDAE} \cite{yang2024spectrally} utilizes a graph-based pruning approach, constructing a graph to capture spatial relationships between Gaussians and applying a band-limited graph filter to selectively down-sample them. To counteract the potential loss of information during this process, a convolutional neural network (CNN) is employed to recover fine details, ensuring a balance between efficient Gaussian placement and the preservation of visual quality.

\subsubsection{Quantitative Comparison of Compaction Methods}

This section presents a comparative analysis of 3DGS compaction methods across three datasets: \TandT, \MipNeRF, and \DeepBlending. Detailed descriptions of the datasets can be found in Sec. \ref{sec:datasets}. As shown in Table \ref{tab:compaction}, the performance of each compression method is evaluated using four metrics: PSNR, SSIM, LPIPS, and the number of Gaussians. Additionally, we compute an overall rank that averages the rankings of the methods across all datasets. To determine the compaction dataset ranks $rank_g$, the quality metrics—PSNR, SSIM, and LPIPS—are given equal weight alongside the number of Gaussians. Specifically, each quality metric contributes one-sixth to the total ranking, while the model size (number of Gaussians, denoted as $k$ Gaussians) accounts for half: \\
$rank_g = \frac{rank(PSNR)}{6} + \frac{rank(SSIM)}{6} + \frac{rank(LPIPS)}{6} + \frac{rank(k\ Gaussians)}{2}$.\\
As for the compression rank calculation the rank for compaction methods is calculated by averaging the rankings over all available datasets, to ensure a fair comparison. 

Table~\ref{tab:compaction} highlights that the most efficient compaction methods are those that minimize the number of Gaussians while maintaining reasonable visual quality. \textit{Octree-GS}\cite{ren2024octree} and \textit{Mini-Splatting}\cite{Fang2024MiniSplattingRS} stand out as the most efficient methods, using the fewest Gaussians while still delivering competitive PSNR values. These methods are ideal for applications requiring tight memory and computation limits. On the other hand, methods like \textit{Taming3DGS (Big)}\cite{taming20243dgs} and \textit{GaussianPro}\cite{cheng2024gaussianpro} offer higher quality at the expense of increased Gaussian count, making them more suitable for use cases that prioritize visual fidelity over extreme compaction.

A closer analysis reveals that even the highest-ranked method in terms of efficiency, \textit{Octree-GS}, does not achieve the best performance across all datasets. This suggests that different methods focus on optimizing distinct aspects of the scene. For example, \textit{Octree-GS} excels on datasets like \DeepBlending{} and \TandT{}, where the texture-rich regions are relatively sparse. Its multi-scale strategy enables flexible adjustments of finer details, which allows users to manually refine resolution levels in such regions. In contrast, methods like \textit{Taming3DGS (Big)} perform better in more texture-dense scenes due to their emphasis on higher Gaussian counts and richer detail capture. This underscores the importance of selecting compaction methods based on the specific application needs.

{
\rowcolors{2}{gray!25}{white}
\footnotesize
\setlength{\tabcolsep}{3pt}
\begin{table*}
\caption{Performance comparison of 3DGS compaction methods across three datasets: \TandT, \MipNeRF, and \DeepBlending. The included metrics are PSNR, SSIM, LPIPS and, number of Gaussians. The best methods in each category are highlighted (\colorbox{lightred}{first}, \colorbox{lightorange}{second}, \colorbox{lightyellow}{third}). The rank represents the average rankings of the methods across all available datasets.  The method highlighted in bold: \textbf{3DGS-30K} is the original 3DGS method. \label{tab:compaction}}
\begin{adjustbox}{center}
\begin{tabular}{ll|lrrr|lrrr|lrrr}
\toprule
Method & Rank & \multicolumn{4}{c|}{Tanks and Temples} & \multicolumn{4}{c|}{Mip-NeRF 360} & \multicolumn{4}{c}{Deep Blending} \\
 & \tiny  & \tiny PSNR$\uparrow$ & \tiny SSIM$\uparrow$ & \tiny LPIPS$\downarrow$ & \tiny k Gauss & \tiny PSNR$\uparrow$ & \tiny SSIM$\uparrow$ & \tiny LPIPS$\downarrow$ & \tiny k Gauss & \tiny PSNR$\uparrow$ & \tiny SSIM$\uparrow$ & \tiny LPIPS$\downarrow$ & \tiny k Gauss \\
\midrule
Octree-GS & \cellcolor{lightred}2.7 & \cellcolor{lightred}24.68 & \cellcolor{lightred}.866 & \cellcolor{lightorange}.153 & 443 & \cellcolor{lightred}28.05 & \cellcolor{lightyellow}.819 & .217 & 657 & \cellcolor{lightred}30.49 & \cellcolor{lightorange}.912 & .241 & \cellcolor{lightred}112 \\
Mini-Splatting & \cellcolor{lightorange}3.4 & 23.18 & .835 & .202 & \cellcolor{lightred}200 & 27.34 & \cellcolor{lightorange}.822 & .217 & \cellcolor{lightred}490 & \cellcolor{lightyellow}29.98 & \cellcolor{lightyellow}.908 & .253 & \cellcolor{lightyellow}350 \\
Taming3DGS & \cellcolor{lightyellow}4.8 & 23.89 & .835 & .207 & \cellcolor{lightorange}290 & 27.29 & .799 & .253 & \cellcolor{lightorange}630 & 27.79 & .822 & .263 & \cellcolor{lightorange}270 \\
Taming3DGS (Big) & \cellcolor{lightyellow}4.8 & \cellcolor{lightyellow}24.04 & .851 & .170 & 1,840 & \cellcolor{lightorange}27.79 & \cellcolor{lightorange}.822 & \cellcolor{lightorange}.205 & 3,310 & \cellcolor{lightorange}30.14 & .907 & \cellcolor{lightyellow}.235 & 2,810 \\
AtomGS & 4.9 & 23.70 & .849 & \cellcolor{lightyellow}.166 & 1,480 & 27.38 & .816 & \cellcolor{lightyellow}.211 & 3,140 &  &  &  &  \\
GaussianPro & 5.0 & \cellcolor{lightorange}24.09 & \cellcolor{lightorange}.862 & .185 & 1,441 & 27.43 & .813 & .219 & 3,403 & 29.79 & \cellcolor{lightred}.913 & \cellcolor{lightorange}.222 & 2,582 \\
Color-cued GS & 5.5 & 23.18 & .830 & .198 & \cellcolor{lightyellow}370 & 27.07 & .797 & .249 & \cellcolor{lightyellow}646 & 29.71 & .902 & .255 & 644 \\
Mini-Splatting-D & 5.7 & 23.23 & \cellcolor{lightyellow}.853 & \cellcolor{lightred}.140 & 4,280 & \cellcolor{lightyellow}27.51 & \cellcolor{lightred}.831 & \cellcolor{lightred}.176 & 4,690 & 29.88 & .906 & \cellcolor{lightred}.211 & 4,630 \\
\textbf{3DGS-30K} & 6.6 & 23.14 & .841 & .183 & 1,783 & 27.21 & .815 & .214 & 3,362 & 29.41 & .903 & .243 & 2,975 \\
\bottomrule
\end{tabular}
\end{adjustbox}
\end{table*}
}

\section{Datasets and Comparison Metrics}
\subsection{Datasets}\label{sec:datasets}

\begin{figure*}
    \centering
    \begin{adjustbox}{center}
        \begin{tabular}{cccc}

            \multicolumn{2}{c}{\textbf{Tanks and Temples}} & \multicolumn{2}{c}{\textbf{Deep Blending}} \\[0.03cm]
            
            \includegraphics[width=0.2\textwidth]{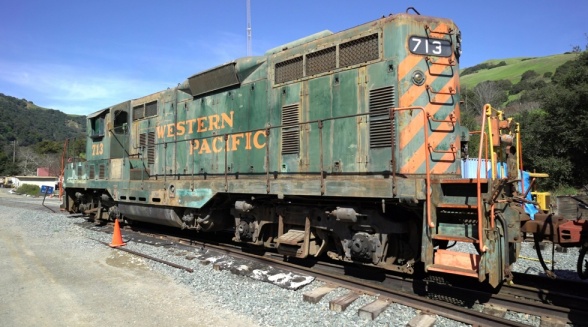} &
            \includegraphics[width=0.2\textwidth]{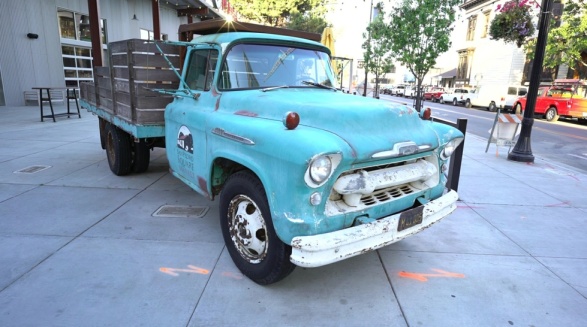} &
            \includegraphics[width=0.2\textwidth]{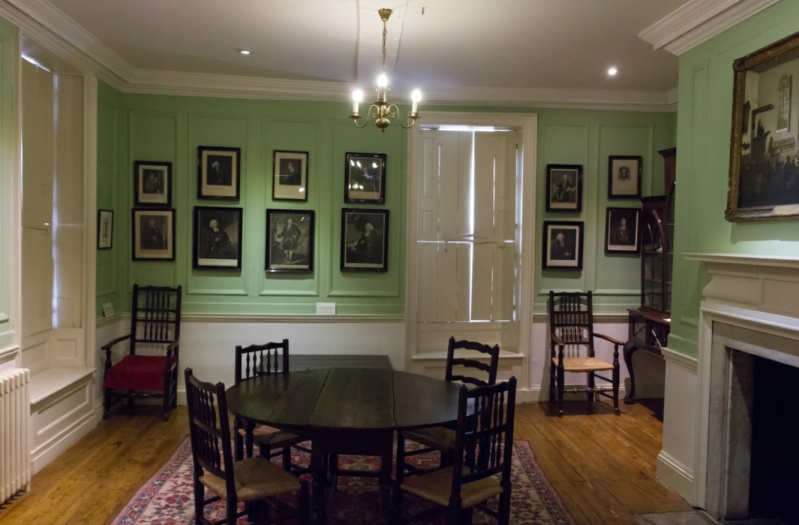} &
            \includegraphics[width=0.2\textwidth]{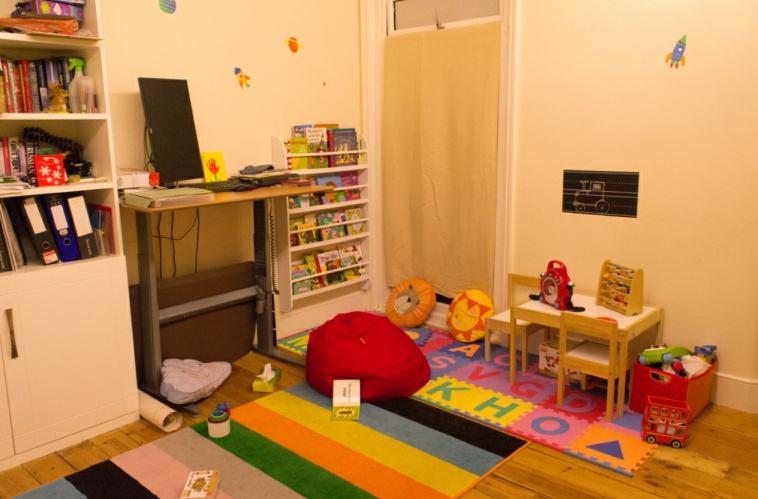} \\
            \text{train} & \text{truck} & \text{Dr Johnson} & \text{playroom} \\

            \\[0.01 cm]
                        
        \end{tabular}
    \end{adjustbox}
    \begin{adjustbox}{center}
        \begin{tabular}{ccc}
    
            \multicolumn{3}{c}{\textbf{Mip-NeRF 360}} \\[0.03cm]

            \includegraphics[width=0.2\textwidth]{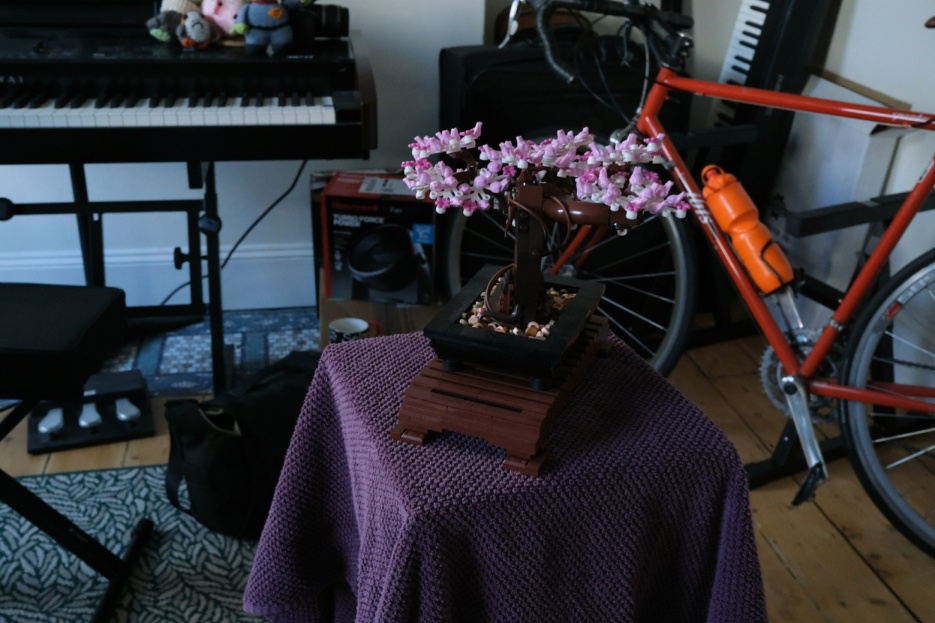} &
            \includegraphics[width=0.2\textwidth]{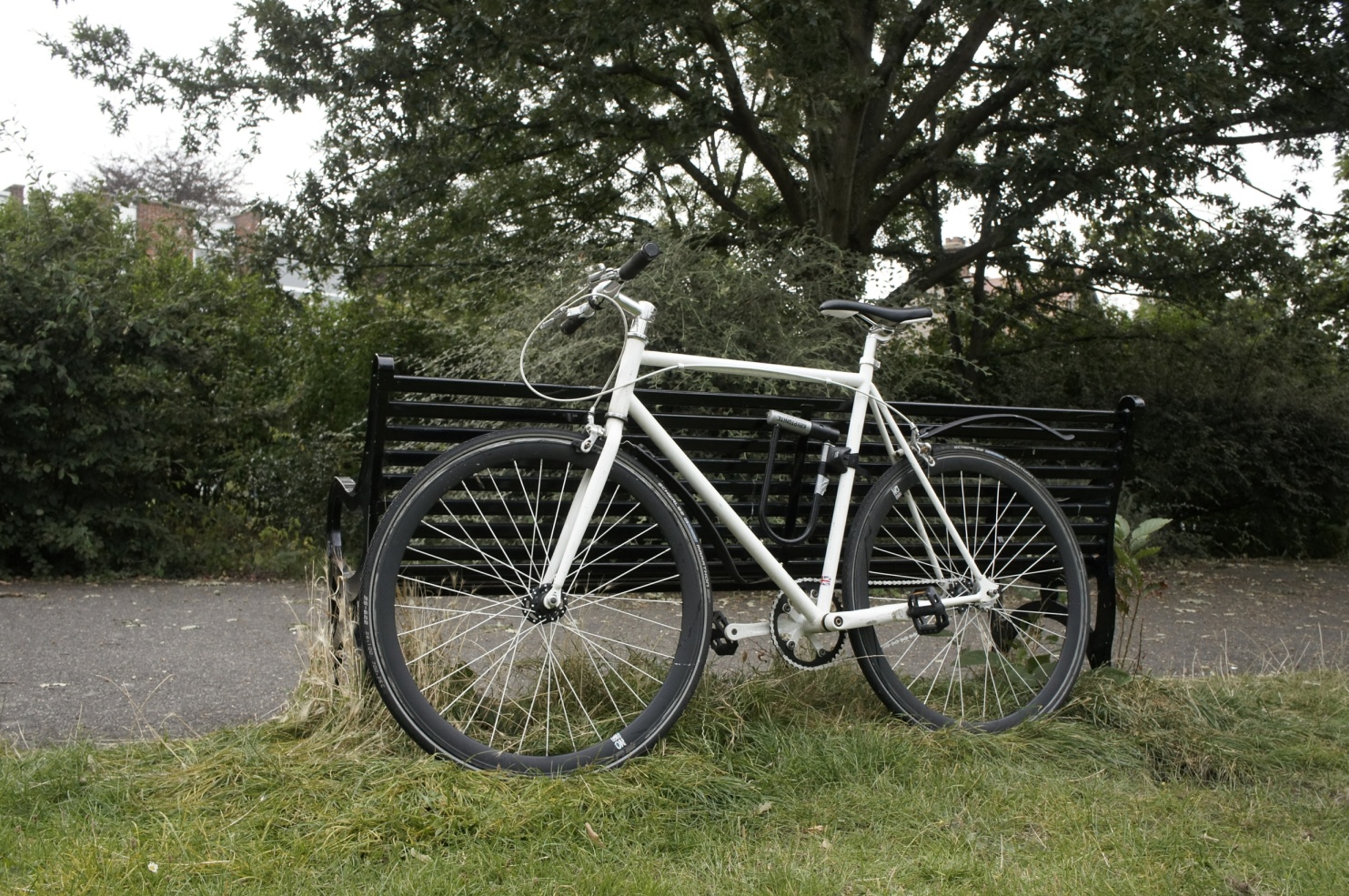} &
            \includegraphics[width=0.2\textwidth]{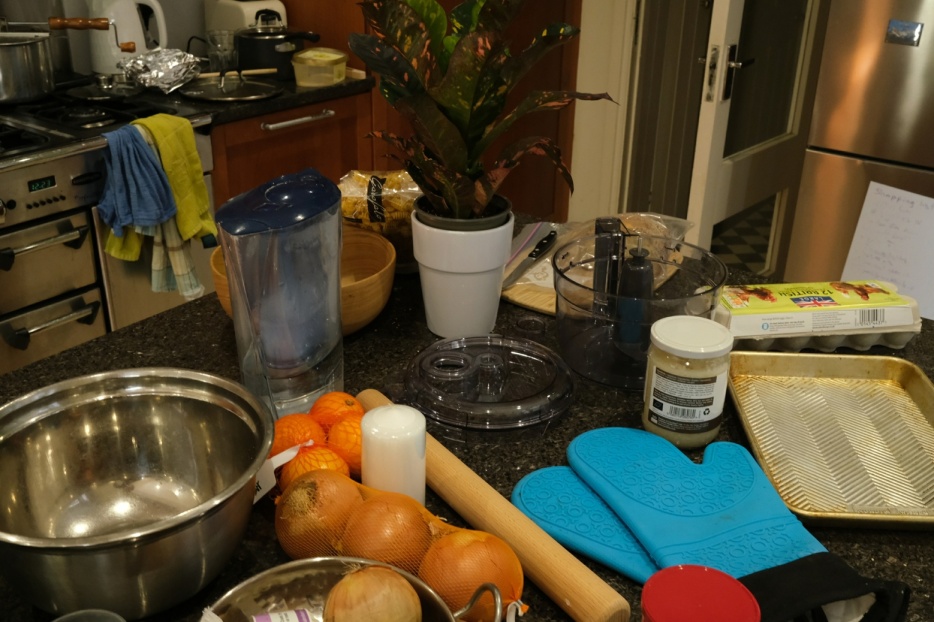} \\
            \text{bonsai} & \text{bicycle} & \text{counter}\\
            \\[0.001cm]
            \includegraphics[width=0.2\textwidth]{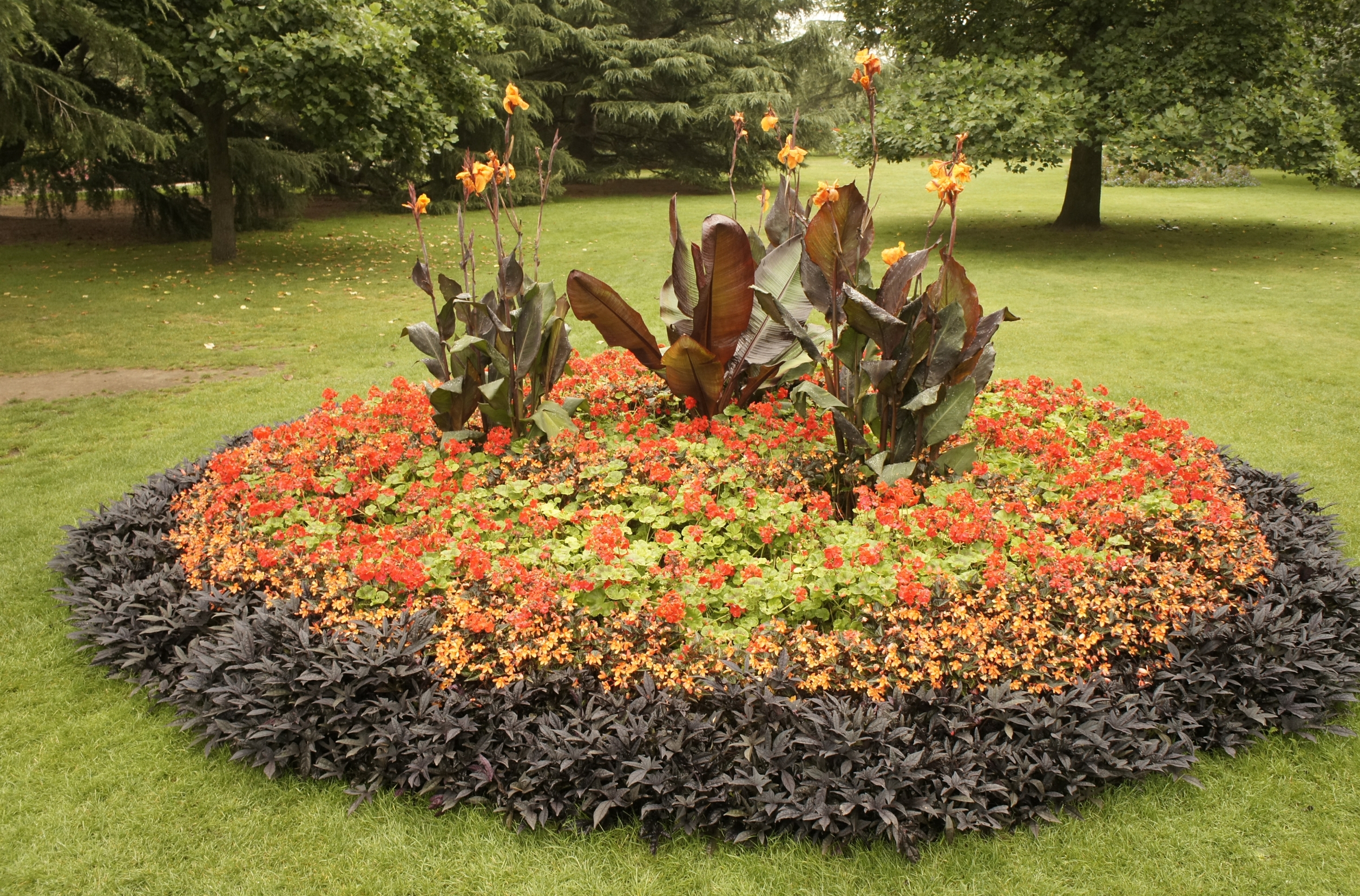} &
            \includegraphics[width=0.2\textwidth]{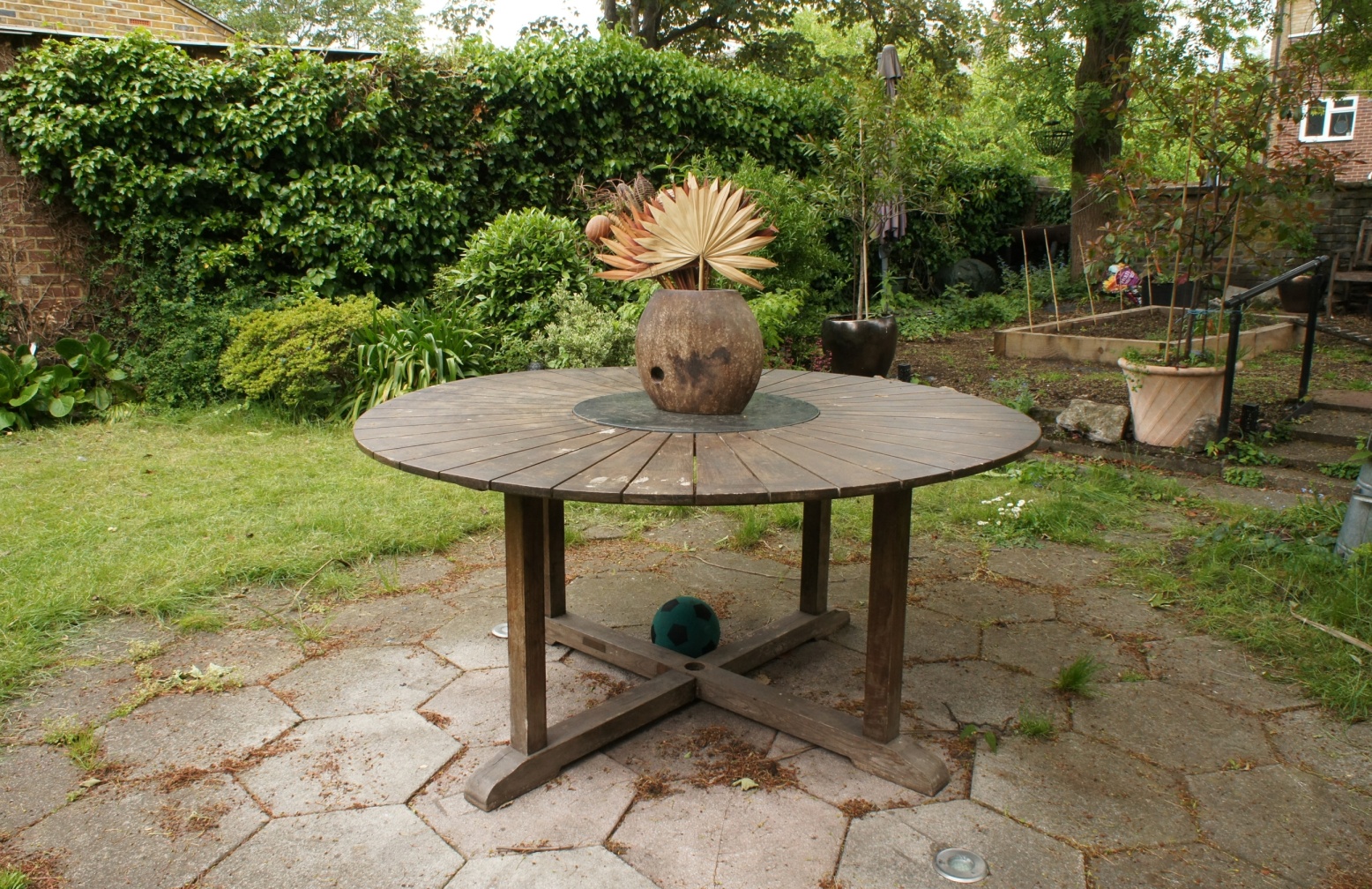} &
            \includegraphics[width=0.2\textwidth]{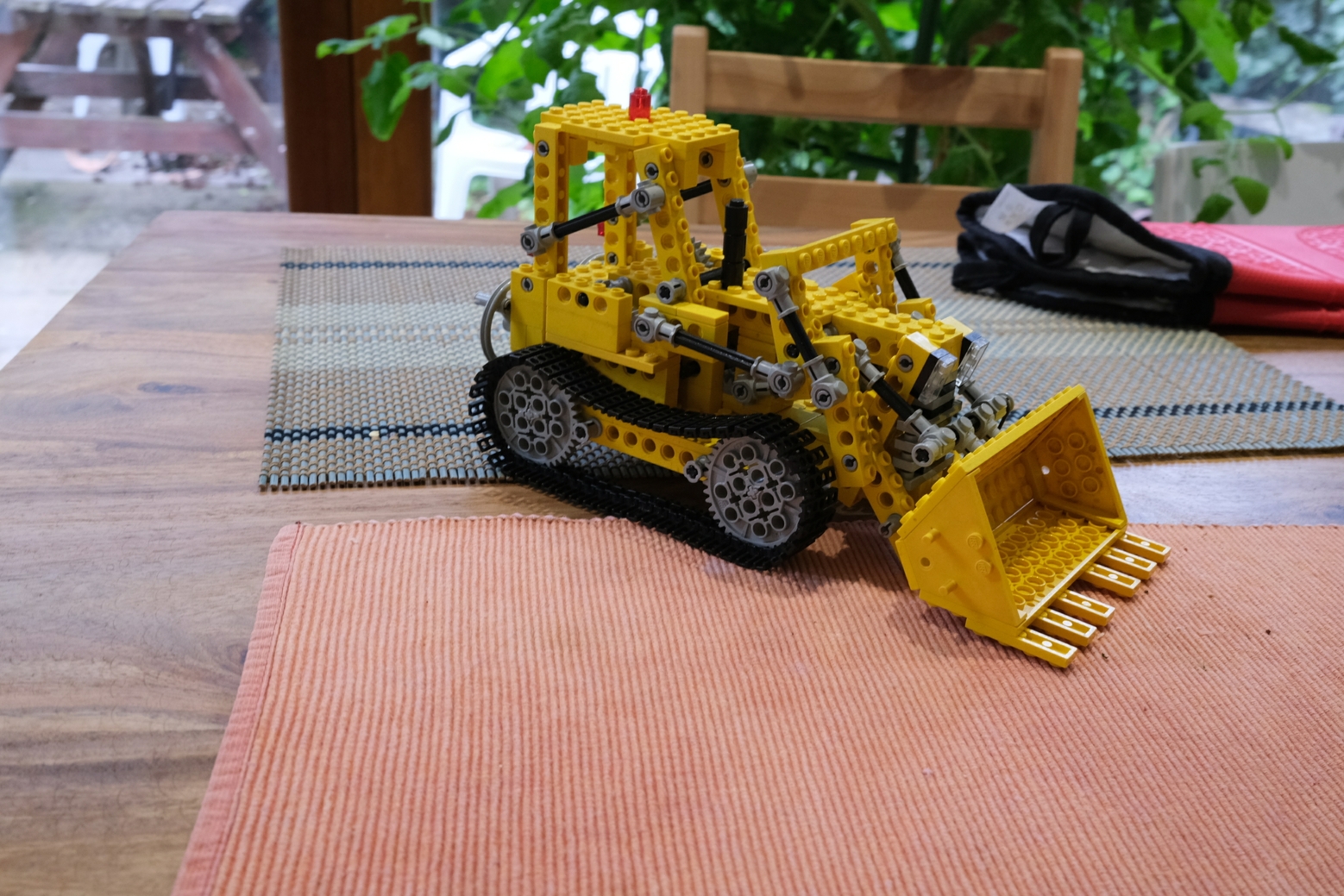} \\
            \text{flowers} & \text{garden} & \text{kitchen} \\
            \\[0.001cm]
            \includegraphics[width=0.2\textwidth]{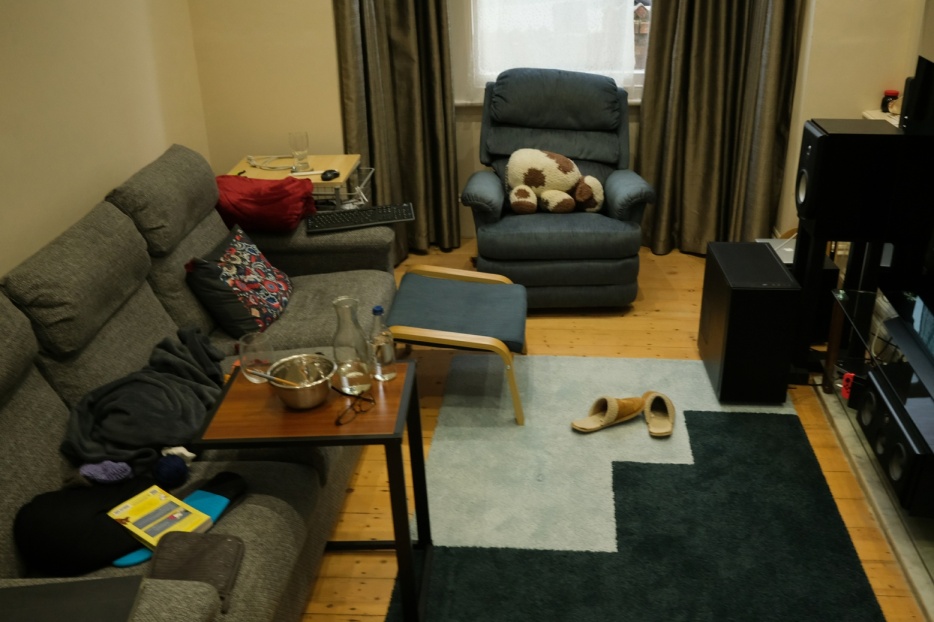} &
            \includegraphics[width=0.2\textwidth]{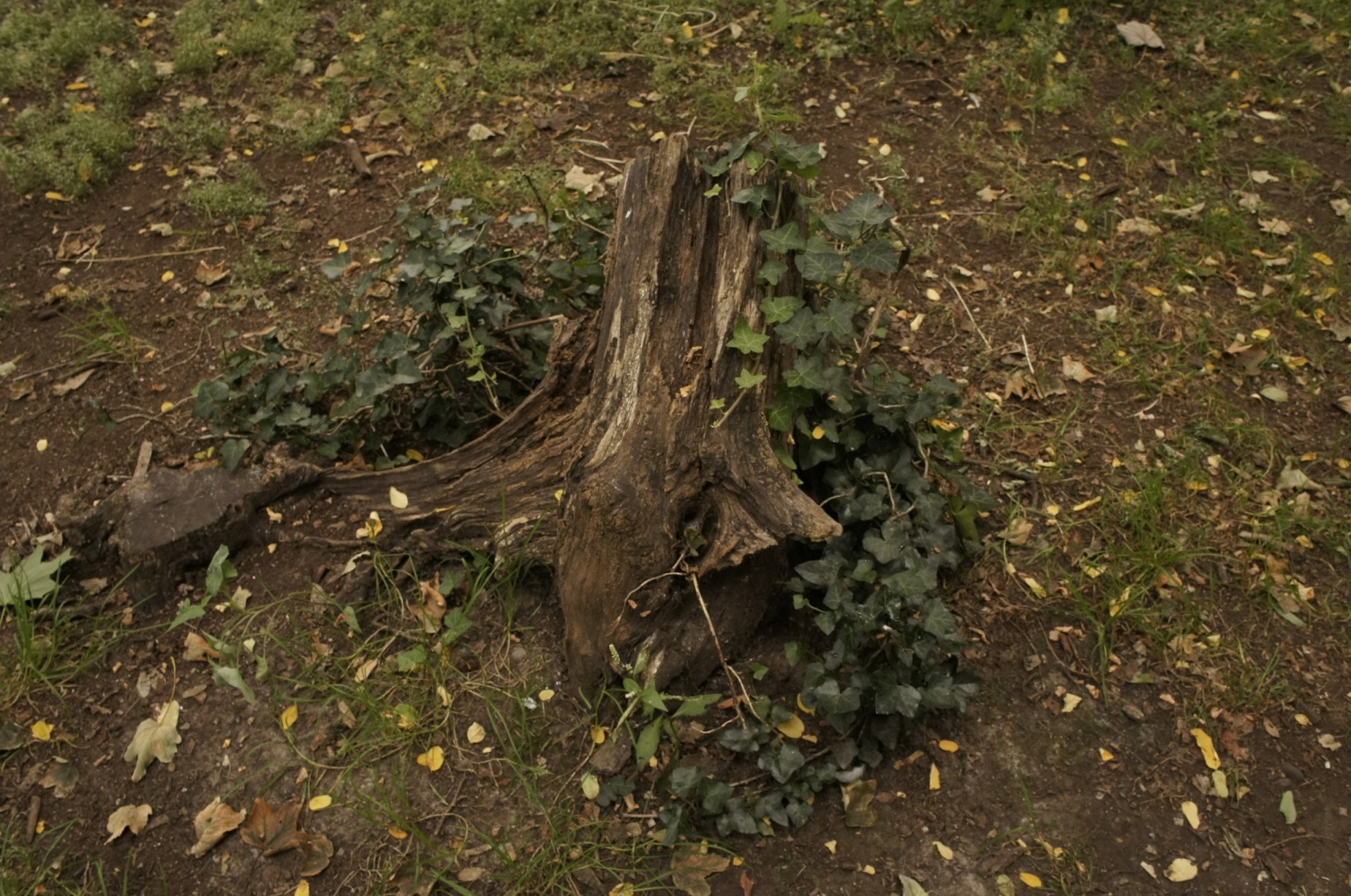} &
            \includegraphics[width=0.2\textwidth]{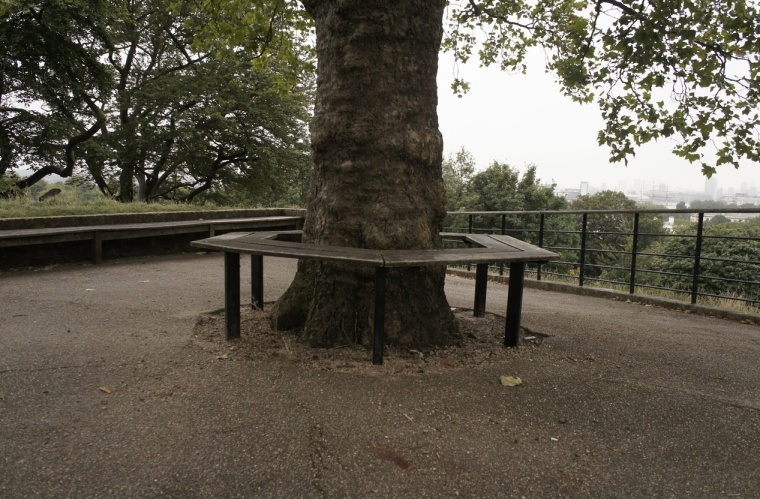} \\
            \text{room} & \text{stump} & \text{treehill} \\

            \\[0.01 cm]

        \end{tabular}
    \end{adjustbox}
    \begin{adjustbox}{center}
        \begin{tabular}{cccc}
            \multicolumn{4}{c}{\textbf{Synthetic NeRF}} \\[0.01cm]
    
            \includegraphics[width=0.15\textwidth]{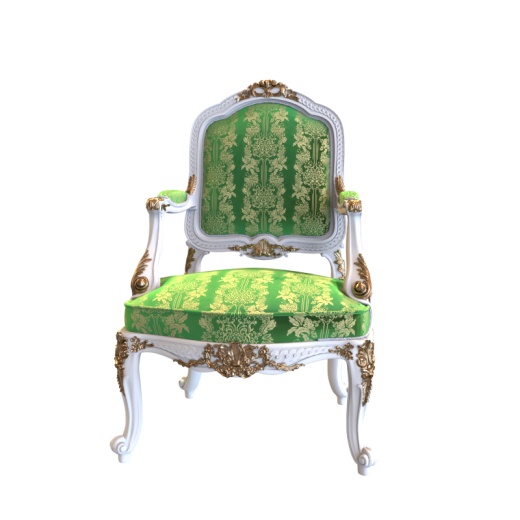} &
            \includegraphics[width=0.15\textwidth]{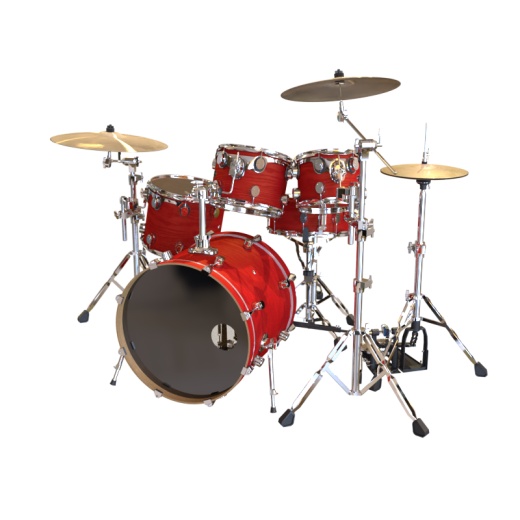} &
            \includegraphics[width=0.15\textwidth]{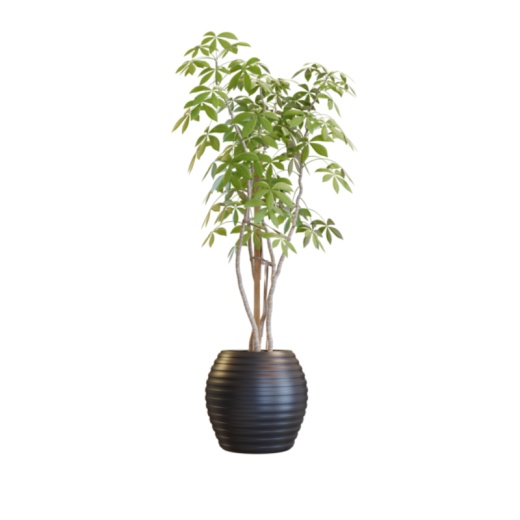} &
            \includegraphics[width=0.15\textwidth]{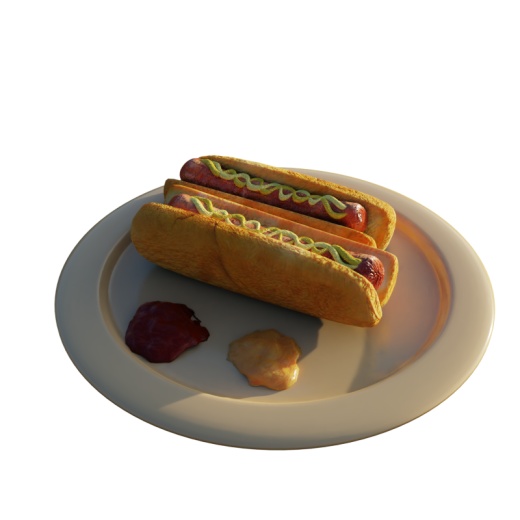} \\
            \text{chair} & \text{drums} & \text{ficus} & \text{hotdog} \\
    
            \includegraphics[width=0.15\textwidth]{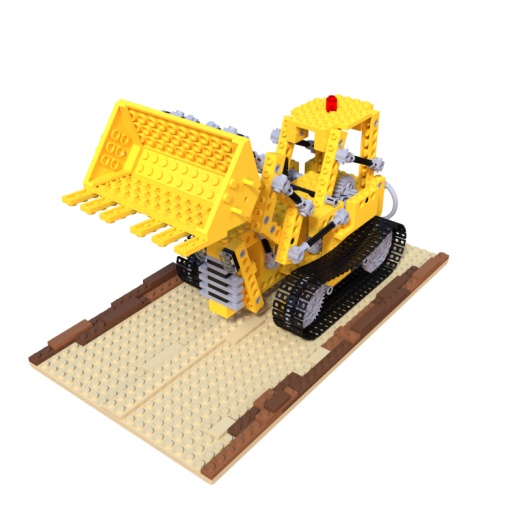} &
            \includegraphics[width=0.15\textwidth]{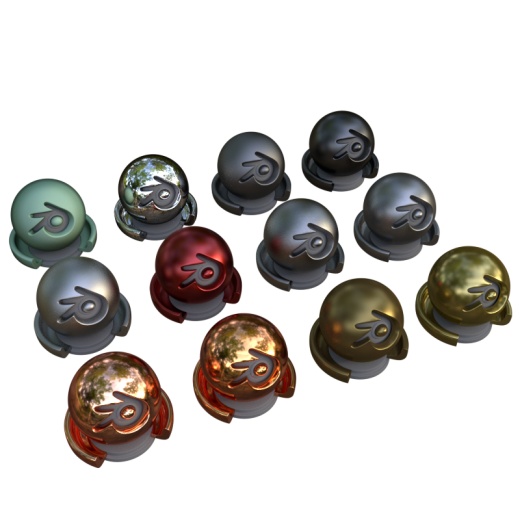} &
            \includegraphics[width=0.15\textwidth]{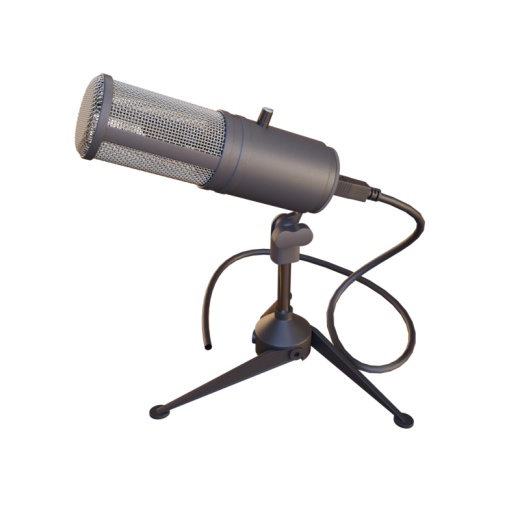} &
            \includegraphics[width=0.15\textwidth]{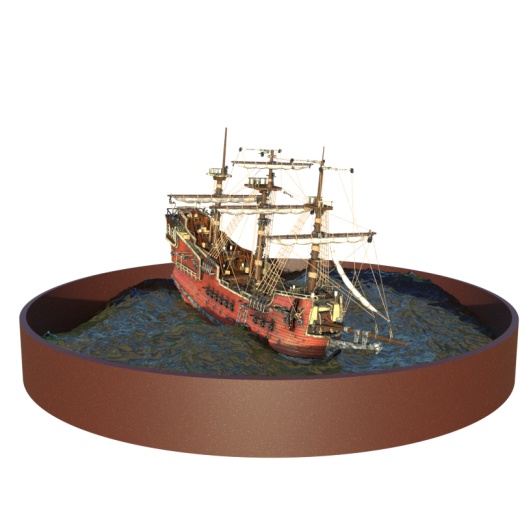} \\
            \text{lego} & \text{materials} & \text{mic} & \text{ship} \\
        \end{tabular}
    \end{adjustbox}
    \\[0.1cm]
    \captionsetup{justification=raggedright}
    \caption{The figure shows a sample image from each scene of the datasets used in evaluation. This selection covers a broad range of scene types, including small to medium-sized natural environments from \TandT{}, \DeepBlending{}, and \MipNeRF{}, alongside highly detailed synthetic scenes with fine textures and reflections provided by \SyntheticNeRF{}. This combination ensures robust evaluation across both real-world and artificial environments.}
    \label{fig:datasets}
\end{figure*}

Performance and quality assessment of 3D Gaussian Splatting algorithms is typically performed on multiple datasets. These datasets provide 3D scenes or objects with various properties, such as varying levels of detail, lighting conditions, and complexities, which allow for comprehensive evaluation of the algorithms. \\
In our survey, we include \TandT\cite{TanksAndTemples}, \MipNeRF\cite{MipNeRF360}, \DeepBlending\cite{DeepBlending} as real-world datasets, and \SyntheticNeRF\cite{SyntheticNeRF} as a synthetic dataset. Figure \ref{fig:datasets} shows a sample image from each included scene. From \TandT{} we include ``truck'' and ``train'' two unbounded outdoor scenes which have a centered view point. The \MipNeRF{} dataset also has a centered view point but includes in- and outdoor scenes. The following scenes are included:  ``bicycle'', ``bonsai'', ``counter'', ``flowers'', ``garden'', ``kitchen'', ``room'', ``stump'', ``treehill''. From the \DeepBlending{} dataset we include ``Dr Johnson'' and ``Playroom'' two indoor scenes with a viewpoint directed outward. The synthetic scenes: ``chair'', ``drums'', ``ficus'', ``hotdog'', ``lego'', ``material'', ``mic'', ``ship'' stem from the \SyntheticNeRF{} dataset.

These scenes align with those used in the 3D Gaussian Splatting (3DGS)~\cite{kerbl3Dgaussians} publication, making them particularly useful for comparing compression methods, as most authors have benchmarked against them. While it would be beneficial to explore larger or more specialized scenes in future work, the lack of accessible data for such comparisons currently limits our scope.

\subsection{Comparison Metrics}\label{sec:metrics}

To evaluate the performance of 3D Gaussian Splatting (3DGS) compression and compaction methods, we rely on a set of well-established metrics that assess both the quality of the rendered scene and the efficiency of the data representation. These metrics include Peak Signal-to-Noise Ratio (PSNR), Structural Similarity Index (SSIM), Learned Perceptual Image Patch Similarity (LPIPS), and model size (in megabytes or number of Gaussians).

\begin{itemize}
    \item \textbf{PSNR} is a widely used metric that quantifies the difference between the original and compressed image in terms of pixel accuracy. Higher PSNR values indicate better fidelity and less distortion.
    \item \textbf{SSIM} measures the perceptual similarity between two images by considering luminance, contrast, and structure. A higher SSIM score represents a closer resemblance between the reference and the rendered scene
    \item \textbf{LPIPS} assesses perceptual quality using a learned model that captures human-like judgments of visual similarity. Lower LPIPS values indicate better perceptual quality.
    \item \textbf{Model Size} is represented either as the file size in megabytes (MB, with $1 \text{ MB} = 1000^2 \text{ Bytes}$) or the total number of Gaussians in the model. Smaller sizes reflect more efficient compression or compaction.
\end{itemize}

\subsection{Testing Protocols}
Initially, our methodology for data compilation for this survey involved parsing various tables from numerous 3DGS compression publications. However, we have then revised our strategy and asked all authors of the publications referenced in this report to provide us with data / results in a standardized format.

In order to ensure consistency and comparability across different approaches, we suggested that authors adhere to the established testing conventions from the original 3DGS project. Specifically, this includes using all 9 scenes from the \MipNeRF{} dataset, incorporating the extra scenes "flowers" and "treehill" (see Sec. \ref{sec:datasets}), and only using the "train" and "truck" scenes from \TandT. For image evaluation, full-resolution images should be used up to a maximum side length of 1600px. For larger test images, downscaling is required so that the longest dimension is 1600px, following the standard 3DGS\cite{kerbl3Dgaussians} resizing method, which uses the PIL .resize() function with bicubic resampling. For the 3 COLMAP datasets (\TandT{}, \DeepBlending{}, \MipNeRF{}), every 8th image must be selected for testing, specifically those images of index $i$ where $i \text{ mod } 8 \equiv 0$. For the \SyntheticNeRF{} dataset, authors should follow the predefined train/evaluation split provided by the dataset.

Table \ref{tab:psnr_comparison} provides a comparative analysis of training and evaluation sizes. The gsplat~\cite{Ye_gsplat} methodology was employed to train \MipNeRF~ scenes at three different resolutions: full resolution, 1600 pixels (longest side), and resolutions downscaled by factors of 2 or 4 (longest side $<1600$ pixels). The findings indicate that the optimal Peak Signal-to-Noise Ratio (PSNR) results are achieved when the evaluation resolution corresponds precisely to the training resolution. Any deviation, whether by increasing or decreasing the evaluation resolution relative to the training resolution, results in a slight reduction in PSNR. Furthermore, utilizing both higher training and evaluation resolutions enhances PSNR. Additionally, the overall quality is enhanced when employing 1 million Gaussians compared to 360k Gaussians, which aligns with expectations, as rendering scenes with 1 million Gaussians captures more details. Consequently, it is advised to carefully select training and evaluation sizes, acknowledging that this decision affects the final PSNR outcome as well as the comparative analysis of different compression approaches.

\footnotesize
\setlength{\tabcolsep}{3pt}
\begin{table}
\centering
\caption{PSNR comparison table for training sizes vs. evaluation sizes. For this table we used gsplat\cite{Ye_gsplat} to train \MipNeRF~scenes either in full resolution, with a resolution of 1600 pixels or scaled down with a factor of 2 or 4 (resulting in a resolution less than 1600 pixels, as used in the implementations~\cite{MipNeRF360,kerbl3Dgaussians}). Evaluation is performed using full resolution or 1600 pixels (longest dimension).}
\label{tab:psnr_comparison}
\begin{tabular}{lcc|cc}
\toprule
{} & \multicolumn{2}{c|}{PSNR} & \multicolumn{2}{c}{PSNR} \\
{} & \multicolumn{2}{c|}{(eval. res. full)} & \multicolumn{2}{c}{(eval. res. 1600px)} \\
{$\#$Gaussians $\xrightarrow{}$} & 360 K & 1 M & 360 K & 1 M \\
\midrule
train. res. full & \cellcolor{lightred}{26.38} & \cellcolor{lightred}{26.97} & 26.33 & 26.67 \\
scaling factor 2/4 & 25.58 & 25.84 & 26.40 & 27.02 \\
train. res. 1600px & 25.82 & 26.05 & \cellcolor{lightred}{26.69} & \cellcolor{lightred}{27.33} \\

\bottomrule
\end{tabular}
\end{table}

\section{Discussion}

The field of 3D Gaussian Splatting (3DGS) has seen rapid advancements, with significant improvements in compression and compaction techniques aimed at overcoming the high storage demands traditionally associated with 3DGS. While the original implementation of 3D Gaussian Splatting~\cite{kerbl3Dgaussians} trained on scenes from the \TandT dataset resulted in a data size exceeding 400 MB, the most effective compression methods achieve more than a 40-fold reduction in size, all the while enhancing visual quality. This state-of-the-art report reveals key insights into these developments, providing a foundation for future optimization and application of 3DGS in real-time rendering.

A crucial finding is that a well-implemented compaction strategy can be an improvement to most methods. Adaptive Density Control (ADC), with densification and pruning, has proven to be effective in significantly enhancing visual quality while reducing the storage footprint. This is vital for applications in resource-constrained environments, such as mobile devices or VR headsets, where storage, memory and processing power are limited but visual fidelity matters to the user.
Structured representations also play a key role in achieving efficient compression.  While various compression techniques have been developed independently, their interplay remains an open research area. Understanding the interactions between different methods can lead to more effective strategies, such as vector quantization reducing redundancy by encoding similar Gaussians with shared representations, or structured representations like octrees and hash grids spatially organizing Gaussians to improve compression. Further research into integrating vector quantization within structured representations could lead to more effective encodings.


Techniques that prune Gaussian attributes, such as spherical harmonics (SH) coefficients, must be balanced against compaction strategies like ADC, which reduces the number of Gaussians. Determining the optimal trade-off between reducing attributes and minimizing Gaussian count remains an important area of study. Some compression methods prioritize extreme size reduction through entropy coding and quantization, while others focus on preserving fidelity using hierarchical encoding. Hybrid approaches that dynamically adjust compression levels based on scene complexity could be highly beneficial. Compression techniques must balance reducing file size with maintaining rendering accuracy. Neural feature encoding techniques, such as tri-planes or hash grids, significantly reduce storage needs but require additional computation during rendering. In contrast, explicit representations with precomputed attributes demand more storage but enable real-time rendering with minimal overhead. Lossy compression techniques like pruning or aggressive quantization can introduce artifacts, particularly in view-dependent rendering. Knowledge distillation techniques that encapsulate high-frequency details into lower-order representations offer a promising approach to maintaining perceptual quality while reducing complexity.

Another key challenge is the lack of standard benchmarks for evaluating 3DGS compression methods. Current studies use different datasets, metrics, and experimental conditions, making direct comparisons difficult. In this survey we propose basic guidelines to enable comparison but establishing a unified benchmarking framework with standardized datasets and evaluation protocols would enhance reproducibility and facilitate progress in the field. In addition, methods should be evaluated not only on file size and reconstruction quality but also on real-time performance and energy efficiency, particularly for deployment on mobile and embedded systems.

Finally, the trade-off between compression efficiency and visual quality remains a central challenge. Approaches like \textit{HAC-highrate} excel in achieving high-quality compression but at the cost of higher computational overhead. On the other hand, more aggressive compression techniques such as \textit{SOG w/o SH} reduce memory usage significantly but may result in noticeable quality degradation, especially in scenes requiring fine detail. This trade-off highlights the need for flexible solutions that can be tuned based on the specific requirements of the application, whether it's focused on minimizing storage or maximizing visual realism. As the field advances, hybrid approaches that integrate machine learning-driven optimizations may offer new opportunities. Deep learning models could be trained to predict optimal compression strategies based on scene characteristics, dynamically selecting techniques that maximize efficiency while preserving quality. 

\section{Conclusion and Future Directions}

3D Gaussian Splatting (3DGS) has emerged as a powerful alternative to neural radiance fields, offering explicit control over scene elements while achieving high rendering fidelity. However, its practicality remains constrained by storage and computational costs. Scalability, ease of use, and adaptability across different platforms can still be improved. This state-of-the-art-report provides an overview of existing compression and compaction techniques, highlighting key advances and challenges in the field and the importance of continuing to refine and standardize 3D Gaussian Splatting (3DGS) compression and compaction techniques to make them more accessible and widely applicable.

Most current 3DGS compression techniques are designed for static scenes. However, real-world applications increasingly demand dynamic and interactive capabilities. Existing methods rely on pre-trained models, limiting adaptability. Developing real-time adaptive compression strategies that adjust based on scene complexity such as changing objects or lighting conditions over time, or changing hardware constraints would be a valuable research direction. Further advancement could enable real-time simulations and enhance interactive applications such as gaming and virtual reality. 

Another improvement could be the creation of multi resolution models with support for level-of-detail (LOD) scaling. Such models would optimize performance by allowing different parts of a scene to be rendered with varying levels of detail, depending on real-time requirements. Octrees or feature grids, provide scalable solutions, but ensuring seamless transitions between different levels of detail without introducing artifacts remains a challenge.
Quantization-aware training and the development of shared codebooks across scenes or applications could further improve compression efficiency, allowing for reduced redundancy and more effective memory usage.

Together, these advancements will help expand the applicability of 3DGS to encompass to real-time applications, immersive environments, or large-scale scene representations making 3DGS a more versatile and efficient tool for addressing future challenges in the field of computational graphics.


\section{Short Summaries of Included Compression Approaches}

In this section, we provide short overviews of the key compression publications surveyed in this report. Each approach offers unique methods to address the challenges of memory and computational efficiency in 3D Gaussian Splatting (3DGS), focusing on aspects such as attribute pruning, vector quantization, and structured representations. By leveraging different strategies, these methods achieve a balance between compression ratio, visual quality, and rendering performance.

The summaries below highlight the primary innovations of each technique, the specific problems they address, and their contributions to advancing 3DGS compression. This section serves as a quick reference for understanding how each method fits into the broader landscape of 3DGS optimization, making it easier to identify the most suitable approach for applications and research needs.

\subsection{ContextGS: Compact 3D Gaussian Splatting with Anchor Level Context Model (\textit{ContextGS})}
The authors of \textit{ContextGS}~\cite{wang2024contextgs} proposes the first auto-regressive model at the anchor level for 3DGS compression. This work divides anchors into different levels and the anchors that are not coded yet can be predicted based on the already coded ones in all the coarser levels, leading to more accurate modeling and higher coding efficiency. To further improve the efficiency of entropy coding, a low-dimensional quantized feature is introduced as the hyperprior for each anchor, which can be effectively compressed. This work can be applied to both Scaffold-GS and vanilla 3DGS.

\subsection{HAC: Hash-grid Assisted Context for 3D Gaussian Splatting Compression (\textit{HAC})}
The paper proposes a Hash-grid Assisted Context (HAC) framework~\cite{chen2024hac} for compressing 3D Gaussian Splatting (3DGS) models by leveraging the mutual information between attributes of unorganized 3D Gaussians (anchors) and hash grid features. Using Scaffold-GS~\cite{lu2024scaffold} as a base model, HAC queries the hash grid by anchor location to predict anchor attribute distributions for efficient entropy coding. The framework introduces an Adaptive Quantization Module (AQM) to dynamically adjust quantization step sizes. Furthermore, this method employs adaptive offset masking with learnable masks to eliminate invalid Gaussians and anchors, by leveraging the pruning strategy introduced by Compact3DGS~\cite{lee2024compact} and additionally removing anchors if all the attached offsets are pruned.

\subsection{Compression of 3D Gaussian Splatting with Optimized Feature Planes and Standard Video Codecs (\textit{CodecGS})}

This method~\cite{lee2025compression3dgaussiansplatting} introduces an effective approach for compressing 3D Gaussian Splatting by employing optimized feature planes and integrating them with standard video codecs. More specifically CodecGS introduces progressive tri-planes, where the tri-plane takes 3D point positions x as input and predicts the corresponding attributes for each point. A two-phase training, starting with standard 3DGS~\cite{kerbl3Dgaussians} training followed by feature plane training ensures densification is consistent. Progressive training addresses the instability of feature plane training due to sparse 3DGS signals. DCT entropy modeling is employed to transform the feature planes. After training, feature planes are normalized and converted to 16-bit integers, corresponding to the YUV 16-bit format, and are encoded by a standard video codec.

\subsection{gsplat (\textit{gsplat})}
This approach leverages 3D Gaussian Splatting as Markov Chain Monte Carlo (3DGS-MCMC)~\cite{mcmc}, interpreting the training process of positioning and optimizing Gaussians as a sampling procedure rather than minimizing a predefined loss function. Additionally, it incorporates compression techniques derived from the Morgenstern et al.\ paper~\cite{morgenstern2024compact}, which organizes the parameters of 3DGS in a 2D grid, capitalizing on perceptual redundancies found in natural scenes, thereby significantly reducing storage requirements. Further compression is achieved by clustering spherical harmonics into discrete elements and storing them as FP16 values. This technique is implemented in gsplat~\cite{Ye_gsplat}, an open-source library designed for CUDA-accelerated differentiable rasterization of 3D Gaussians, equipped with Python bindings.


\subsection{Compact3D: Compressing Gaussian Splat Radiance Field Models with Vector Quantization (\textit{Compact3D})}
This approach~\cite{navaneet2023compact3d} introduces a vector quantization method based on the K-means algorithm to quantize the Gaussian parameters in 3D Gaussian splatting, as many Gaussians may share similar parameters. Only a small codebook is stored along with the index of the code for each Gaussian, resulting in a large reduction in the storage of the learned radiance fields and a reduction of the memory footprint at rendering time. Additionally, the indices are further compressed by sorting the Gaussians based on one of the quantized parameters and storing the indices using a method similar to Run-Length-Encoding (RLE). To reduce the number of Gaussians, this method applies a regularizer to encourage zero opacity before pruning Gaussians with opacity smaller than a threshold.

\subsection{CompGS: Efficient 3D Scene Representation via Compressed Gaussian Splatting \textit{CompGS}}
The paper CompGS~\cite{liu2024compGS} proposes a hybrid primitive structure with anchor primitives to predict the attributes of coupled primitives, resulting in compact residual representations. A rate-constrained optimization scheme further enhances compactness by jointly minimizing both rendering distortion and bit rate. The bit rate of both anchor and coupled primitives is modeled by entropy estimation.

\subsection{End-to-End Rate-Distortion Optimized 3D Gaussian Representation (\textit{RDO-Gaussian})}
This paper~\cite{wang2024end} introduces RDO-Gaussian, an end-to-end Rate-Distortion Optimized 3D Gaussian representation. The authors achieve flexible, continuous rate control by formulating 3D Gaussian representation learning as a joint optimization of rate and distortion. Rate-distortion optimization is realized through dynamic pruning and entropy-constrained vector quantization (ECVQ). Gaussian pruning involves learning a mask to eliminate redundant Gaussians, and adaptive SH pruning assigns varying SH degrees to each Gaussian based on material and illumination needs. The covariance and color attributes are discretized through ECVQ, which performs vector quantization.

\subsection{Reducing the Memory Footprint of 3D Gaussian Splatting (\textit{Reduced3DGS})}
This approach~\cite{papantonakis2024reducing} addresses three main issues contributing to large storage sizes in 3D Gaussian Splatting (3DGS). To reduce the number of 3D Gaussian primitives, the authors introduce a scale- and resolution-aware redundant primitive removal method. This extends opacity-based pruning by incorporating a redundancy score to identify regions with many low-impact primitives. To mitigate storage size due to spherical harmonic coefficients, they propose adaptive adjustment of spherical harmonic (SH) bands. This involves evaluating color consistency across views and reducing higher-order SH bands when view-dependent effects are minimal. Additionally, recognizing the limited need for high dynamic range and precision for most primitive attributes, they develop a codebook using K-means clustering and apply 16-bit half-float quantization to the remaining uncompressed floating point values.

\subsection{Compact 3D Scene Representation via Self-Organizing Gaussian Grids (\textit{SOG})}

Compressing 3D data is challenging, but many effective solutions exist for compressing 2D data (such as images). The authors propose a new method~\cite{morgenstern2024compact} to organize 3DGS parameters into a 2D grid, drastically reducing storage requirements without compromising visual quality. This organization exploits perceptual redundancies in natural scenes. They introduce a highly parallel sorting algorithm, PLAS, which arranges Gaussian parameters into a 2D grid, maintaining local neighborhood structure and ensuring smoothness. This solution is particularly innovative because no existing method efficiently handles a 2D grid with millions of points. During training, a smoothness loss is applied to enforce local smoothness in the 2D grid, enhancing the compressibility of the data. The key insight is that smoothness needs to be enforced during training to enable efficient compression.

\subsection{MesonGS: Post-training Compression of 3D Gaussians via Efficient Attribute Transformation (\textit{MesonGS})}

MesonGS~\cite{xie2024mesongs} employs universal Gaussian pruning by evaluating the importance of Gaussians through forward propagation, considering both view-dependent and view-independent features. It transforms rotation quaternions into Euler angles to reduce storage requirements and applies region adaptive hierarchical transform (RAHT) to reduce entropy in key attributes. Block quantization is performed on attribute channels by dividing them into multiple blocks and performing quantization for each block individually, using vector quantization for compressing less important attributes. Geometry is compressed using an octree, and all elements are packed with the LZ77 codec. A finetune scheme is implemented post-training to restore quality.

\subsection{Compressed 3D Gaussian Splatting for Accelerated Novel View Synthesis (\textit{Compressed3D})}

The authors propose a compressed 3D Gaussian representation~\cite{niedermayr2024compressed} consisting of three main steps: 1. sensitivity-aware clustering, where scene parameters are measured according to their contribution to the training images and encoded into compact codebooks via sensitivity-aware vector quantization; 2. quantization-aware fine-tuning, which recovers lost information by fine-tuning parameters at reduced bit-rates using quantization-aware training; and 3. entropy encoding, which exploits spatial coherence through entropy and run-length encoding by linearizing 3D Gaussians along a space-filling curve. Furthermore, a renderer for the compressed scenes utilizing GPU-based sorting and rasterization is proposed, enabling real-time novel view synthesis on low-end devices.

\subsection{Compact 3D Gaussian Representation for Radiance Field (\textit{Compact3DGS})}
This approach~\cite{lee2024compact} introduces a Gaussian volume mask to prune non-essential Gaussians and a compact attribute representation for both view-dependent color and geometric attributes. The volume-based masking strategy combines opacity and scale to selectively remove redundant Gaussians. For color attribute compression, spatial redundancy is exploited by incorporating a grid-based (Instant-NGP) neural field, allowing efficient representation of view-dependent colors without storing attributes per Gaussian. Given the limited variation in scale and rotation, geometric attribute compression employs a compact codebook-based representation to identify and reuse similar geometries across the scene. Additionally, the authors propose quantization and entropy-coding as post-processing steps for further compression.

\subsection{EAGLES: Efficient Accelerated 3D Gaussians with Lightweight EncodingS (\textit{EAGLES})}

The authors of this approach~\cite{girish2024eagles} observed that in 3DGS, the color and rotation attributes account for over 80\% of memory usage; thus, they propose compressing these attributes via a latent quantization framework. Additionally, they quantize the opacity coefficients of the Gaussians, improving optimization and resulting in fewer floaters or visual artefacts in novel view reconstructions. To reduce the number of redundant Gaussians resulting from frequent densification (via cloning and splitting), the approach employs a pruning stage to identify and remove Gaussians with minimal influence on the full reconstruction. For this, an influence metric is introduced, which considers both opacity and transmittance.

\subsection{Scaffold-GS: Structured 3D Gaussians for View-Adaptive Rendering (\textit{Scaffold-GS})}
Scaffold-GS~\cite{lu2024scaffold} introduces anchor points that leverage scene structure to guide the distribution of local 3D Gaussians. Attributes like opacity, color, rotation, and scale are dynamically predicted for Gaussians linked to each anchor within the viewing frustum, enabling adaptation to different viewing directions and distances. Initial anchor points are derived by voxelizing the sparse, irregular point cloud from Structure from Motion (SfM), forming a regular grid. Gaussians are spatially quantized using voxels to refine and grow the anchors, with new anchors created at the centers of significant voxels, which are identified by their average gradient over N training steps. Random elimination and opacity-based pruning regulate anchor growth and refinement.

\subsection{LightGaussian: Unbounded 3D Gaussian Compression with 15x Reduction and 200+ FPS (\textit{LightGaussian})}

LightGaussian~\cite{fan2024lightgaussian} aims to transform 3D Gaussians to a more efficient and compact form, avoiding the scalability issues that arise from a large number of SfM (Structure from Motion) points for unbounded scenes. Inspired by Network Pruning, the method identifies Gaussians that minimally contribute to scene reconstruction and employs a pruning and recovery process, thereby efficiently reducing redundancy in Gaussian counts while maintaining visual effects. Additionally, LightGaussian utilizes knowledge distillation and pseudo-view augmentation to transfer spherical harmonics coefficients to a lower degree. Furthermore, the authors propose a Gaussian Vector Quantization based on the global significance of Gaussians to quantize all redundant attributes, achieving lower bit-width representations with minimal accuracy losses.

\section{Short Summaries of Included Compaction Approaches}
Compaction and pruning jointly involve determining whether to introduce or eliminate Gaussians based on criteria aimed at improving scene accuracy and optimizing computational resources. The original criteria used in 3D Gaussian Splatting often fail to capture high-frequency details, leading to over-reconstruction or under-reconstruction. To address this, refined methods have introduced criteria that better balance computational efficiency and realistic scene modelling. Compaction concentrates on identifying where additional kernels are needed to capture missing details, especially in complex or high-frequency areas, while pruning removes redundant or ineffective ones, ensuring that superfluous kernels—those that do not add value to the reconstruction—are eliminated and optimizing both efficiency and rendering fidelity.

\subsection{Mini-Splatting: Representing Scenes with a Constrained Number of Gaussians (\textit{Mini-Splatting})}

Mini-Splatting~\cite{Fang2024MiniSplattingRS} enhances Gaussian distribution through Blur Split, which refines Gaussians in blurred regions, and Depth Reinitialization, which repositions Gaussians based on newly generated depth points, calculated from the mid-point of ray intersections with Gaussian ellipsoids, thus avoiding artifacts from alpha blending. For simplification, Intersection Preserving retains Gaussians with the greatest visual impact, while Sampling maintains geometric integrity and rendering quality, reducing complexity.

\subsection{Octree-GS: Towards Consistent Real-time Rendering with LOD-Structured 3D Gaussians (\textit{Octree-GS})}

Octree-GS~\cite{ren2024octree} introduces an octree structure to 3D Gaussian splatting. Starting with a sparse point cloud, an octree is constructed for the bounded 3D space, where each level corresponds to a set of anchor Gaussians assigned to different levels of detail (LOD). This method selects the necessary LOD based on the observation view, gradually accumulating Gaussians from higher LODs for final rendering. The model is trained using standard image reconstruction and volume regularization losses.

\subsection{Taming 3DGS: High-Quality Radiance Fields with Limited Resources (\textit{Taming3DGS})}

Taming 3DGS~\cite{taming20243dgs} employs a global scoring approach to guide the addition of Gaussians, ensuring efficient densification. Each Gaussian is assigned a score based on four factors: 1) gradient, 2) pixel coverage, 3) per-view saliency, and 4) core attributes like opacity, depth, and scale. Gaussians with the top B scores, where B is the desired number of new Gaussians, are then split or cloned to optimize the scene's representation. By calculating a composite score that reflects both the scene’s structural complexity and visual importance, only the most critical areas are targeted for Gaussian splitting or cloning, resulting in more effective scene representation.

\subsection{AtomGS: Atomizing Gaussian Splatting for High-Fidelity Radiance Field (\textit{AtomGS})}

AtomGS~\cite{liu2024atomgs} prioritizes fine details through Atom Gaussians, which are isotropic and uniformly sized to align closely with the scene's geometry, while large Gaussians are merged to cover smooth surfaces. In addition, Geometry-Guided Optimization uses an Edge-Aware Normal Loss and multi-scale SSIM to maintain geometric accuracy. The Edge-Aware Normal Loss is calculated as the product of the normal map, derived from the pre-optimized 3DGS, and the edge map, which is derived from the gradient magnitude of the ground truth RGB image.

\subsection{Color-cued Efficient Densification Method for 3D Gaussian Splatting (\textit{Color-cued GS})}

This method~\cite{Kim_2024_CVPR} introduces a simple yet effective modification to the densification process in the original 3D Gaussian Splatting (3DGS). It leverages the view-independent (0th) spherical harmonics (SH) coefficient gradient to better assess color cues for densification, while using the 2D position gradient more coarsely to refine areas where structure-from-motion (SfM) struggles to capture fine structures.

\subsection{GaussianPro: 3D Gaussian Splatting with Progressive Propagation (\textit{GaussianPro})}

GaussianPro~\cite{cheng2024gaussianpro} generates depth and normal maps that guide the growth and adjustment of Gaussians. It employs patch matching to propagate depth and normal information from neighboring pixels to generate new values. Geometric filtering and selection then identify pixels needing additional Gaussians, which are initialized using the propagated information. It also introduces a planar loss to ensure Gaussians match real surfaces more closely. This method enforces consistency between the Gaussian's rendered normal and the propagated normal using L1 and angular loss.

\bibliographystyle{habbrv} 
\bibliography{datasets,
    survey,
    methods_compression}

\appendix
\section{Additional figures on attribute statistics.}
\label{ap:stats}

\begin{figure*}[htb]
    \centering
    \includegraphics[width=0.8\linewidth]{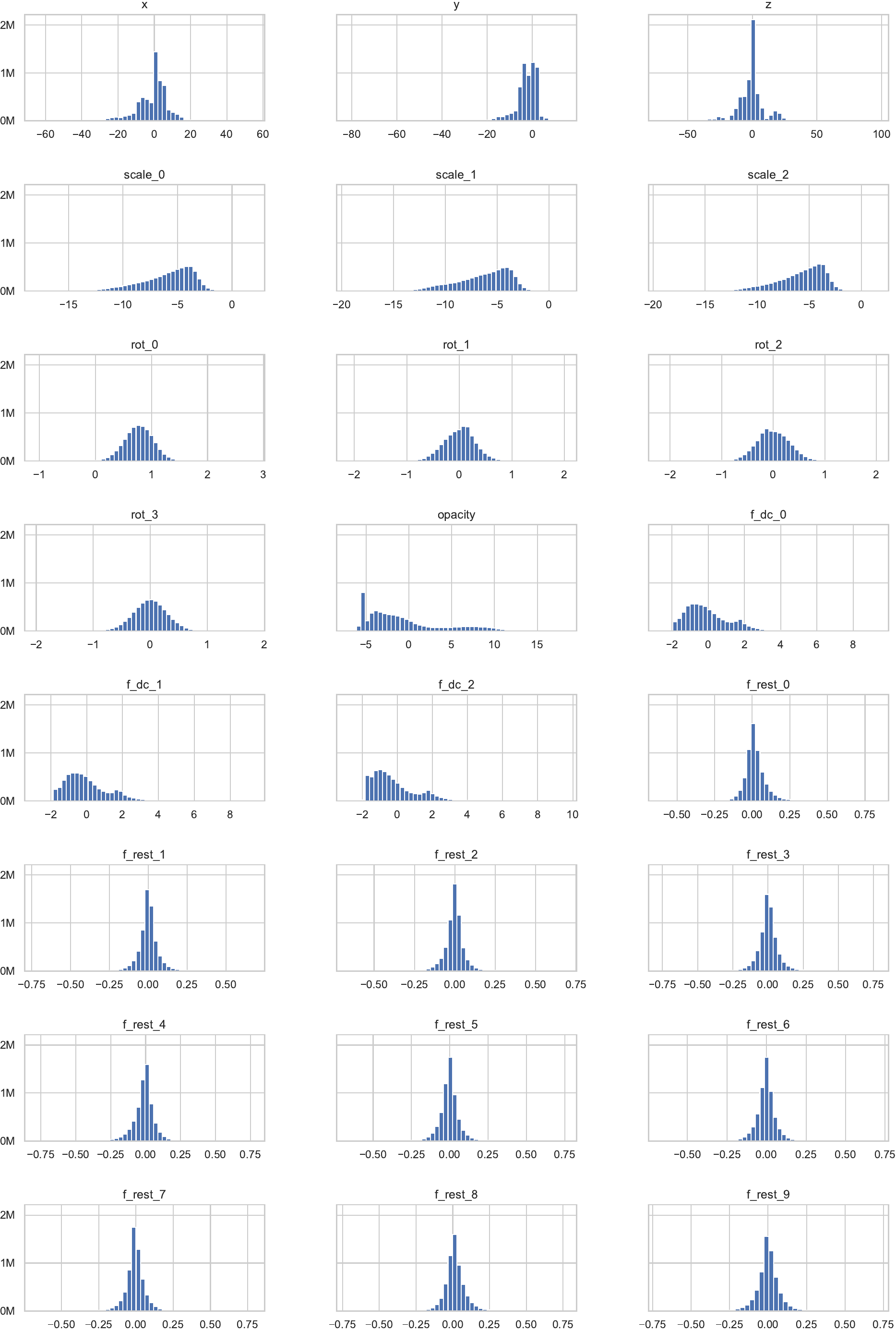}
    \caption{Histograms of attributes for all splats of the \textit{bicycle} scene, as provided by 3DGS\cite{kerbl3Dgaussians}. Only the first 12 of 48 spherical harmonics attributes are shown for brevity. All attributes are plotted as stored in their files, with their values not activated.}
    \label{fig:most_cols_hist}
\end{figure*}

\begin{figure*}[htb]
    \centering
    \includegraphics[width=\linewidth]{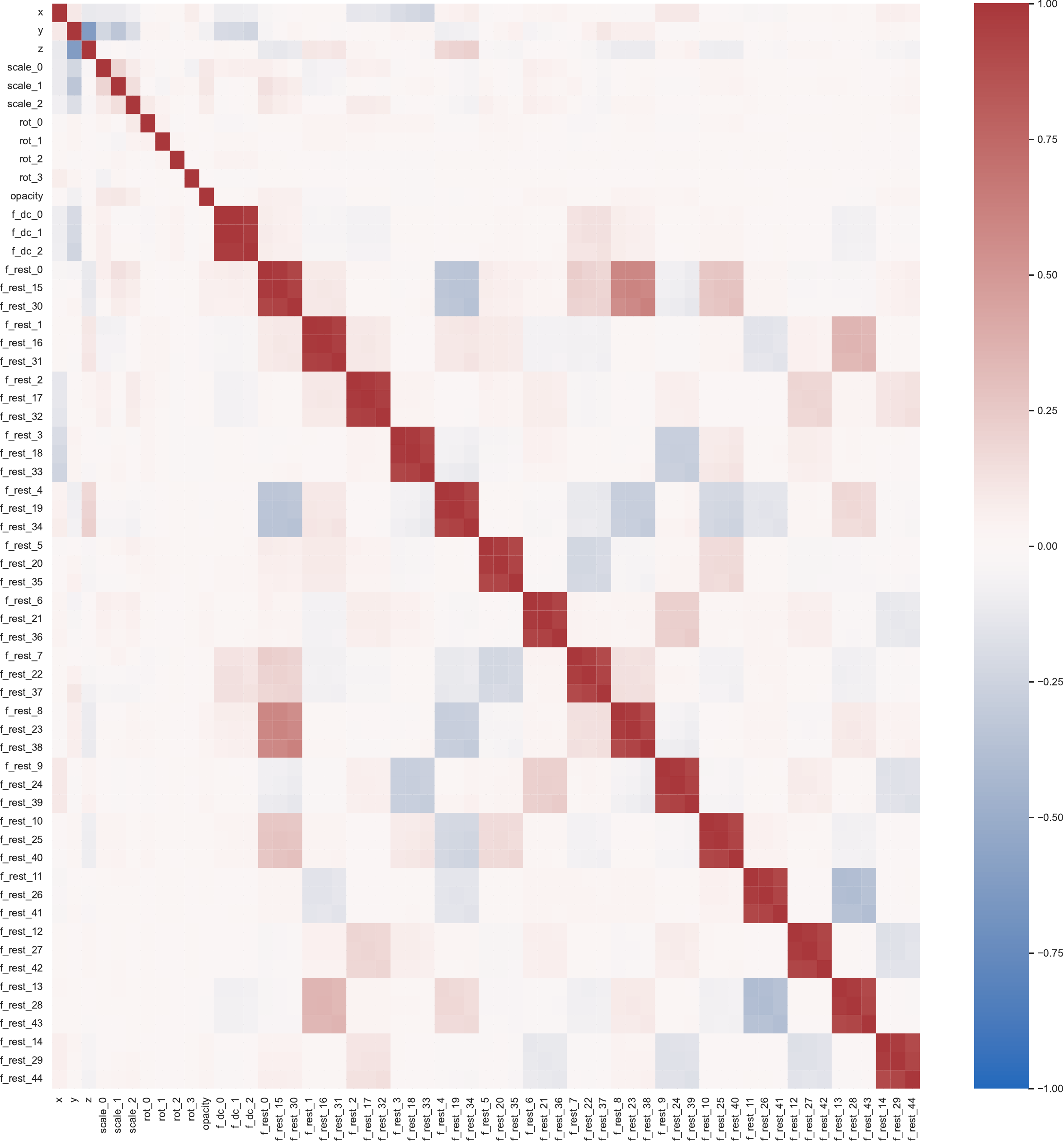}
    \caption{
    A correlation heatmap for all attributes of all splats of the \textit{bicycle} scene, as provided by 3DGS\cite{kerbl3Dgaussians}. The color channel correlation seen in the smaller version of this map in Figure \ref{fig:relevant_cols_corr} is seen repeated in the rest of the attributes of the spherical harmonics (SH), although slowly decreasing in the higher degrees. But there are additional correlations between blocks of these SH attributes, showing potential for compression. The plot also demonstrates the large part of the required space the spherical harmonics attributes take in 3DGS with third-degree SH.
    }
    \label{fig:all_cols_corr}
\end{figure*}

\section{Additional figures for compression and compaction.}
\label{ap:compressionCompaction}

\begin{figure*}
    \hspace{1cm}
    \includegraphics[width=0.4\textwidth]{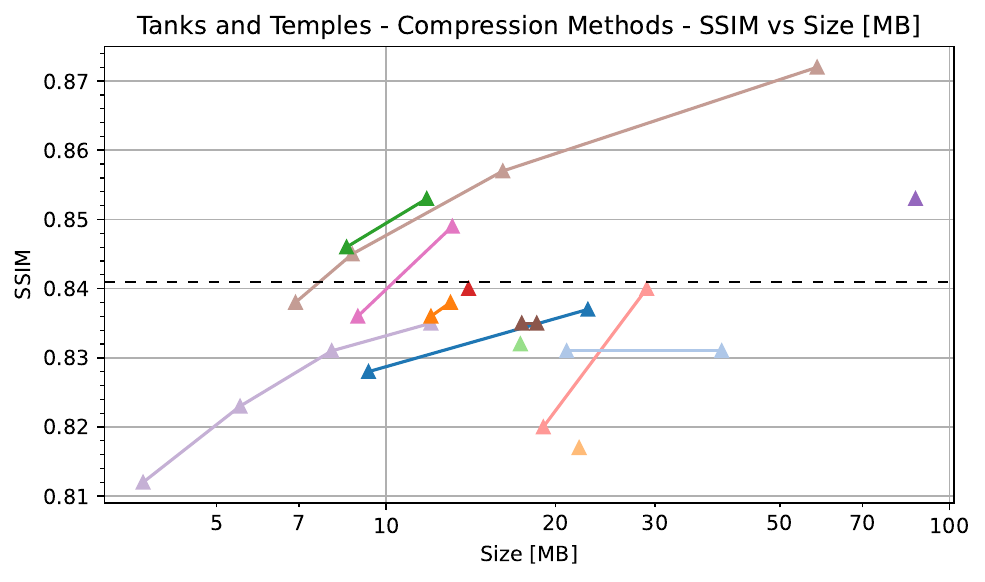}
    \includegraphics[width=0.497\textwidth]{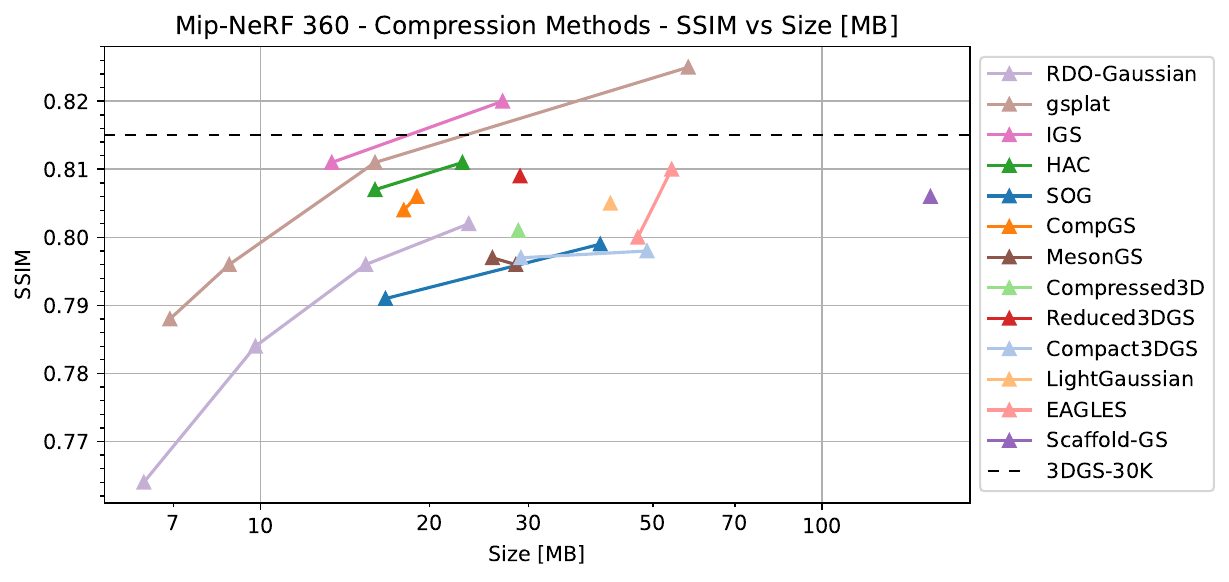}

    \hspace{1cm}
    \includegraphics[width=0.4\textwidth]{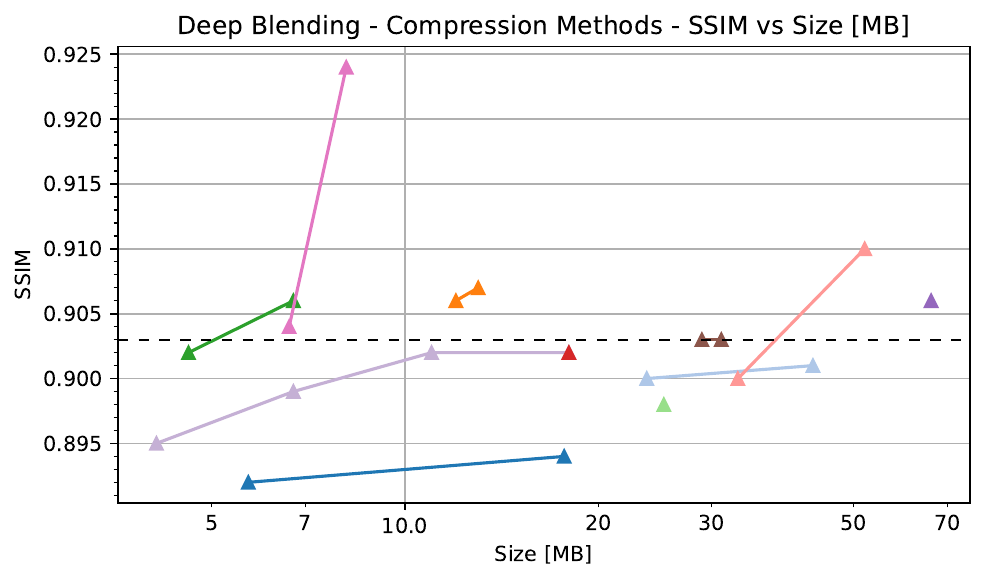}
    \includegraphics[width=0.4\textwidth]{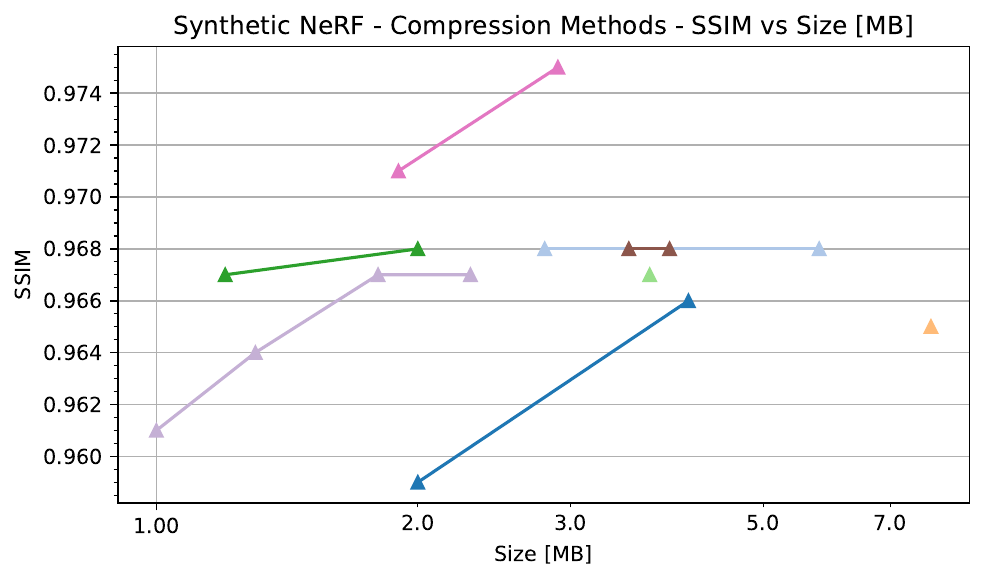}
    \caption{SSIM vs. Model Size (MB) for 3D Gaussian Splatting Compression Methods. The graphs compare different 3DGS compression methods across the \TandT{}, \MipNeRF{}, \DeepBlending{}, and \SyntheticNeRF{} datasets. The x-axis represents the model size (in MB), while the y-axis represents the SSIM, indicating the visual quality.}
\end{figure*}

\begin{figure*}
    \hspace{1cm}
    \includegraphics[width=0.4\textwidth]{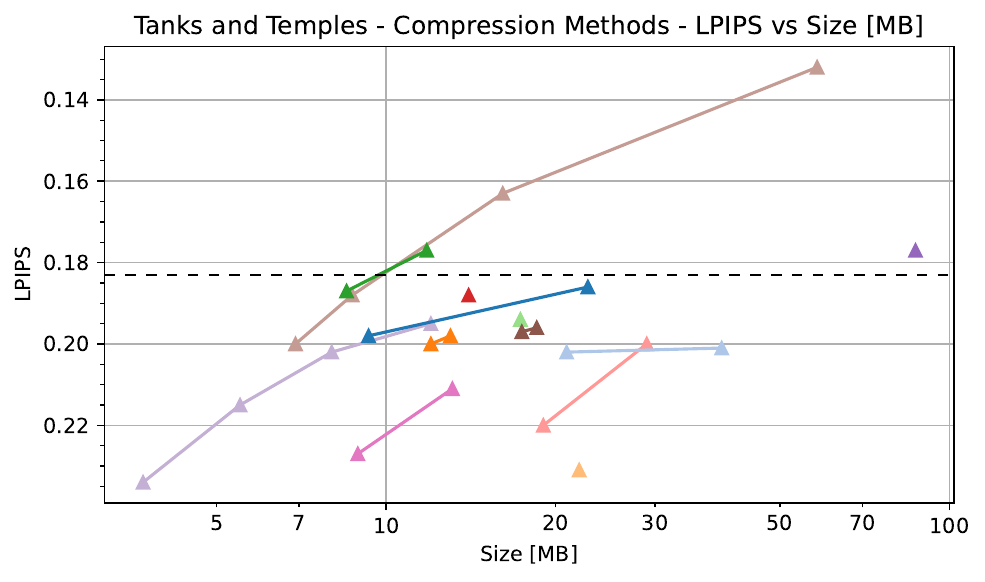}
    \includegraphics[width=0.497\textwidth]{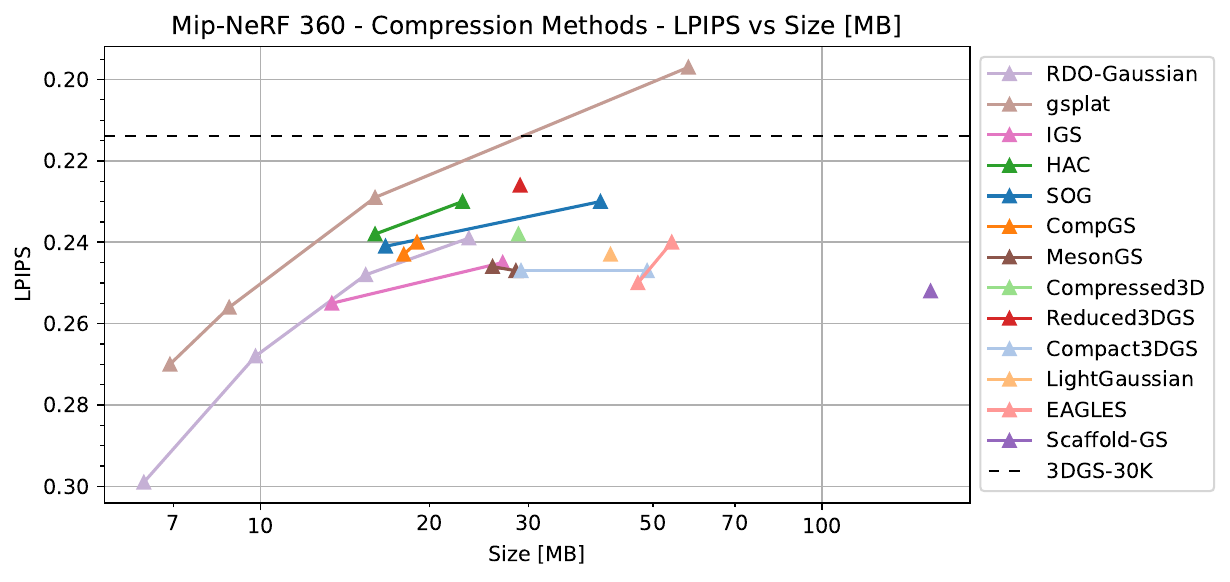}

    \hspace{1cm}
    \includegraphics[width=0.4\textwidth]{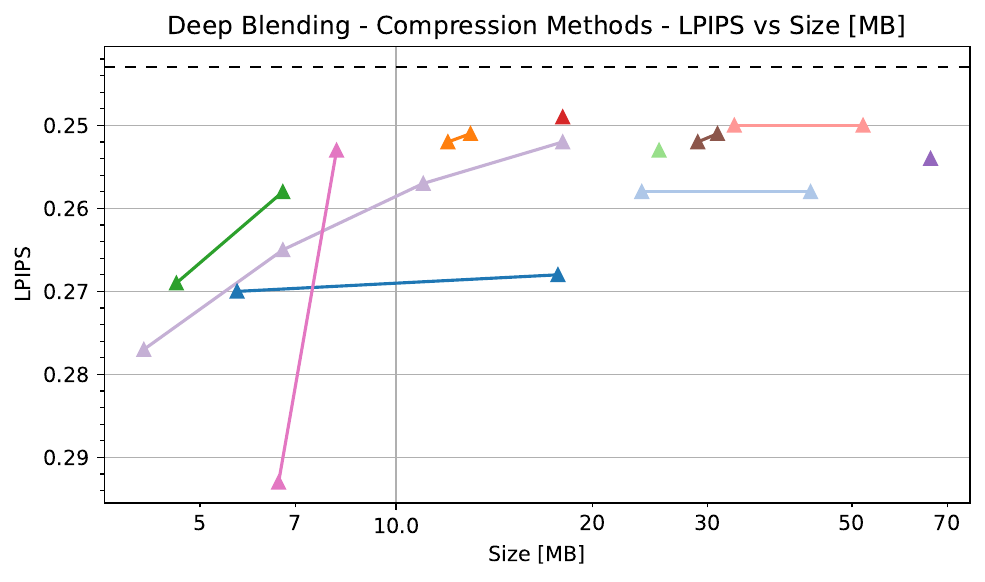}
    \includegraphics[width=0.4\textwidth]{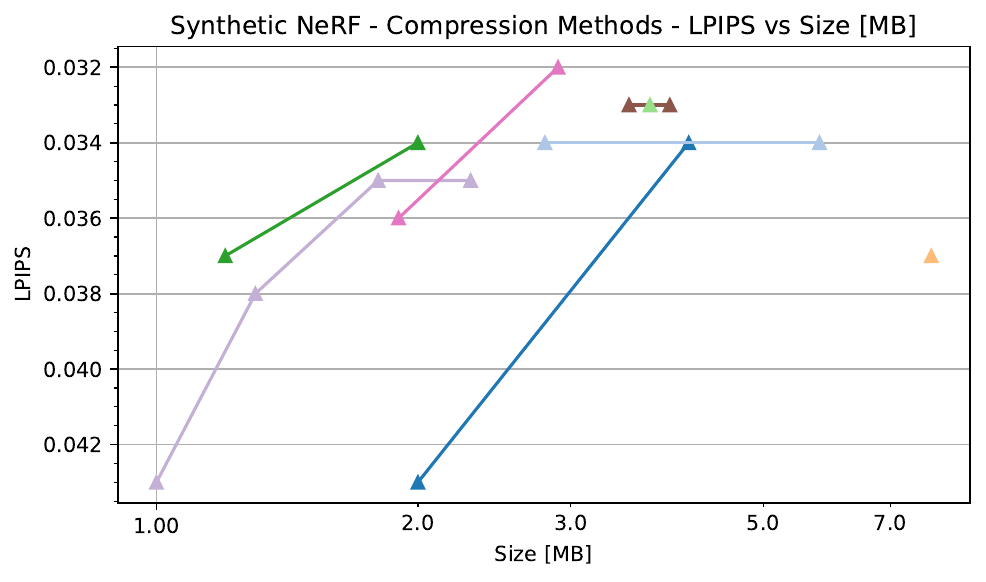}
    \caption{LPIPS vs. Model Size (MB) for 3D Gaussian Splatting Compression Methods. The graphs compare different 3DGS compression methods across the \TandT{}, \MipNeRF{}, \DeepBlending{}, and \SyntheticNeRF{} datasets. The x-axis represents the model size (in MB), while the y-axis represents the LPIPS, indicating the visual quality.}
\end{figure*}

\begin{figure*}
    \hspace{1cm}
    \includegraphics[width=0.4\textwidth]{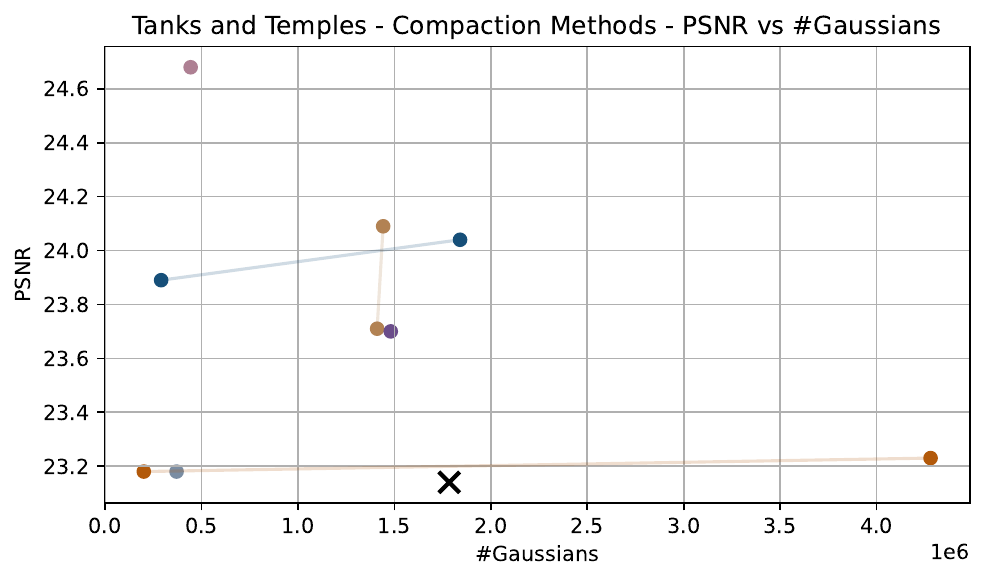}
    \includegraphics[width=0.4\textwidth]{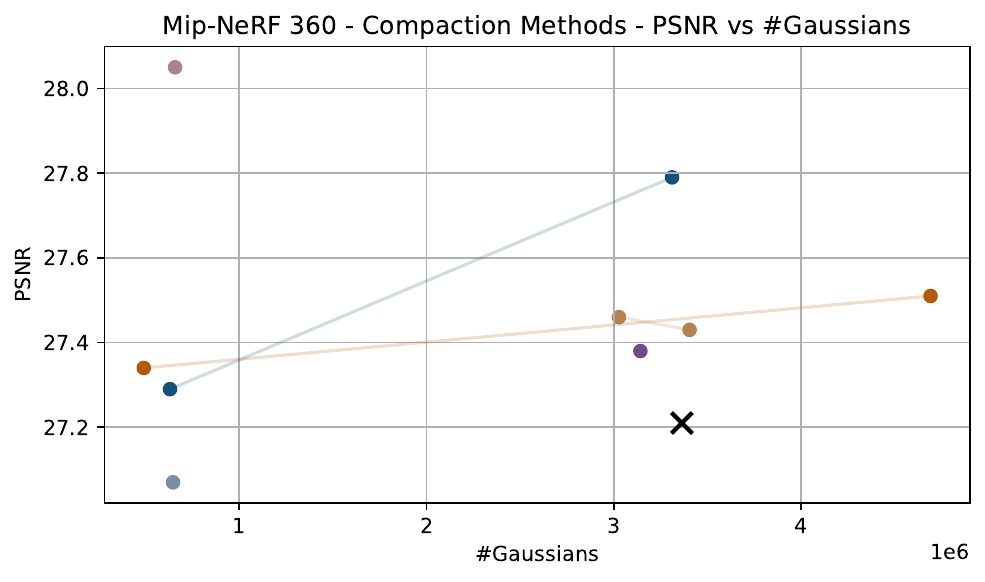}

    \hspace{1cm}
    \includegraphics[width=0.497\textwidth]{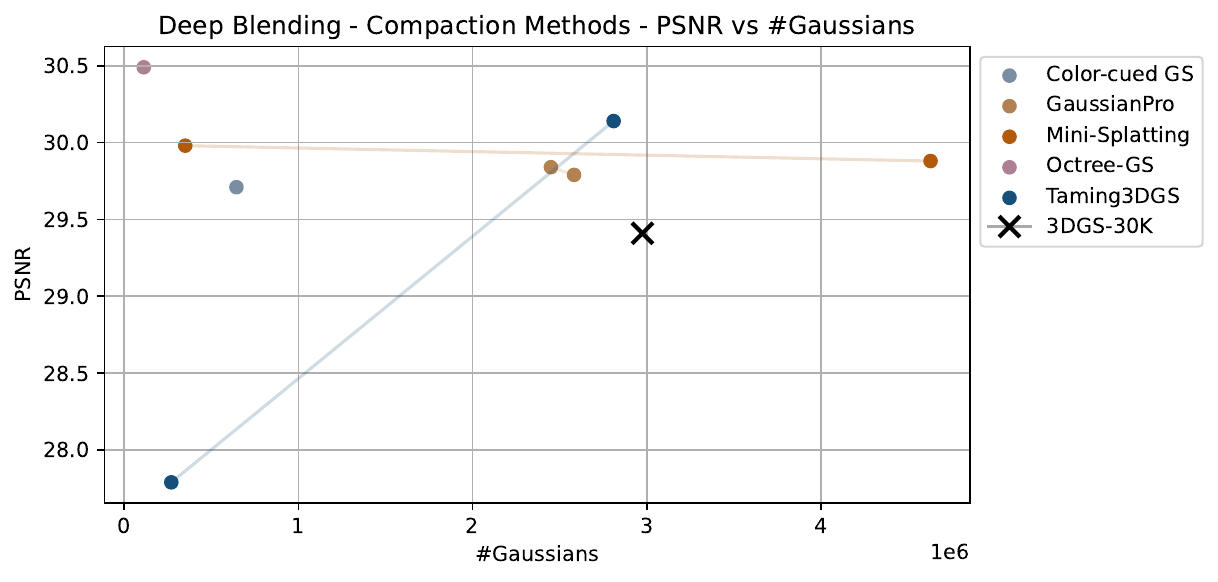}
    \caption{PSNR vs. \#Gaussians $\times 10^6$ for 3D Gaussian Splatting Compression Methods. The graphs compare different 3DGS compression methods across the \TandT{}, \MipNeRF{}, \DeepBlending{}, and \SyntheticNeRF{} datasets. The x-axis represents the number of Gaussians $\times 10^6$, while the y-axis represents the PSNR, indicating the visual quality.}
\end{figure*}

\begin{figure*}
    \hspace{1cm}
    \includegraphics[width=0.4\textwidth]{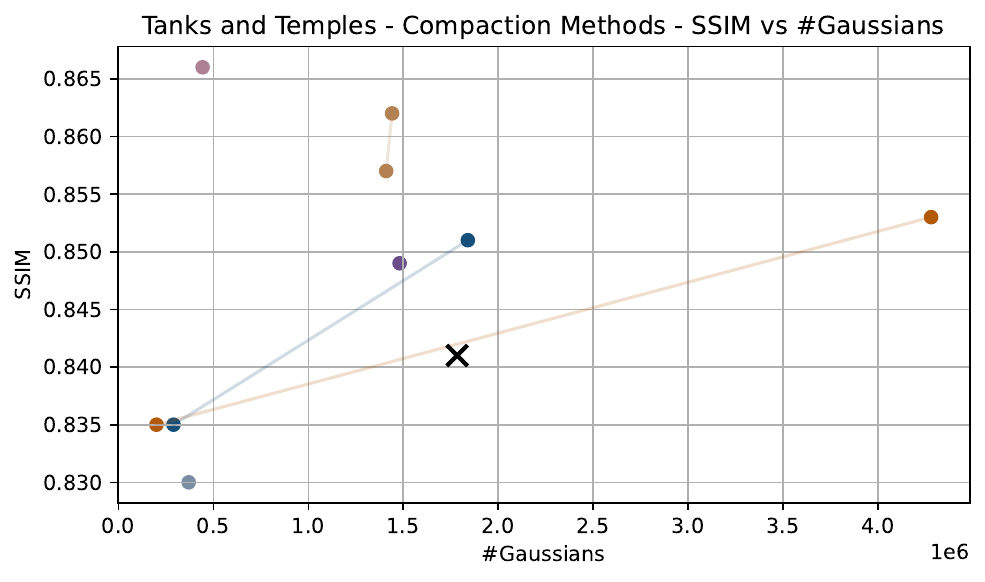}
    \includegraphics[width=0.4\textwidth]{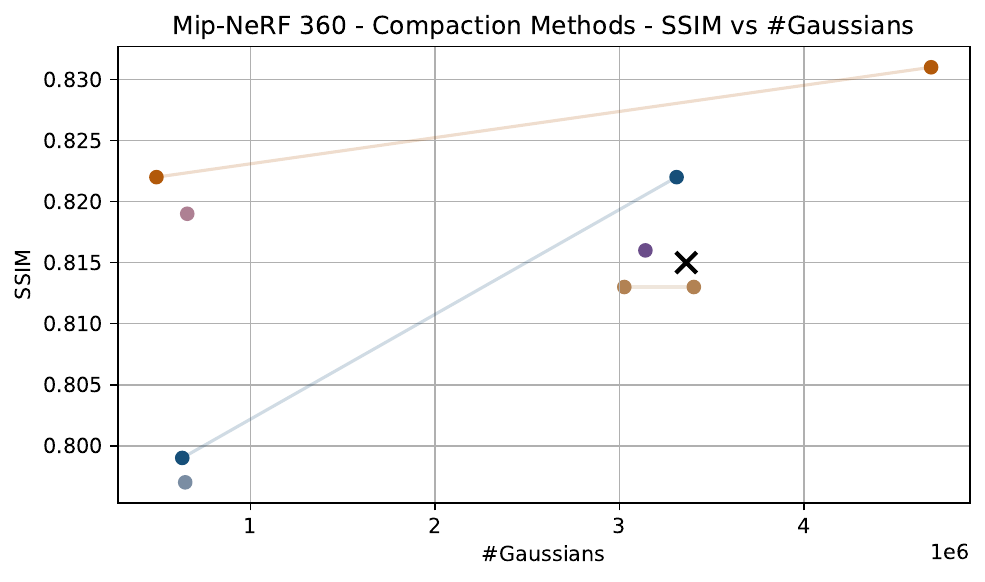}
    
    \hspace{1cm}
    \includegraphics[width=0.497\textwidth]{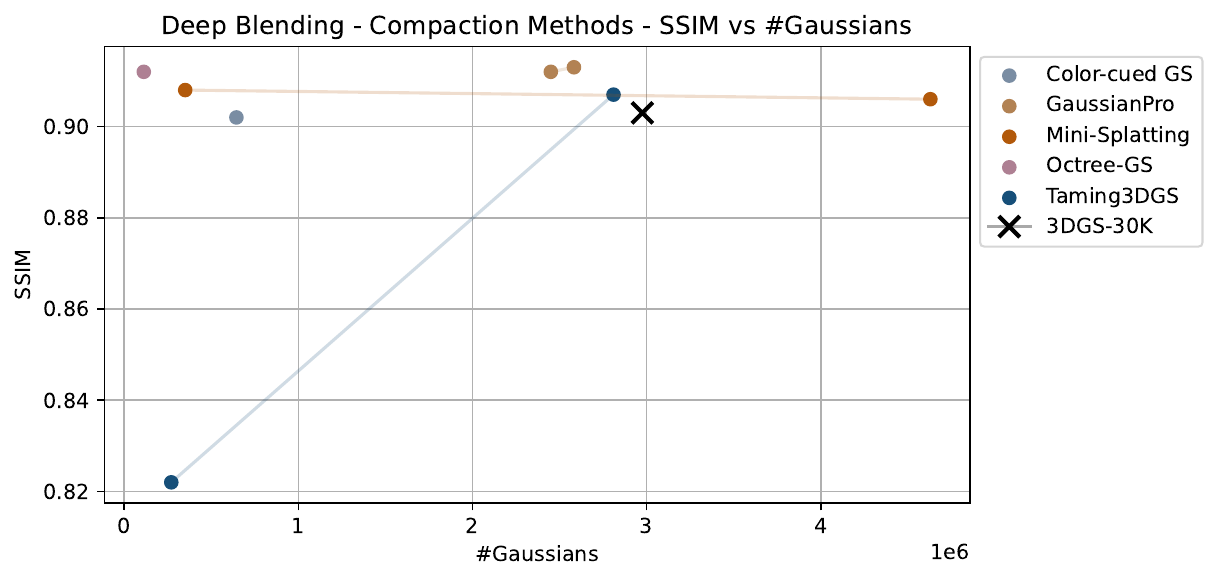}
    \caption{SSIM vs. \#Gaussians $\times 10^6$ for 3D Gaussian Splatting Compression Methods. The graphs compare different 3DGS compression methods across the \TandT{}, \MipNeRF{}, \DeepBlending{}, and \SyntheticNeRF{} datasets. The x-axis represents the number of Gaussians $\times 10^6$, while the y-axis represents the SSIM, indicating the visual quality.}
\end{figure*}

\begin{figure*}
    \hspace{1cm}
    \includegraphics[width=0.4\textwidth]{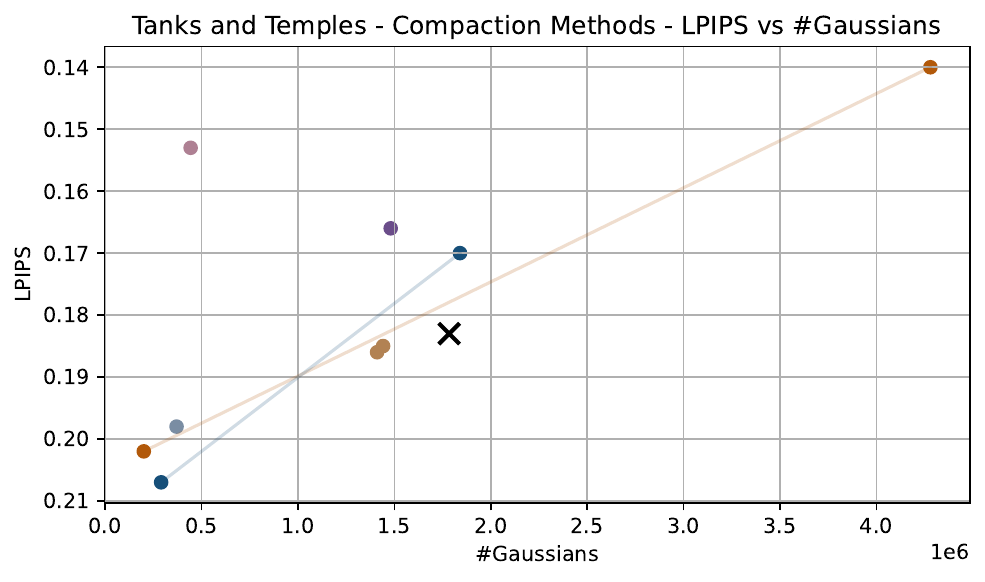}
    \includegraphics[width=0.4\textwidth]{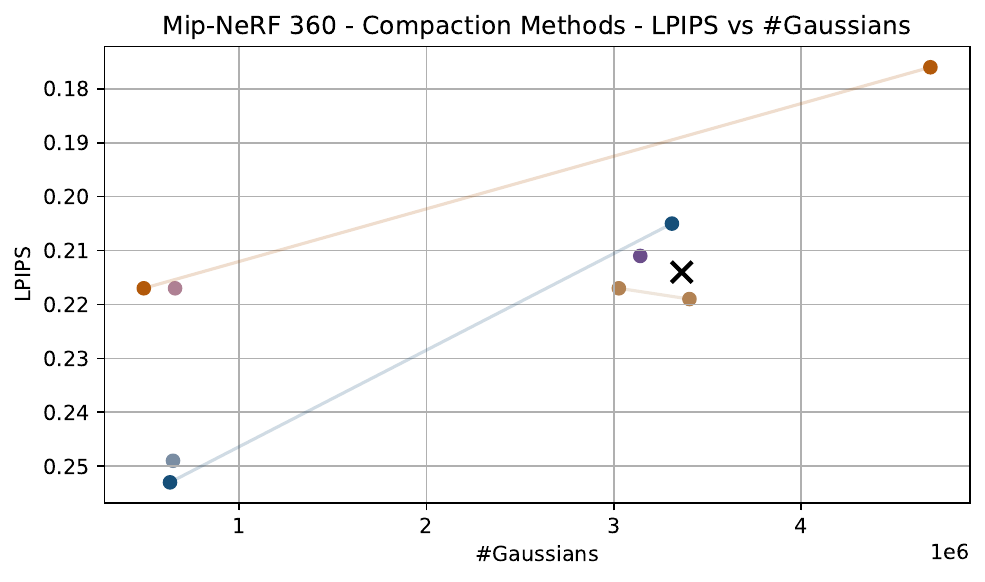}

    \hspace{1cm}
    \includegraphics[width=0.497\textwidth]{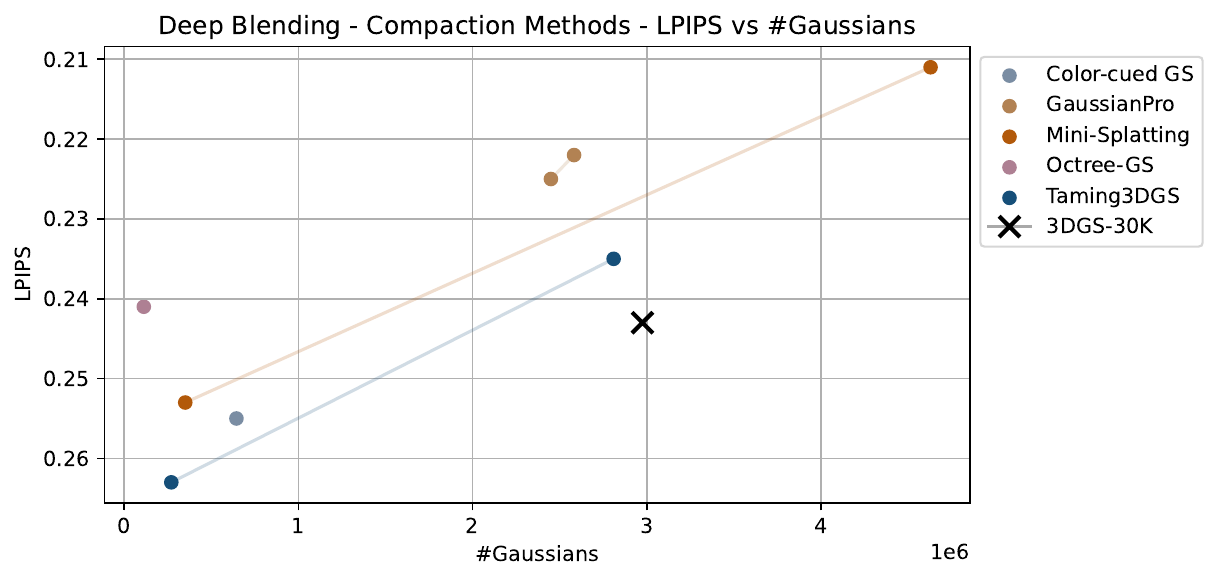}
    \caption{LPIPS vs.  \#Gaussians $\times 10^6$ for 3D Gaussian Splatting Compression Methods. The graphs compare different 3DGS compression methods across the \TandT{}, \MipNeRF{}, \DeepBlending{}, and \SyntheticNeRF{} datasets. The x-axis represents the  number of Gaussians $\times 10^6$, while the y-axis represents the LPIPS, indicating the visual quality.}
\end{figure*}

\end{document}